%% file: main.tex
\documentclass{article}

\PassOptionsToPackage{numbers, compress}{natbib}

\usepackage[preprint]{arxiv}

\usepackage[utf8]{inputenc} 
\usepackage[T1]{fontenc}    
\usepackage{url}            
\usepackage{booktabs}       
\usepackage{amsfonts}       
\usepackage{nicefrac}       
\usepackage{microtype}      
\usepackage{xcolor}         
\usepackage{colortbl}

\input{math_definition}

\usepackage{xspace}
\usepackage{graphicx}
\newcommand{\ourmethod}{{\sc {OVO}}\xspace}
\newcommand{\ourmethodbold}{{\textbf{\textsc {OVO}}}\xspace}
\newcommand\mypara[1]{\vspace{1.0mm}\noindent\textbf{#1}}
\usepackage{multirow}
\usepackage{bbding}
\usepackage{wrapfig} 
\usepackage{caption, subcaption}
\usepackage{amssymb}
\usepackage{pifont}
\usepackage{sidecap}
\usepackage{pifont}
\sidecaptionvpos{figure}{c}
\usepackage{enumitem}

\usepackage[citecolor=citecolor,pagebackref=true,breaklinks=true,letterpaper=true,colorlinks,bookmarks=false]{hyperref}
\newcommand{\cmark}{\ding{51}}%
\def\eg{{\em e.g.}}
\def\ie{{\em i.e.}}

\newcommand{\app}{{\color{purple}{Appendix}}\xspace}

\definecolor{ForestGreen}{rgb}{0.13, 0.55, 0.13}
\definecolor{Maroon}{rgb}{0.69, 0.19, 0.0}
\definecolor{my_pink}{rgb}{1,0.43,0.41}
\definecolor{my_gray}{rgb}{0,0,0.2}
\definecolor{citecolor}{RGB}{30,130,255}
\definecolor{LightCyan}{rgb}{0.88,1,1}

\definecolor{fig1_table}{RGB}{30,144,221}
\definecolor{fig1_object}{RGB}{112, 48, 160}
\definecolor{fig1_road}{RGB}{190,0,192}
\definecolor{fig1_car}{RGB}{96,144,235}
\definecolor{fig2_loss}{RGB}{131,214,208}

\definecolor{unknown}{RGB}{153, 153, 153}
\definecolor{car}{RGB}{100, 150, 245}
\definecolor{bicycle}{RGB}{100, 230, 245}
\definecolor{motorcycle}{RGB}{30, 60, 150}
\definecolor{truck}{RGB}{80, 30, 180}
\definecolor{otherVehicle}{RGB}{100, 80, 250}
\definecolor{person}{RGB}{255, 30, 30}
\definecolor{bicyclist}{RGB}{255, 40, 200}
\definecolor{motorcyclist}{RGB}{150, 30, 90}
\definecolor{road}{RGB}{255, 0, 255}
\definecolor{parking}{RGB}{255, 150, 255}
\definecolor{sidewalk}{RGB}{75, 0, 75}
\definecolor{otherGround}{RGB}{175, 0, 75}
\definecolor{building}{RGB}{255, 200, 0}
\definecolor{fence}{RGB}{255, 120, 50}
\definecolor{vegetation}{RGB}{0, 175, 0}
\definecolor{trunk}{RGB}{135, 60, 0}
\definecolor{terrain}{RGB}{150, 240, 80}
\definecolor{pole}{RGB}{255, 240, 150}
\definecolor{trafficSign}{RGB}{255, 0, 0}

\definecolor{ceiling}{RGB}{214, 38, 40}
\definecolor{floor}{RGB}{43, 160, 43}
\definecolor{wall}{RGB}{158, 216, 229}
\definecolor{window}{RGB}{114, 158, 206}
\definecolor{chair}{RGB}{204, 204, 91}
\definecolor{bed}{RGB}{255, 186, 119}
\definecolor{sofa}{RGB}{147, 102, 188}
\definecolor{table}{RGB}{30, 119, 181}
\definecolor{tvs}{RGB}{188, 188, 33}
\definecolor{furniture}{RGB}{255, 127, 12}
\definecolor{other}{RGB}{196, 175, 214}
\definecolor{out}{RGB}{220, 238, 230}

\definecolor{supervised}{RGB}{232, 232, 232}

\title{\ourmethodbold: Open-Vocabulary Occupancy}

\author{Zhiyu Tan$^1$\thanks{equal contribution} \quad Zichao Dong$^1$\footnotemark[1] \quad Cheng Zhang$^2$ \quad Weikun Zhang$^3$ \quad
Hang Ji$^4$ \quad Hao Li$^1$ \\ [5pt]
{$^1$Fudan University \quad  
$^2$Carnegie Mellon University} \\
{$^3$Zhejiang University \quad
$^4$Karlsruhe Institute of Technology}}

\begin{document}

\maketitle

\input{section/abstract}
\input{section/intro}
\input{section/related}

\input{section/approach}

\input{section/exp}
\input{section/disc}

{
\small
\bibliographystyle{plainnat}
\bibliography{main}
}

\clearpage
\appendix
\begin{center}
\textbf{\Large Supplementary Material}
\end{center}

In this supplementary material, we provide details and additional results omitted in the main paper. 
\begin{itemize}[leftmargin=6mm]
    \item[$\bullet$] Appendix~\ref{app_example}: additional implementation details (Section 3.2 and Section 4.1 of the main paper)
    \item[$\bullet$] Appendix~\ref{app_ablation}: additional ablation studies (Section 4.3 of the main paper)
    \begin{itemize}
        \item[$\circ$] Impact of the sample size of base classes
        \item[$\circ$] Impact of the choice of base classes
        \item[$\circ$]Impact of the choice of novel classes
    \end{itemize}
    \item[$\bullet$] Appendix~\ref{app_qualitative}: additional qualitative results (Section 4.3 of the main paper)
\end{itemize}

\input{appendix/app_content}

\input{appendix/app_ablation}
\input{appendix/app_qualitative}

\end{document}

%% file: math_definition.tex
\usepackage{amsmath,amsfonts,bm}

\def\1{\bm{1}}

\def\vp{{\bm{p}}}

\def\vv{{\bm{v}}}

\DeclareMathAlphabet{\mathsfit}{\encodingdefault}{\sfdefault}{m}{sl}
\SetMathAlphabet{\mathsfit}{bold}{\encodingdefault}{\sfdefault}{bx}{n}

\newcommand{\R}{\mathbb{R}}

%% file: section/abstract.tex
\begin{abstract}
Semantic occupancy prediction aims to infer dense geometry and semantics of surroundings for an autonomous agent to operate safely in the 3D environment. Existing occupancy prediction methods are almost entirely trained on human-annotated volumetric data. Although of high quality, the generation of such 3D annotations is laborious and costly, restricting them to a few specific object categories in the training dataset. To address this limitation, this paper proposes \textbf{Open Vocabulary Occupancy} (\ourmethod), a novel approach that allows semantic occupancy prediction of \emph{arbitrary classes} but \emph{without} the need for 3D annotations during training. Keys to our approach are (1) knowledge distillation from a pre-trained 2D open-vocabulary segmentation model to the 3D occupancy network, and (2) pixel-voxel filtering for high-quality training data generation. The resulting framework is simple, compact, and compatible with most state-of-the-art semantic occupancy prediction models. On NYUv2 and SemanticKITTI datasets, OVO achieves competitive performance compared to supervised semantic occupancy prediction approaches. Furthermore, we conduct extensive analyses and ablation studies to offer insights into the design of the proposed framework.
Our code is publicly available at: \url{https://github.com/dzcgaara/OVO}.
\end{abstract}

%% file: section/intro.tex
\section{Introduction} \label{s_intro}

In recent years, 3D scene understanding has attracted significant attention due to its potential applications in various fields, such as robotics, augmented and virtual reality~\cite{brazil2022omni3d}, and autonomous driving~\cite{geiger2013vision,semantickitti,wang2019pseudo}. One fundamental task for perceiving objects and scenes in 3D is the prediction of the occupancy status of each voxel in the scene, which provides essential information for scene understanding and scene reconstruction~\cite{cao2022monoscene,li2023voxformer,miao2023occdepth,huang2023tri}. Various methods have been proposed to tackle this problem, most of which rely on labeled 3D datasets to train a model for a single task with supervision. The majority of existing methods for occupancy networks rely on a predefined set of semantic class labels that can potentially be assigned to a voxel. The size of the label set is bounded by the number of categories in the training dataset. However, given that the English language alone encompasses hundreds of thousands of nouns, it is probable that the small size of the label set significantly limits the scalability of current Occupancy networks.

The primary reason for the limited label sets in current methods is the high cost associated with collecting and labeling 3D annotations~\cite{nyu,semantickitti}. Creating training datasets requires human annotators to assign a semantic class label to every voxel in thousands of images, which is an extremely labor-intensive and expensive task even for small label sets. As the number of labels increases, the annotation task becomes even more complex, as the human annotator must be aware of the fine-grained candidate labels. Moreover, inter-annotator consistency becomes problematic when objects in an image can fit multiple descriptions or are subject to a hierarchy of labels.

To predict the occupancy status of each voxel for arbitrary semantic classes without requiring annotations during training, we need a new approach that overcomes the limitations of existing methods. The zero-shot learning approach has been successful in object detection and segmentation tasks. However, extending this approach to the 3D scene open vocabulary task is challenging due to the complexity of the task and the need for a flexible and scalable method.

To address these issues, we propose a new method called Open Vocabulary Occupancy (OVO) to predict the occupancy status of each voxel in a 3D scene in a zero-shot manner. Overall, the OVO method leverages a combination of 3D feature extraction, voxel grouping, and open-vocabulary queries to predict the occupancy status of each voxel in a 3D scene for arbitrary semantic classes. The method is designed to overcome the limitations of existing methods that rely on a limited set of semantic class labels and require extensive annotations during training.

\begin{figure}[t!]
  \centering
   \includegraphics[width=\textwidth]{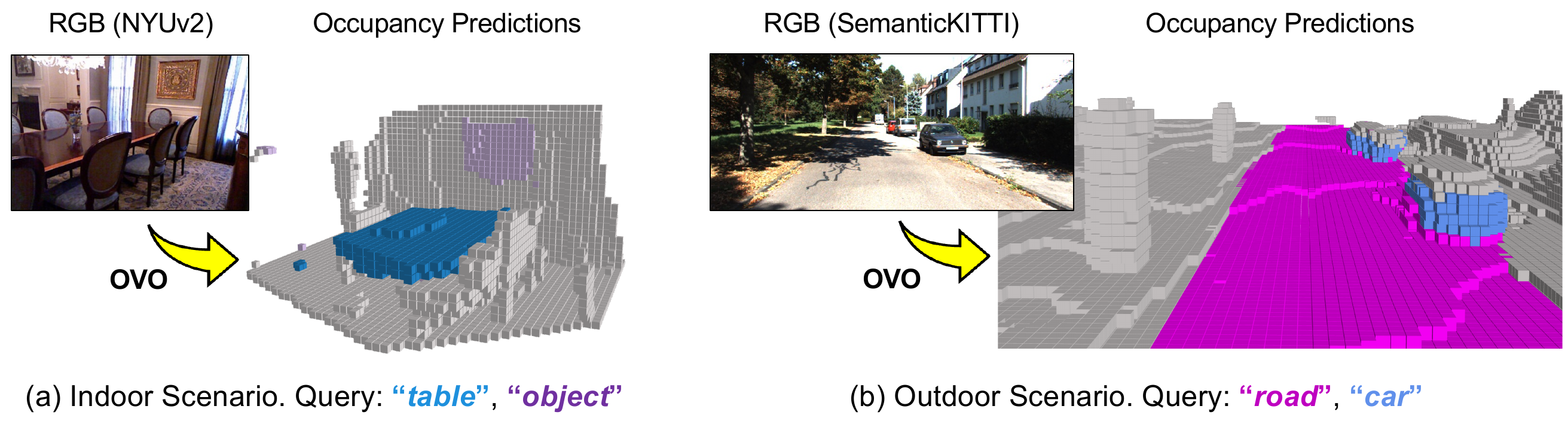}
   \caption{\small \textbf{Open-Vocabulary Occupancy (OVO)} from a monocular camera. 
   Existing occupancy networks rely on fully-labeled training examples for dense semantic occupancy prediction. 
   Given an RGB input, our method can predict dense semantic occupancy for object categories that have \emph{not} been annotated in the training data. For example, the voxels of \emph{novel} object categories can be well-captured by using arbitrary text queries in both indoor (\eg, class ``{\color{fig1_table}{\emph{table}}}'' and ``{\color{fig1_object}{\emph{object}}}'' in \textbf{(a)}) and outdoor (\eg, class ``{\color{fig1_road}{\emph{road}}}'' and ``{\color{fig1_car}{\emph{car}}}'' in \textbf{(b)}) scenarios.}
  \label{fig:sample} 
  \vspace{-2mm}
\end{figure}

Our proposed method represents the first attempt to address the open vocabulary occupancy prediction task, which is significant due to its potential for advancing 3D scene understanding. We validate the effectiveness of our proposed method through quantitative experiments on the SemanticKITTI dataset and the NYU dataset, demonstrating its superiority over existing methods in predicting occupancy status for arbitrary semantic classes. Furthermore, the flexibility of our model is highlighted through several examples shown in Figure~\ref{fig:sample}, where our method can predict the occupancy status of arbitrary semantic classes without requiring annotations during training. In summary, our proposed method offers a promising approach for predicting the occupancy status of each voxel in a 3D scene for arbitrary semantic classes. By removing the need for labeled data for each semantic class, our method is more scalable and adaptable to various applications, thus opening up new possibilities for 3D scene understanding. We believe that our proposed method has the potential to contribute significantly to the development of more robust and flexible 3D scene understanding methods, which can further advance the state-of-the-art in the field.

\mypara{Contribution.} 
(1) We present Open Vocabulary Occupancy (OVO), a model-agnostic approach designed specifically for semantic occupancy tasks. To the best of our knowledge, this is the first attempt to apply an open vocabulary methodology to the occupancy task domain.
(2) We introduce novel feature alignment objectives for knowledge distillation from an off-the-shelf 2D open-vocabulary model. We show that high-quality voxel selection for training is the key to the success of OVO.  
(3) Our proposed framework achieves competitive results on NYUv2~\cite{nyu} and SemanticKITTI~\cite{semantickitti} datasets, demonstrating the effectiveness of the OVO framework and the 2D-to-3D distillation approach in addressing complex occupancy tasks.

%% file: section/related.tex
\section{Related Work}

\mypara{3D Occupancy Prediction.}
The task of occupancy prediction involves predicting the occupancy status and corresponding semantic labels for each voxel in a given 3D scene. 
MonoScene ~\cite{cao2022monoscene} is a pioneering 3D voxel reconstruction framework that enables outdoor scene understanding using only a monocular camera. The framework incorporates a module for 2D feature line of sight projection (FLoSP) to link the 2D and 3D U-Net and a 3D context relation prior (CRP) layer to enhance contextual learning. 
VoxFormer ~\cite{li2023voxformer} is a transformer-based framework that operates in two stages. It starts by using sparse visible and occupied queries from a depth map and then propagates them to dense voxels using self-attention mechanisms.
OccDepth ~\cite{miao2023occdepth} is a stereo-based method that uses a stereo soft feature assignment module to lift stereo features to 3D space. It distills depth information from a teacher model and applies it to train a depth-augmented occupancy perception module.
TPVFormer ~\cite{huang2023tri} is a pioneering surround-view 3D reconstruction framework that utilizes sparse LiDAR semantic labels as the only source of supervision. The framework generalizes BEV to a Tri-Perspective View (TPV), which involves expressing the features of 3D space through three slices perpendicular to the $x$, $y$, and $z$ axis. TPVFormer uses 3D point queries to decode occupancy with arbitrary resolution.

\mypara{Open-Vocabulary Scene Understanding.}
Recent works, such as CLIP~\cite{radford2021learning} and ALIGN~\cite{jia2021scaling}, have pushed the limits of image-text modeling by collecting million-scale image-text pairs and training joint image-text models using contrastive learning. These models can be transferred to various classification datasets and achieve impressive performances.
\cite{zareian2021open} involves pretraining the backbone model using image captions and fine-tuning the pretrained model with detection datasets.  
ViLD~\cite{gu2021open} is a proposed two-stage detector that leverages knowledge distillation from the CLIP/ALIGN model to advance zero-shot object detection.
LSeg~\cite{li2022language} proposes an open-vocabulary segmentation method that utilizes pre-trained CLIP text encoders. The method trains an image encoder to predict pixel embeddings that are aligned with the text embeddings of their corresponding pixel labels. OpenSeg~\cite{ghiasi2022scaling} utilizes segmentation masks and their features to represent images and employs weakly supervised learning to align visual-semantic relationships based on predicted masks.
OpenScene ~\cite{peng2022openscene} utilized the pre-trained text encoder from CLIP to achieve open-vocabulary 3D scene understanding. Their method involved fusing 2D and 3D features and aligning the resulting combination with text embeddings.
The Semantic Abstraction (SemAbs)~\cite{ha2022semantic} method takes an RGB-D image as input, which equips 2D Vision-Language Models with 3D spatial capabilities by leveraging relevancy maps from CLIP, enabling open-world 3D scene understanding while maintaining zero-shot robustness.
In CLIP-FO3D~\cite{zhang2023clip}, local features are extracted by splitting region crops to super-pixels and patches. Pixel-level CLIP features are then fused by multi-scale region crops. Their 3D scene understanding model is trained to align with the above pixel-level CLIP features.

%% file: section/approach.tex
\section{Proposed Method: Open Vocabulary Occupancy}

\subsection{Background}
\label{s_background}
The task of semantic occupancy prediction aims to predict the geometry and semantics of voxels within a 3D scene. In this work, we focus on camera-based occupancy prediction. However, our framework is model-agnostic and can be integrated with other input modalities such as 3D scans and point clouds. We begin with reviewing key concepts relevant to the understanding of our approach. 

\mypara{Architecture for Semantic Occupancy Prediction.} An occupancy network~\cite{cao2022monoscene,li2023voxformer} mainly consists of four components: (1) a 2D network ($E_\text{2D}$) that extracts the 2D features from an input image, (2) a 2D-to-3D feature transformation module that lifts the 2D representations to the 3D space, (3) a 3D network ($E_\text{3D}$) that refines the 3D feature map by learning the semantic relationships within the scene, and (4) an occupancy head that infers geometry and semantics for each voxel in the 3D feature map.

\mypara{Notations.} 
In open vocabulary tasks for 2D images, such as object detection or segmentation, dataset categories are divided into base and novel subsets ($C_B$ and $C_N$). Task-specific features align to the CLIP feature space on base class $C_B$ through vision distillation and text distillation. For 3D open vocabulary tasks, challenges include complex 3D data representation, limited annotated data availability, scalability issues, and the lack of established frameworks and benchmarks. 

In this paper, we present a model-agnostic framework designed for open-vocabulary occupancy tasks. To validate our approach, we utilize MonoScene~\cite{cao2022monoscene} as the occupancy network, the CLIP text encoder for text distillation, and LSeg~\cite{li2022language} as the open vocabulary image encoder for vision distillation, which predicts pixel embeddings that align with the text embeddings of their corresponding pixel labels.

\begin{figure}[t]
  \centering
  \includegraphics[width=1.0\textwidth,page=1]{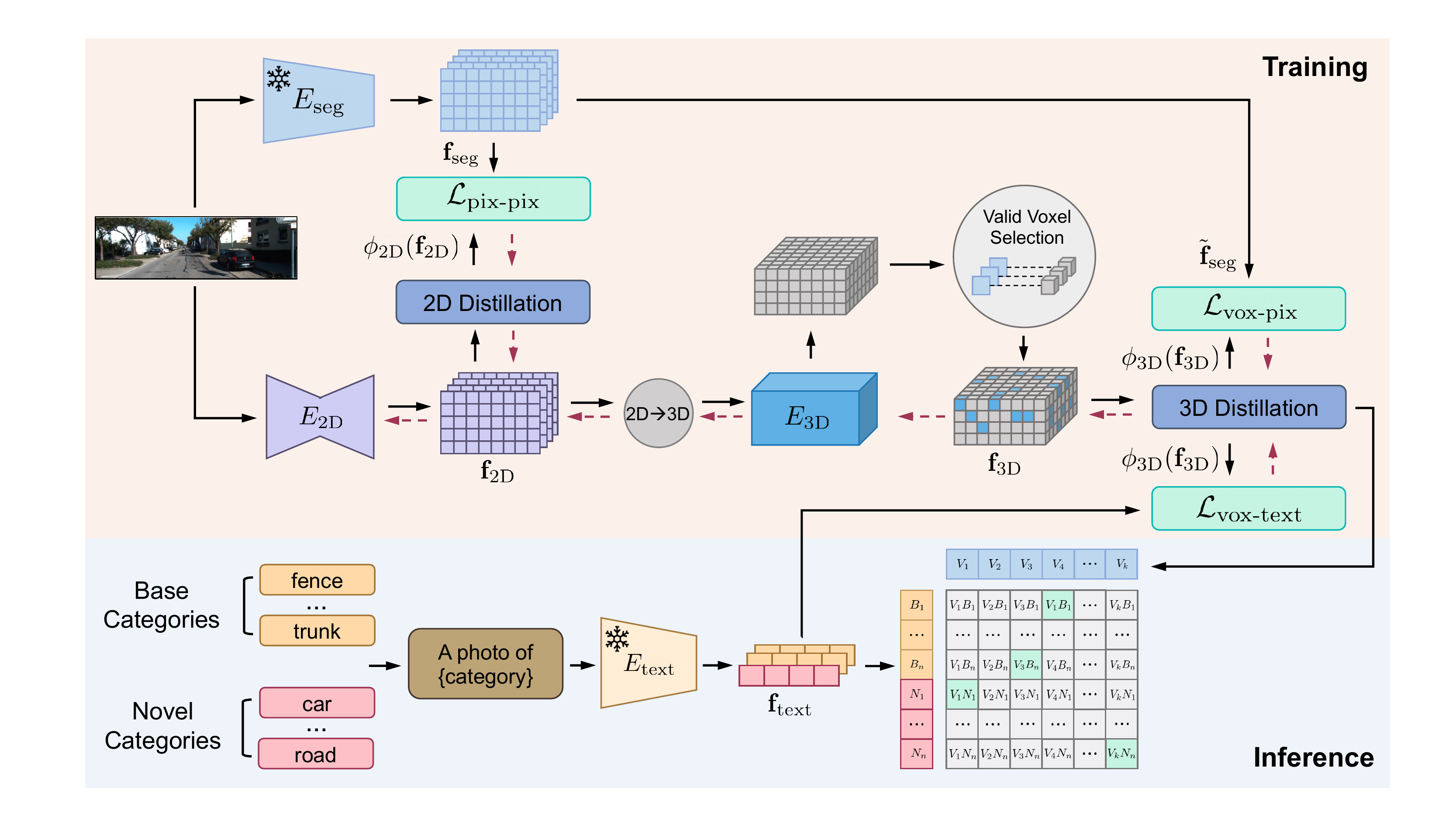}
  \caption{\small \textbf{Overall pipeline of Open-Vocabulary Occupancy (OVO)}. Our framework enables knowledge distillation (see Section~\ref{ss_align}) from a pre-trained 2D \emph{open-vocabulary} segmentation model ($E_\text{seg}$ in the upper row) to the 3D occupancy network ($E_\text{2D}$--$E_\text{3D}$ in the middle row). We also propose a simple yet effective voxel filtering mechanism for high-quality training data selection (see Section~\ref{ss_selection} and Figure~\ref{fig:pairs}). The whole pipeline is trained end-to-end and only the parameters of the 3D occupancy network will be updated. {\color{purple}{Red dashed}} arrows indicate the backward pass for the three {\color{fig2_loss}{\textbf{feature alignment losses}}}.
  During inference, text embeddings (bottom row) of both base and novel categories can be used to predict the semantic label for each voxel. 
  }
  \label{fig:framework} \vspace{-2mm}
\end{figure}

\subsection{Knowledge Distillation via Feature Alignment} 
\label{ss_align}
Given a pre-trained 2D open-vocabulary segmentation model, how can we distill its knowledge of recognizing arbitrary object categories into a 3D occupancy network? We provide details on the three alignment techniques used to achieve to goal. Figure~\ref{fig:framework} illustrates our overall framework.

\mypara{Voxel-to-Pixel Alignment.} The voxel-to-pixel alignment is a key module in OVO, which allows voxel features to mimic 2D features generated by LSeg, enabling the inference of voxel categories using text embeddings. First, we filter the voxel features generated by the 3D backbone to obtain valid voxels. These valid voxels are then fed into the 3D distillation module to align the dimensions of the 3D features with the 2D features. Then, $\mathcal{L}_\text{vox-pix}$ is computed based on the similarity between the 3D features and the 2D features. The formulation is as follows:
\begin{align}\label{eq:rw}
    \mathcal{L}_\text{vox-pix} = \frac{1}{|\Omega|}\sum_{{\vv} \in \Omega} w \bigl( \rho (\vv) \bigl) \Bigl[ 1 - \bigl\langle \phi_\text{3D} ( \textbf{f}_\text{3D} ), 
    \tilde{\textbf{f}}_\text{seg}  \bigl\rangle \Bigl],
\end{align}
where $\rho(\cdot)$ is a mapping function that maps a 3D coordinate $\vv$ to a 2D coordinate, $\textbf{f}_\text{3D} = E_\text{3D} (\vv)$ is the voxel feature generated by MonoScene 3D encoder, $\tilde{\textbf{f}}_\text{seg} = E_\text{seg} \bigl( \rho (\vv) \bigl)$ refers to LSeg feature corresponding to voxel located at $\vv$, 
$\Omega$ is the valid voxel set obtained by the valid voxel selector, and
$w \bigl( \rho (\vv) \bigl)$ is the confidence score predicted by LSeg and we take it as a re-weight value in training.

\mypara{Voxel-to-Text alignment.} We generate the text embeddings offline by feeding the category texts with prompt templates, \eg, ``a photo of \{category\}'', into the text encoder $E_\text{text}$. Then, we compute cosine similarities between the voxel and text embeddings. A softmax activation is applied to generate final predictions. Our goal is to train the voxel embeddings such that they can be classified by text embeddings. As depicted in Figure~\ref{fig:framework}, it replaces the learnable classifier with text embeddings. Only $\tau(C_B)$, the text embeddings of $C_B$, are used for training. For the voxels that do not match any ground truth in $C_B$, they are assigned to the background category. The loss for voxel-to-text alignment can be written as: 
\begin{align}
    \mathcal{L}_\text{vox-text} = \frac{1}{|\Omega|}\sum_{{\vv} \in \Omega} 1 - \bigl\langle \phi_\text{3D} ( \textbf{f}_\text{3D} ), \textbf{f}_\text{text}  \bigl\rangle,
\end{align}
where $\textbf{f}_\text{text} = E_\text{text} \bigl( \tau ({\vv}) \bigl)$ and $\tau(\cdot)$ returns the category label (in text form) assigned to voxel $\vv$.

\mypara{Pixel-to-Pixel Alignment.} Pixel-to-Pixel alignment aims at transferring knowledge from a teacher image encoder $E_\text{seg}$ to a student encoder $E_\text{2D}$. The categories of these pixels predicted by LSeg can belong to $C_B$ or $C_N$, as the network is capable of generalizing. Specifically, we first use the 2D distillation to lift the dimension of the image feature generated by the 2D backbone. Then we apply 2D pixel-to-pixel alignment for the 2D feature map obtained from Lseg~\cite{li2022language} and the image feature after distillation. The alignment loss function is defined as follows:
\begin{align}
    \mathcal{L}_\text{pix-pix} = \sum_{\vp} 1 - \bigl\langle \phi_\text{2D} (\textbf{f}_\text{2D}), \textbf{f}_\text{seg} \bigl\rangle,
\end{align}
where $\textbf{f}_\text{seg} = E_\text{seg} (\vp) \in \R^{C_1}$ is the 2D feature for pixel $\vp$ from the LSeg image encoder,   $\textbf{f}_\text{2D} = E_\text{2D} (\vp) \in \R^{C_2}$ is the 2D feature for pixel $\vp$ from the MonoScene 2D encoder, and $\phi_\text{2D}(\cdot)$: $\R^{C_2} \rightarrow \R^{C_1}$ is the alignment layer (convolutions or MLPs).

\mypara{Final Loss.}
Our final loss for one image can be written as
{\begin{equation}
    \mathcal{L}_\text{total} = \lambda_1 \mathcal{L}_\text{pix-pix}  + \lambda_2 \mathcal{L}_\text{vox-pix} + \lambda_3 \mathcal{L}_\text{vox-text},
\end{equation}}
where, $\lambda_1$, $\lambda_2$ and $\lambda_3$ are hyper-parameters to balance the three alignment losses, we set it $\lambda_1=0.1$, $\lambda_2=1$ and $\lambda_3=1$ in all the experiments. The whole pipeline can be trained end-to-end efficiently.

\input{table/pairs}

\subsection{Training Voxel Selection}
\label{ss_selection}
Due to the range of viewpoints in the image, object occlusion, and inaccuracies in 2D model predictions, creating sample pairs directly using coordinate mapping may result in a substantial number of incorrect pairing associations.
Based on our observations and experiments, the inappropriate selection of training pairs greatly affects the training performance for feature alignment. In order to tackle this problem, we have designed three strategies for keeping only the high-quality voxels for model training, as illustrated in Figure~\ref{fig:pairs}.
\begin{itemize}[leftmargin=3mm,topsep=0pt,itemsep=0pt,noitemsep]
    \item \mypara{Image Range Filter.} Based on the geometric relationship, voxels in the 3D space whose 2D projection falls outside the range of the 2D image, as determined by the camera pose, will be excluded. Since we are unable to assign 2D labels to these voxels, it is reasonable to filter them out.

    \item \mypara{Occlusion Filter.} 
    This strategy is designed to tackle the problem of occlusion by selectively filtering out voxels with a significant amount of occlusion. As semantic information is derived exclusively from voxels, we eliminate those that are obscured by other voxels, as they are unable to contribute to the semantic comprehension of the scene.

    \item \mypara{Label Consistency Filter.} 
    In order to eliminate inaccurate predictions from LSeg~\cite{li2022language}, we develop an additional filtering process to eliminate voxels that has an inconsistent semantic label to their corresponding 2D pixel. 
\end{itemize}

The diagram in Figure~\ref{fig:pairs} illustrates the process of selecting valid voxels. The light grey voxels represent the ones that have been filtered out by the image range filter, indicating that they are not within the desired range of the image. The dark grey voxels, on the other hand, are occluded by other objects in the scene, making them unreliable for further analysis. The solid arrow in the diagram represents a valid voxel that has passed the filtering process and can be used for subsequent operations. However, there are also voxels with inconsistent labels, indicated by the dashed arrow. 

\subsection{Optimization and Inference}
In the data preprocessing stage, we divide semantic labels into base classes and novel classes. We merge all the novel classes into a single unknown class for training. Specifically, during the training process, the number of classes is equal to the number of base classes plus one.
For each pixel, we calculate the similarity between the pixel feature obtained by the LSeg encoder and all word embeddings. The highest score is then utilized as the confidence weight to reweight $\mathcal{L}_{vp}$ during training.
The 2D distillation module is inference-free, meaning that during inference, we only need to obtain the voxel features distilled by the 3D distillation module. During the inference stage, we determine the specific labels for novel classes by calculating the similarity between voxel features and word embeddings.

%% file: table/pairs.tex
\begin{SCfigure}[]
  \begin{minipage}{0.55\linewidth}
    \includegraphics[width=\linewidth]{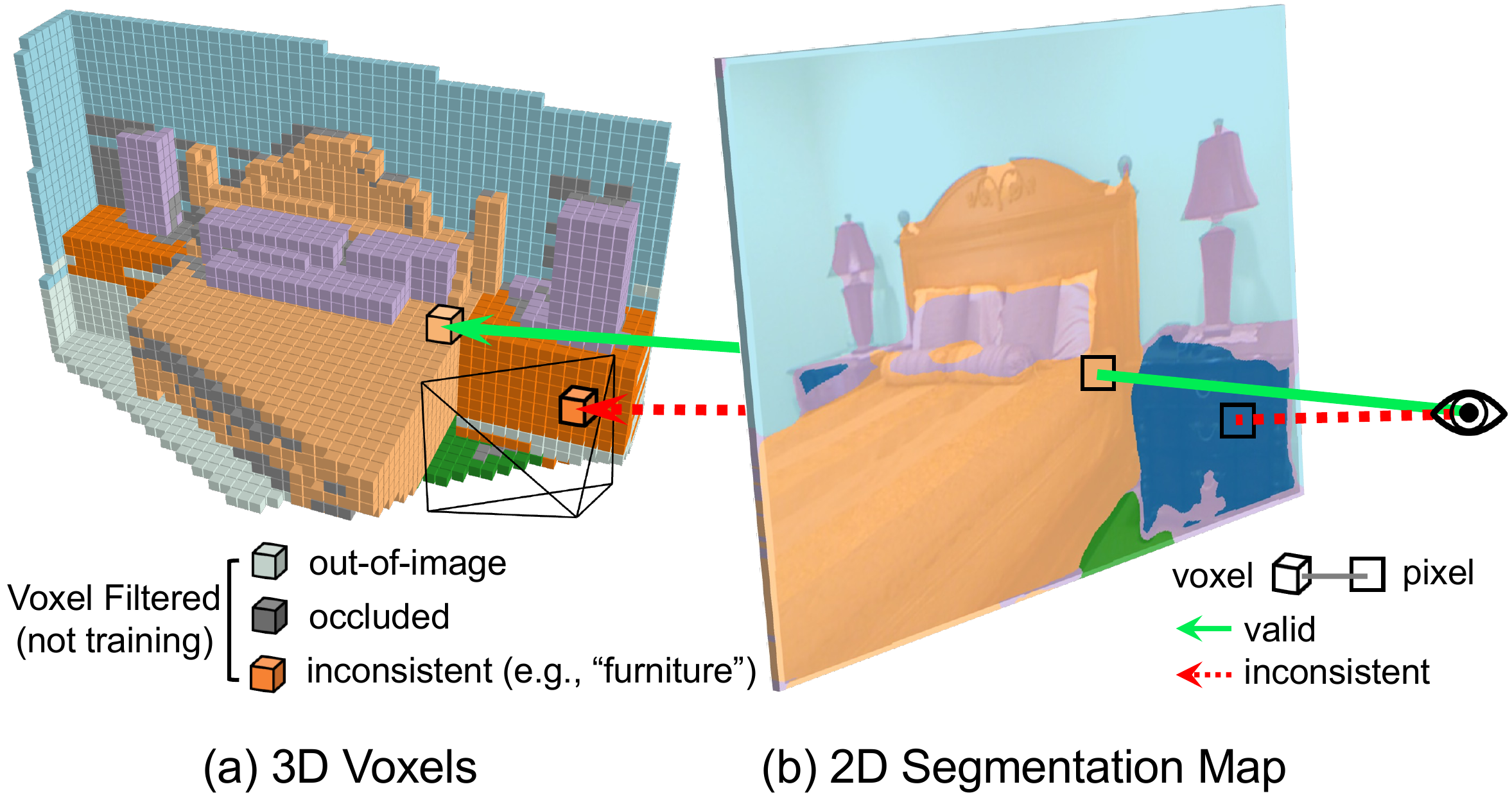} 
    \centering
    \tiny{
    \raisebox{0.5ex}{\colorbox{floor}{}}~floor \raisebox{0.5ex}{\colorbox{wall}{}}~wall \raisebox{0.5ex}{\colorbox{bed}{}}~bed  \raisebox{0.5ex}{\colorbox{table}{}}~table \raisebox{0.5ex}{\colorbox{furniture}{}}~furniture \raisebox{0.5ex}{\colorbox{other}{}}~other \raisebox{0.5ex}{\colorbox{unknown}{}}~occluded \raisebox{0.5ex}{\colorbox{out}{}}~out-of-image}
  \end{minipage}
  \vspace{2mm}
  \caption{\small \textbf{Valid voxel selection for OVO training} (Section~\ref{ss_selection}). We project (a) each 3D voxel to (b) the 2D segmentation map and consider three criteria to filter out invalid voxels: (1) geometrically out-of-image, (2) physically occluded, and (3) label inconsistent between voxel-pixel mapping. For (3) in this example, the ``{\color{furniture}{\emph{furniture}}}'' voxels are removed in the OVO training due to the incorrect prediction to ``{\color{table}{\emph{table}}}'' in pixels by the 2D segmentation model. {\color{green}{Green}} solid ({\color{red}{red}} dashed) arrow indicates valid (inconsistent) voxel-pixel mapping. 
  }
  \label{fig:pairs}
\end{SCfigure}

%% file: section/exp.tex
\vspace{-1mm}
\section{Experiments} \label{s_exp}
\vspace{-1mm}

We validate the proposed Open-Vocabulary Occupancy (OVO) framework in both indoor and outdoor scenarios. We first introduce the experimental setup in Section~\ref{subsec:setting}, then present the main results in Section~\ref{subsec:sota}, and finally show detailed ablation studies and analyses in Section~\ref{subsec:ablation}. Please be referred to \app for additional details and results.

\vspace{-1mm}
\subsection{Setup} \label{subsec:setting}
\mypara{Datasets.}
We conduct experiments on two popular benchmarks for semantic occupancy prediction: \textbf{(1) NYUv2}~\cite{nyu} consists of 1449 indoor scenes. We follow the split in~\cite{cao2022monoscene} and use 795 scenes for training and 654 scenes for inference at scale 1:4. NYUv2 has 12 classes as in~\cite{cao2022monoscene}. We change the label ``object'' to ``other'' for better comparison. \textbf{(2) SemanticKITTI}~\cite{semantickitti} is a large-scale dataset for outdoor scene understanding. It provides both LiDAR sweeps and RGB images. In our experiments, we only use RGB images as the input and generate $256 \times 256\times 32$ occupancy scenes. Following \cite{cao2022monoscene}, we divide train/val split into 3834/815 scenes. We use sequence 08 for inference.

\mypara{Experimental Protocols.} 
\textbf{(a) Benchmark.} We follow \cite{cao2022monoscene} to process the original class labels. In addition, we conduct zero-shot experiments by selection base and novel class labels. In NYUv2~\cite{nyu}, we choose ``bed'', ``table'', and ``other'' as novel classes. In SemanticKITTI, we choose “car”, “road”, and “building” as novel classes.
\textbf{(b) Evaluation Metric.} The main focus of OVO is to determine the semantic class of voxels in an open vocabulary manner. Therefore, we use the mIoU metric to evaluate the performance of semantic scene completion.

\mypara{Baseline Methods.}
To our knowledge, our method is the first framework that allows open-vocabulary semantic occupancy prediction. We thus mainly consider supervised methods as baselines, including AICNet~\cite{AICNet}, 3DSketch~\cite{3DSketch}, MonoScene~\cite{cao2022monoscene}, SSCNet~\cite{sscnet} and TPVFormer~\cite{tpvformer}. The quantitive results of these baselines are taken from previous papers~\cite{cao2022monoscene,tpvformer,sscnet}.

\mypara{Implementation Details.} 
We kept the parameters consistent with MonoScene, except for changing the weight decay from $1 \times 10^{-4}$ to $1 \times 10^{-3}$. The design of the 2D distillation network involves downsampling the original image feature by factors of 2, 4, 8, and 16, followed by convolution operations. The resulting features are then concatenated and passed through two additional convolution layers to align the features to a dimension of 512.
The design of the 3D distillation network varies for different datasets. For the NYU dataset, the 3D distillation network consists of three convolution layers. On the other hand, for the SemanticKITTI dataset, the 3D distillation network is designed with five layers of MLP. The reason for the different structures of the 3D distillation network is that SemanticKITTI has a significantly larger number of voxels compared to NYUv2. Using the original 3D convolutional network would result in excessive GPU memory usage. Therefore, before distillation, we filter out the valid voxels and flatten them. These flattened voxel features are then inputted into an MLP to align the features to a dimension of 512.
We train OVO on the NYUv2 dataset for 100 epochs using 8 V100 GPUs, which took approximately 11.5 hours. Additionally, we train OVO on the SemanticKITTI dataset for 30 epochs, which took approximately 49.5 hours.

\subsection{Main Results} \label{subsec:sota}
\input{table/nyu_main}
\input{table/kitti_main}

\mypara{Results on NYUv2}~\cite{nyu}.
We first evaluate the performance of OVO on indoor scenes using the NYUv2 dataset and show the results in Table~\ref{tab:sota_nyu}. We can see our approach achieves competitive results in novel classes. It is worth noting that \ourmethod outperforms some supervised methods such as AICNet~\cite{AICNet} and SSCNet~\cite{sscnet} for the novel classes, even though these methods use depth information for prediction.
We observe a minor performance decrease in the base class. To gain further understanding, we conduct an experiment following the setup of the original MonoScene but merge novel classes only during the preprocessing stage. This modified version is denoted as MonoScene$\star$ in Table~\ref{tab:sota_nyu}. The results show that MonoScene$\star$ is worse than the fully-supervised MonoScene, suggesting that the performance drop from base classes in OVO may not be due to our model design. We hypothesize the label merging process degrades the performance.

\mypara{Results on SemanticKITTI}~\cite{semantickitti}.
Next, we validate OVO in SemanticKITTI dataset and show results in Table~\ref{tab:sota_kitti}. Again, the experimental results demonstrate that the OVO method surpasses those of AICNet and 3DSketch on novel class, while achieving comparable results on the base class, justifying the robustness and generalizability of our approach in different scenarios.

Overall, the results obtained from both the NYUv2 and SemanticKITTI datasets provide strong evidence of the superior performance of our OVO method, particularly for novel classes, while maintaining competitive performance on the base class. These findings underscore the potential of our approach in advancing semantic occupancy prediction tasks in various real-world scenarios.

\subsection{Ablation and Analyses} \label{subsec:ablation}
In this section, we first investigate the effectiveness of the three proposed feature alignment modules. Then, we examine the impact of re-weighting on the results. Next, we explore the influence of training voxel quality on the experimental results, demonstrating the effectiveness of the voxel filter. Finally, we evaluate the computational efficiency and present the visualization results. \emph{Please refer to the \app for details regarding the selection of novel classes and experiments related to the quantity of training data.} All the ablation studies were conducted on the NYUv2 dataset.

\mypara{Feature Alignment.} 
We conducted a series of ablation experiments to validate the effectiveness of each knowledge distillation module. Based on the voxel-pixel alignment, we sequentially incorporate voxel-text alignment and 2D alignment to compare the mIoU values of the novel classes under the same experimental conditions. The experimental results are presented in Table~\ref{table:align_ablation}. The results of the experiments indicate that all the designed modules are effective in improving performance.

\input{table/ablation_ovo}

\mypara{Impact of Re-weighting.}
As mentioned in Equation~\ref{eq:rw}, we apply re-weighting for voxel-pixel feature alignment. The last two rows in Table~\ref{table:align_ablation} show the effectiveness of re-weighting. The results demonstrate that incorporating re-weighting during the training process can significantly improve the performance of OVO. This indicates that noisy labels harm the feature alignment while assigning higher loss weight to pixels with higher confidence given by LSeg~\cite{li2022language} would benefit model training.

\mypara{Impact of Training Voxel Quality.}
We apply different combinations of voxel filters to filter the voxels and investigate the impact of voxel quality on the performance of OVO. 
We analyze the effect of the valid voxel selection module and show results in Table~\ref{table:voxel_select}. We gradually apply the three voxel filtering mechanisms to the original voxel set.  
We can observe that although the number of voxels involved in the training decreases, the performance actually improves, suggesting that the quality of voxels is more critical than the quantity for semantic occupancy prediction. 

\input{table/selection_ovo}

\setlength{\intextsep}{0pt}
\begin{wrapfigure}{r}{0.35\textwidth}
  \centering
  \includegraphics[width=4.5cm]{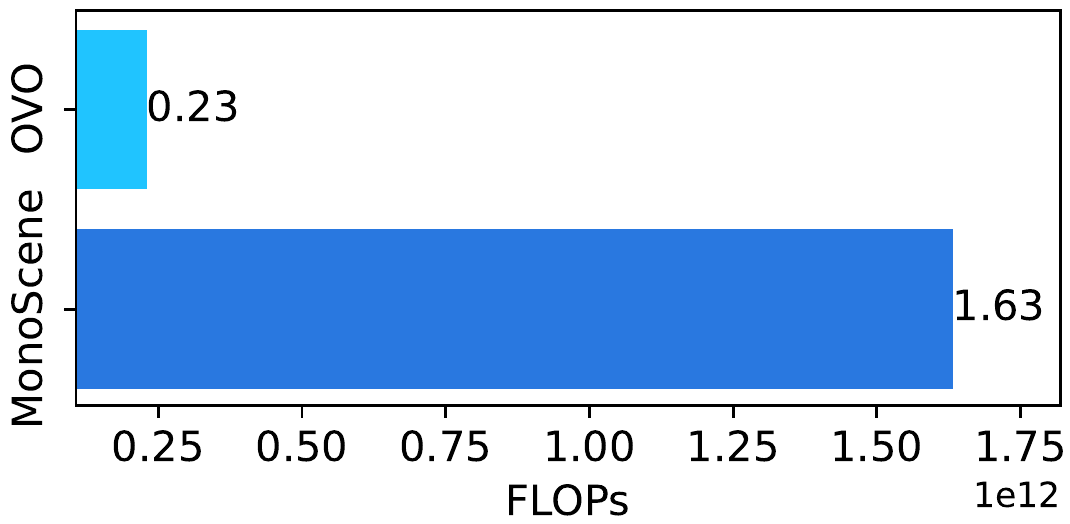} \vspace{-2mm}
  \caption{\small \textbf{Computational efficiency.} \ourmethod leads to only 14\% additional computational cost compared to the original MonoScene~\cite{cao2022monoscene}.}
  \label{fig:flops}
  \vspace{-3mm}
\end{wrapfigure}

\mypara{Efficiency.}
Our OVO framework consists of two components: the original occupancy network (\ie, MonoScene in our experiment) and the newly-added OVO modules (see Figure~\ref{fig:framework}). We show the FLOPs in Figure~\ref{fig:flops}. The proposed OVO modules slightly increase the computational load by 14\%.

\mypara{Qualitative Results.}
Finally, we show qualitative results on the NYUv2 dataset and the SemanticKITTI dataset in Figure~\ref{fig:nyu_qualitative} and Figure~\ref{fig:kitti_qualitative}, respectively. OVO is able to recognize object categories that have not been annotated during the training.

\input{table/figure_qual_nyu}
\input{table/figure_qual_kitti}

%% file: table/nyu_main.tex
\begin{table}[t]
\caption{\small \textbf{Performance (mIoU$\uparrow$) on NYUv2}~\cite{nyu}. \textbf{C}: camera; \textbf{D}: depth; $\dagger$: TSDF}
\label{tab:sota_nyu}
\vspace{1mm}
\footnotesize
\centering
\tabcolsep 1.4pt
\begin{tabular}{l|c|cccc|ccccccccc}
\toprule
 & & \multicolumn{4}{c|}{{\textbf{(a) Novel Class}}}  & \multicolumn{9}{c}{\textbf{(b) Base Class}}\\
\multicolumn{1}{l|}{Method} & Input & \rotatebox{90}{\raisebox{0.5ex}{\colorbox{bed}{}}~{bed}} & \rotatebox{90}{\raisebox{0.5ex}{\colorbox{table}{}}~{table}} & \rotatebox{90}{\raisebox{0.5ex}{\colorbox{other}{}}~{other}} & {mean} & \rotatebox{90}{\raisebox{0.5ex}{\colorbox{ceiling}{}}~{ceiling}} & \rotatebox{90}{\raisebox{0.5ex}{\colorbox{floor}{}}~{floor}} & \rotatebox{90}{\raisebox{0.5ex}{\colorbox{wall}{}}~{wall}} & \rotatebox{90}{\raisebox{0.5ex}{\colorbox{window}{}}~{window}} & \rotatebox{90}{\raisebox{0.5ex}{\colorbox{chair}{}}~{chair}} & \rotatebox{90}{\raisebox{0.5ex}{\colorbox{sofa}{}}~{sofa}} & \rotatebox{90}{\raisebox{0.5ex}{\colorbox{tvs}{}}~{tvs}} & \rotatebox{90}{\raisebox{0.5ex}{\colorbox{furniture}{}}~{furniture}} & {mean} \\ 
\midrule
\multicolumn{14}{l}{\emph{\textbf{Fully-supervised}}}\\
\rowcolor{supervised} (1) AICNet~\cite{AICNet}   & C, D        & 35.87        & 11.11          & 6.45           & 17.81                      & 7.58             & 82.97          & 9.15          & 0.05            & 6.93           & 22.92         & 0.71         & 15.90              & 18.28                  \\
\rowcolor{supervised}(2) SSCNet~\cite{sscnet}   &C, D   & 32.10        & 13.00          & 10.10           & 18.40                         & 15.10             & 94.70          & 24.40          & 0.00            & 12.60          & 35.0         & 7.80         & 27.10              & 27.10                  \\
\rowcolor{supervised}(3) 3DSketch~\cite{3DSketch}   &C$\dagger$   & 42.29        & 13.88          & 8.19           & 21.45                         & 8.53             & 90.45          & 9.94          & 5.67            & 10.64          & 29.21         & 9.38         & 23.83              & 23.46                  \\
\rowcolor{supervised}(4) MonoScene~\cite{cao2022monoscene}   & C      & 48.19        & 15.13          & 12.94          & 25.42                         & 8.89             & 93.50          & 12.06         & 12.57           & 13.72          & 36.11         & 15.22        & 27.96              & 27.50              \\
\midrule
\multicolumn{14}{l}{\emph{\textbf{Zero-shot}}}  \\
(5) MonoScene$\star$  & C        & --        & --           & --           & --                      & 8.10             & 93.49          & 9.94          & 10.32            & 13.24          & 34.47         & 11.75         & 26.41              & 25.96               \\
(6) \ourmethodbold (ours)   &C        & 41.61        & 10.45           & 8.39           & 20.15                     & 7.77             & 93.16          & 7.77          & 6.95            & 10.01          & 33.83         & 8.22         & 25.64              & 24.17                 \\
\bottomrule
\end{tabular}\vspace{-3mm}
\end{table}

%% file: table/kitti_main.tex
\begin{table}[t]
\caption{\small \textbf{Performance (mIoU$\uparrow$) on SemanticKITTI}~\cite{semantickitti}. \textbf{C}: camera; \textbf{D}: depth; $\dagger$: TSDF} 
\label{tab:sota_kitti}
\vspace{1mm}
\centering
\scriptsize
\tabcolsep 2pt
\begin{tabular}{l|c|cccc|ccccccccccccccccc}
\toprule
& & \multicolumn{4}{c|}{\textbf{(a) Novel Class}} & \multicolumn{17}{c}{\textbf{(b) Base Class}}  \\
Method & Input &\rotatebox{90}{\raisebox{0.5ex}{\colorbox{car}{}}~car}   & \rotatebox{90}{\raisebox{0.5ex}{\colorbox{road}{}}~road}  & \rotatebox{90}{\raisebox{0.5ex}{\colorbox{building}{}}~building} & mean & \rotatebox{90}{\raisebox{0.5ex}{\colorbox{sidewalk}{}}~sidewalk} & \rotatebox{90}{\raisebox{0.5ex}{\colorbox{parking}{}}~parking} & \rotatebox{90}{\raisebox{0.5ex}{\colorbox{otherGround}{}}~other ground} & \rotatebox{90}{\raisebox{0.5ex}{\colorbox{truck}{}}~truck} & \rotatebox{90}{\raisebox{0.5ex}{\colorbox{bicycle}{}}~bicycle} & \rotatebox{90}{\raisebox{0.5ex}{\colorbox{motorcycle}{}}~motorcycle} & \rotatebox{90}{\raisebox{0.5ex}{\colorbox{otherVehicle}{}}~other vehicle} & \rotatebox{90}{\raisebox{0.5ex}{\colorbox{vegetation}{}}~vegetatation} & \rotatebox{90}{\raisebox{0.5ex}{\colorbox{trunk}{}}~trunk} & \rotatebox{90}{\raisebox{0.5ex}{\colorbox{terrain}{}}~terrain} & \rotatebox{90}{\raisebox{0.5ex}{\colorbox{person}{}}~person} & \rotatebox{90}{\raisebox{0.5ex}{\colorbox{bicyclist}{}}~bicyclist} & \rotatebox{90}{\raisebox{0.5ex}{\colorbox{motorcyclist}{}}~motorcyclist} & \rotatebox{90}{\raisebox{0.5ex}{\colorbox{fence}{}}~fence} & \rotatebox{90}{\raisebox{0.5ex}{\colorbox{pole}{}}~pole}  & \rotatebox{90}{\raisebox{0.5ex}{\colorbox{trafficSign}{}}~traffic sign} & mean   \\ \midrule
\multicolumn{22}{l}{\emph{\textbf{Fully-supervised}}}\\
\rowcolor{supervised}(1) AICNet~\cite{AICNet}  &C, D     & 15.3 & 39.3 & 9.6     & 21.4       & 18.3    & 19.8   & 1.6         & 0.7  & 0.0    & 0.0       & 0.0          & 9.6       & 1.9  & 13.5   & 0.0   & 0.0      & 0.0         & 5.0  & 0.1  & 0.0         & 4.4   \\
\rowcolor{supervised}(2) 3DSketch~\cite{3DSketch} &C$\dagger$   & 17.1 & 37.7 & 12.1    & 22.3        & 19.8    & 0.0    & 0.0         & 0.0  & 0.0    & 0.0       & 0.0          & 12.1      & 0.0  & 16.1   & 0.0   & 0.0      & 0.0         & 3.4  & 0.0  & 0.0         &  3.2  \\
\rowcolor{supervised}(3) MonoScene~\cite{cao2022monoscene} &C  & 18.8 & 54.7 & 14.4    & 29.3        & 27.1    & 24.8   & 5.7         & 3.3  & 0.5    & 0.7       & 4.4          & 14.9      & 2.4  & 19.5   & 1.0   & 1.4      & 0.4         & 11.1 & 3.3  & 2.1         & 7.7   \\ 
\rowcolor{supervised}(4) TPVFormer~\cite{tpvformer}   & C$\times$6      & 23.8        & 56.5          & 13.9          & 31.4                         & 25.9             & 20.6          & 0.9         & 8.1           & 0.4          & 0.1         & 4.4        & 16.9              & 2.3 & 30.4 & 0.5 & 0.9 & 0.0 & 5.9 & 3.1 & 1.5  &7.6            \\
\midrule
\multicolumn{22}{l}{\emph{\textbf{Zero-shot}} } \\
(5) \ourmethodbold (ours) &C     & 13.3 & 53.9 & 9.7     & 25.7       & 26.5    & 14.4   & 0.1         & 0.7  & 0.4    & 0.3       & 2.5          & 17.2      & 2.3  & 29.0   & 0.6   & 0.7      & 0.0         & 5.4  & 3.0  & 1.7         & 6.6  \\

\bottomrule
\end{tabular}\vspace{-3mm}
\end{table}

%% file: table/ablation_ovo.tex
\begin{table}[t]
\small
\tabcolsep 11pt
\centering
\caption{\small \textbf{Ablation study} (NYUv2~\cite{nyu}). \textbf{RW}: re-weighting for voxel-pixel alignment (Section~\ref{ss_align}).}
\label{table:align_ablation}
\vspace{1mm}
\begin{tabular}{ccc|c|cccc} \toprule
\multicolumn{3}{c|}{Feature Alignment} & \multicolumn{1}{c|}{\multirow{2.5}{*}{RW}} & \multicolumn{4}{c}{Novel Class (mIoU$\uparrow$)}  \\ \cmidrule{1-3} \cmidrule{5-8}
Voxel-Pixel & Voxel-Text & Pixel-Pixel & \multicolumn{1}{c|}{} & bed & table & other & mean \\ \midrule 
\cmark & \multicolumn{1}{l}{} & \multicolumn{1}{l|}{} &  & 37.15 & 8.58 & 7.09 & 17.61 \\
\cmark & \cmark & \multicolumn{1}{l|}{} &  & 38.24 & 8.10 & 7.17 & 17.84 \\
\cmark & \cmark & \cmark &  & 38.94 & 8.40 & 7.46 & 18.26 \\
\cmark & \cmark & \cmark & \cmark & \textbf{41.61} & \textbf{10.45} & \textbf{8.39} & \textbf{20.15} \\ \bottomrule
\end{tabular} \vspace{-3mm}
\end{table}

%% file: table/selection_ovo.tex
\begin{table}[t]
\small
\tabcolsep 10pt
\centering
\caption{\small \textbf{Analysis of valid voxel selection} (NYUv2~\cite{nyu}). \# Voxel: number of voxels kept for OVO training.}
\label{table:voxel_select}
\vspace{1mm}
\begin{tabular}{ccc|r|cccc} \toprule
\multicolumn{3}{c|}{Voxel Filtering} & \multicolumn{1}{c|}{\multirow{2.5}{*}{\# Voxel}} &\multicolumn{4}{c}{Novel Class (mIoU$\uparrow$)}  \\ \cmidrule{1-3} \cmidrule{5-8} 
Out-of-Image & Occluded & Inconsistent & & \multicolumn{1}{c}{bed} & \multicolumn{1}{c}{table} & \multicolumn{1}{c}{other} & \multicolumn{1}{c}{mean} \\ \midrule
\cmark &  &  & 2,719K & 30.15 & 5.29 & 7.67 & 14.37 \\
\cmark & \cmark &  &  1,132K & 32.97 & 6.12 & 7.71 & 15.60 \\
\cmark & \cmark & \cmark & 360K & \textbf{41.61} & \textbf{10.45} & \textbf{8.39} & \textbf{20.15} \\ \bottomrule
\end{tabular} \vspace{-3mm}
\end{table}

%% file: table/figure_qual_nyu.tex
\begin{figure}[htbp]
\centering     

\begin{minipage}{0.23\linewidth}
    \vspace{3pt}
    \centerline{\includegraphics[width=\textwidth]{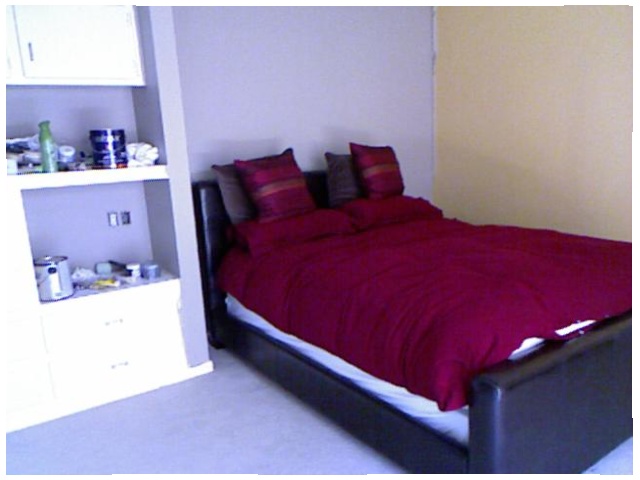}}
\end{minipage}
\begin{minipage}{0.23\linewidth}
    \vspace{3pt}
    \centerline{\includegraphics[width=\textwidth, viewport=500 0 2800 1700,clip]{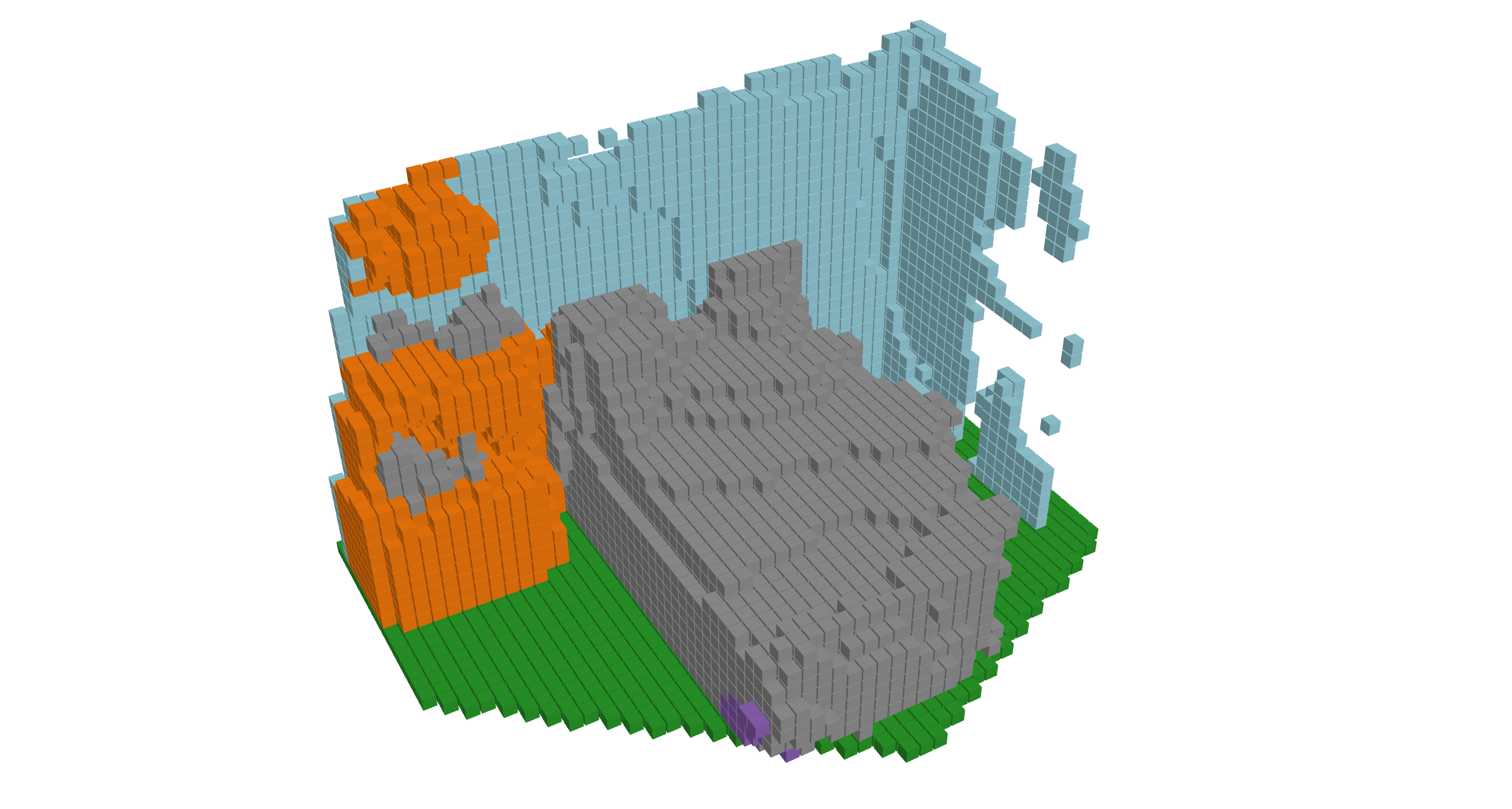}}
\end{minipage}
\begin{minipage}{0.23\linewidth}
    \vspace{3pt}
    \centerline{\includegraphics[width=\textwidth, viewport=500 0 2800 1700,clip]{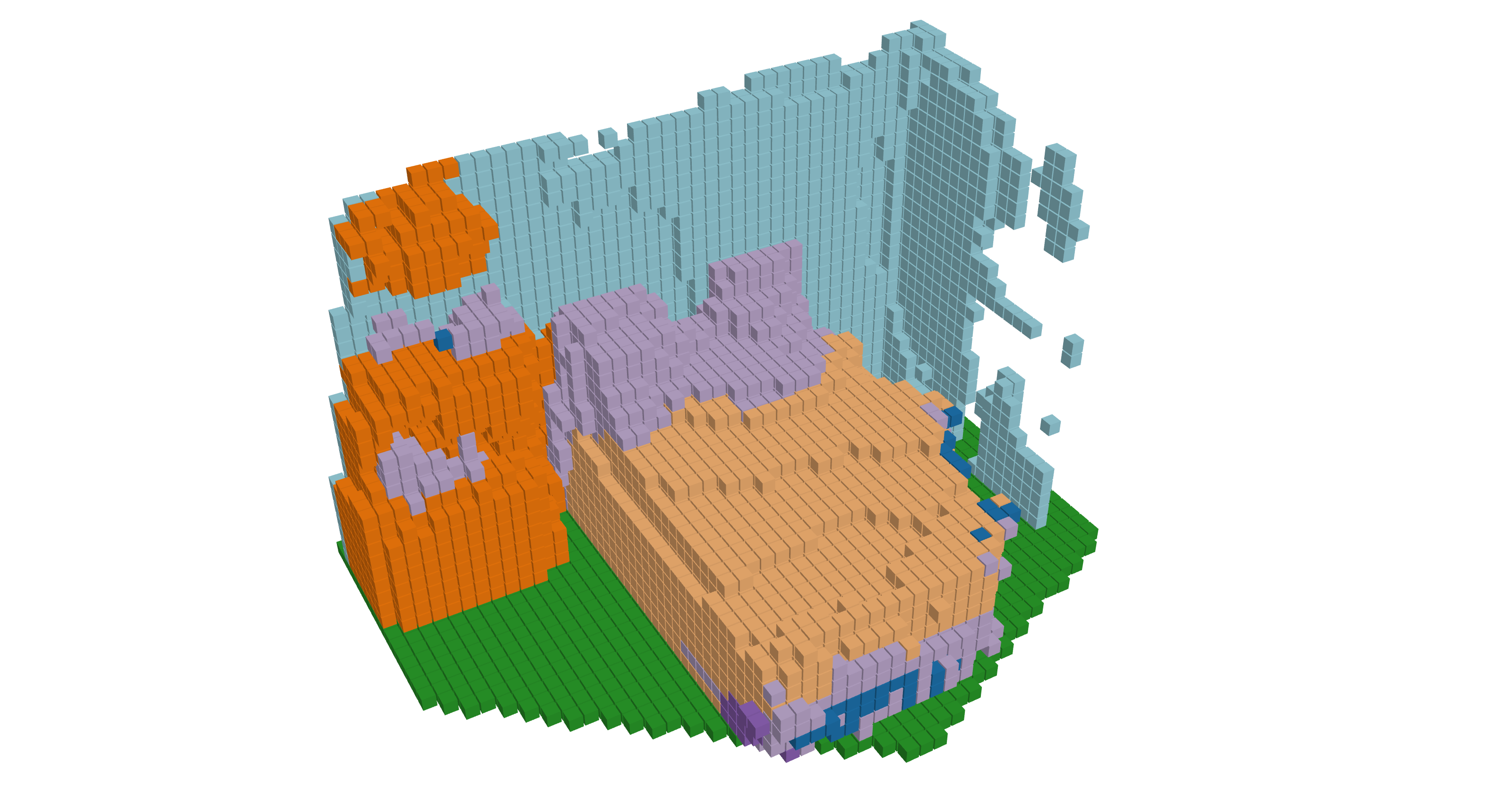}}
\end{minipage}
\begin{minipage}{0.23\linewidth}
    \vspace{3pt}
    \centerline{\includegraphics[width=\textwidth, viewport=500 0 2800 1700,clip]{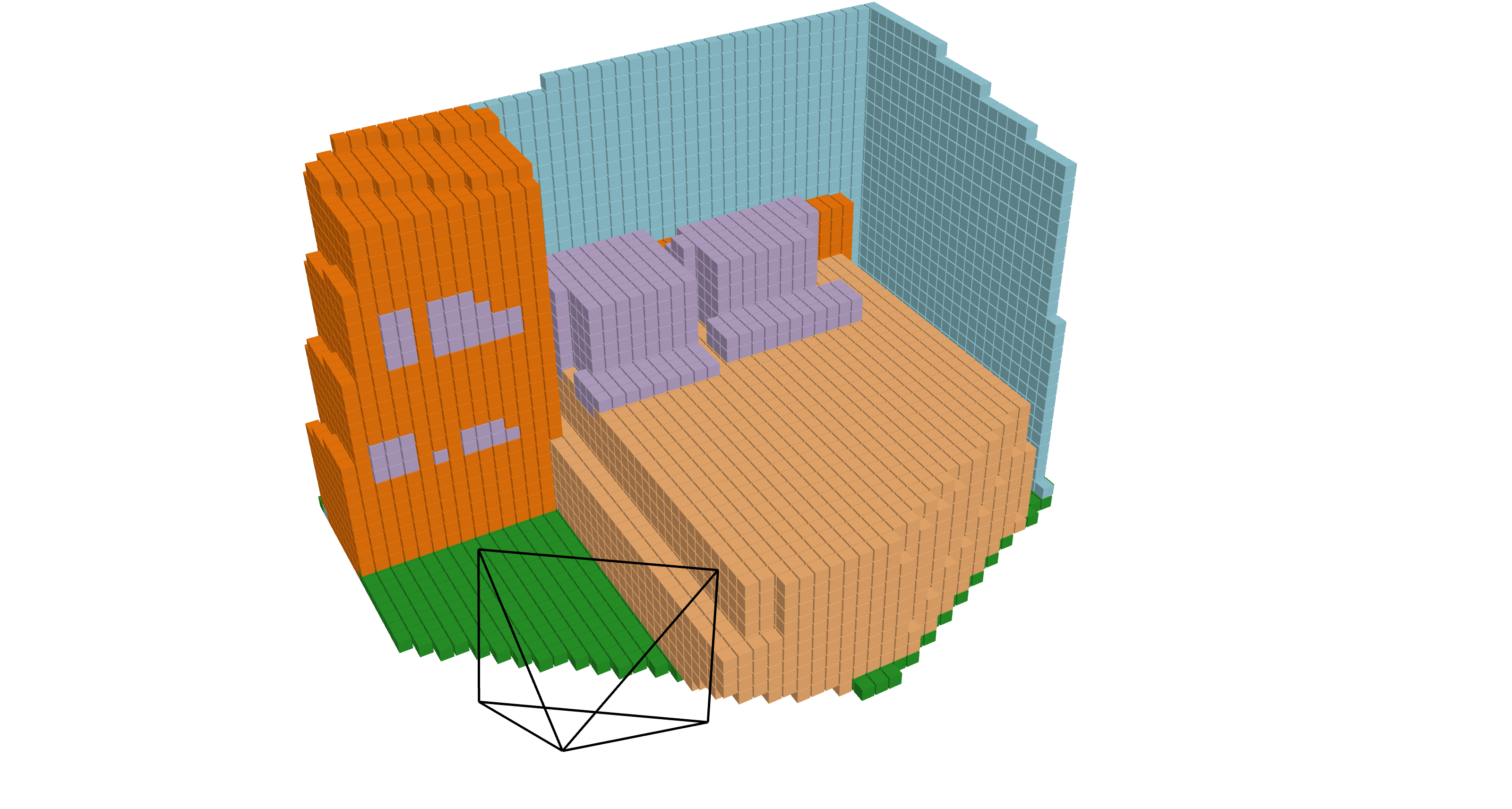}}
\end{minipage}

\begin{minipage}{0.23\linewidth}
    \vspace{3pt}
    \centerline{\includegraphics[width=\textwidth]{figure/nyu_exp/NYU0994_0000_color.jpg}}
\end{minipage}
\begin{minipage}{0.23\linewidth}
    \vspace{3pt}
    \centerline{\includegraphics[width=\textwidth, viewport=500 0 2800 1700,clip]{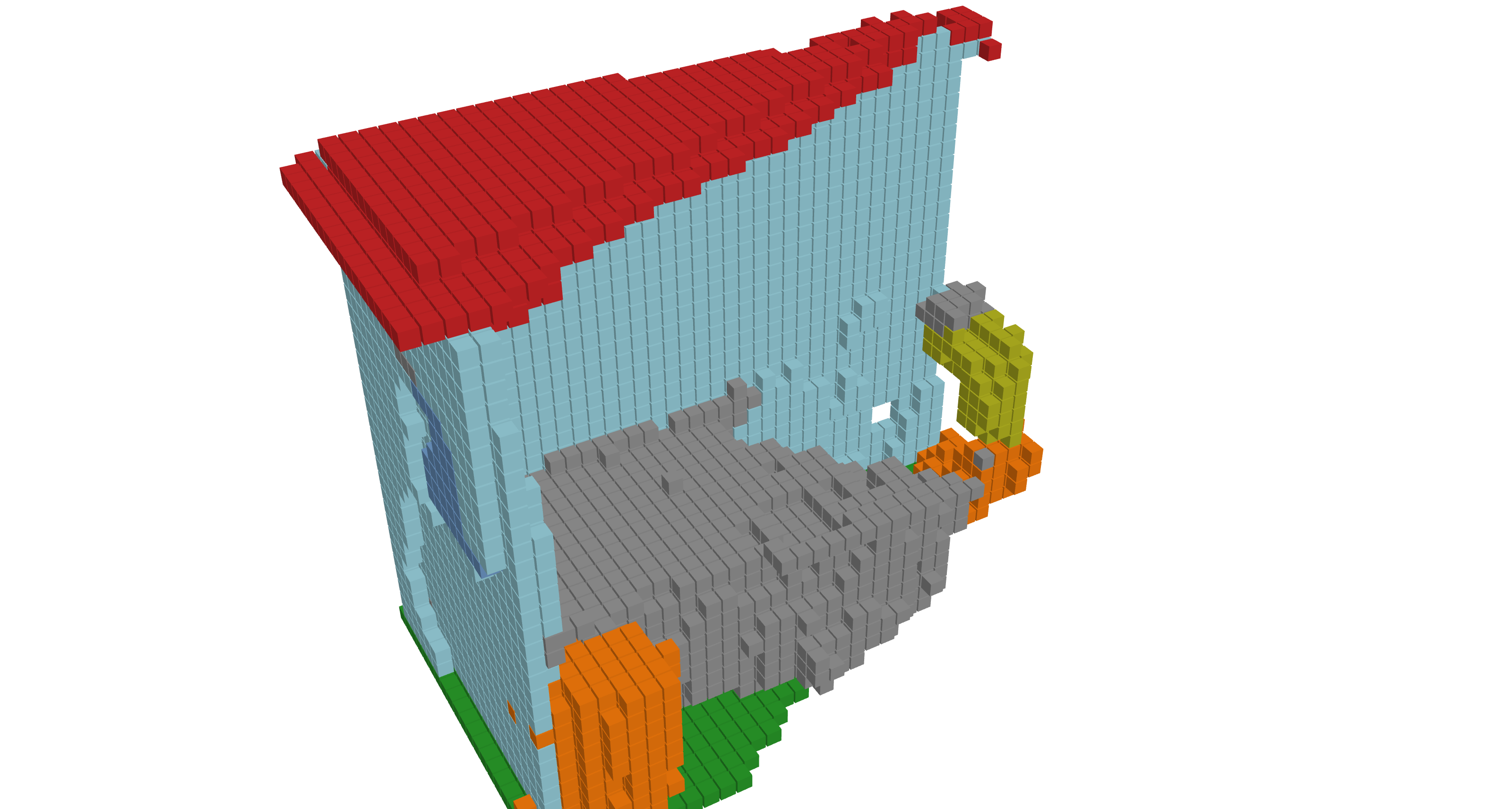}}
\end{minipage}
\begin{minipage}{0.23\linewidth}
    \vspace{3pt}
    \centerline{\includegraphics[width=\textwidth, viewport=500 0 2800 1700,clip]{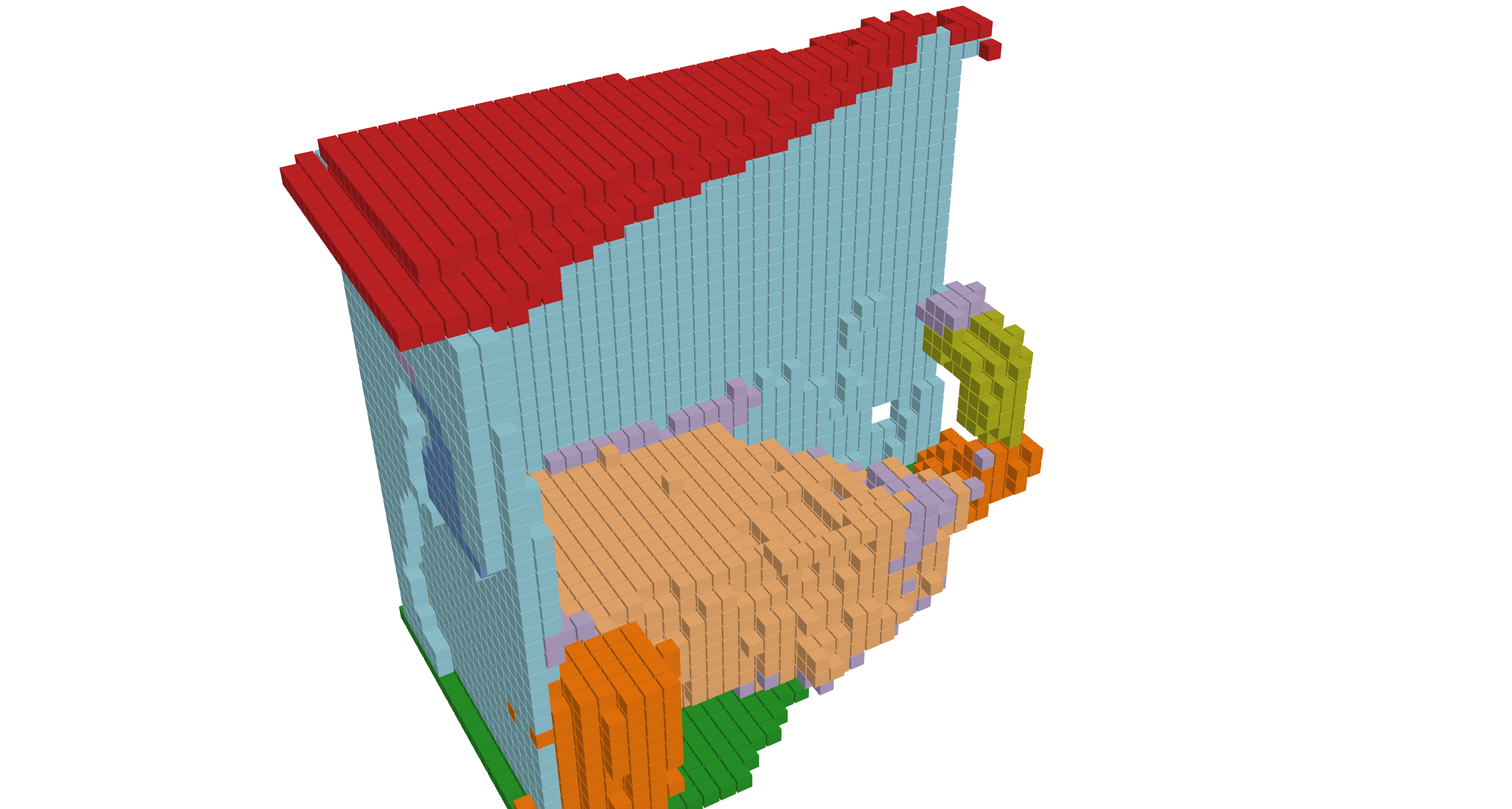}}
\end{minipage}
\begin{minipage}{0.23\linewidth}
    \vspace{3pt}
    \centerline{\includegraphics[width=\textwidth, viewport=500 0 2800 1700,clip]{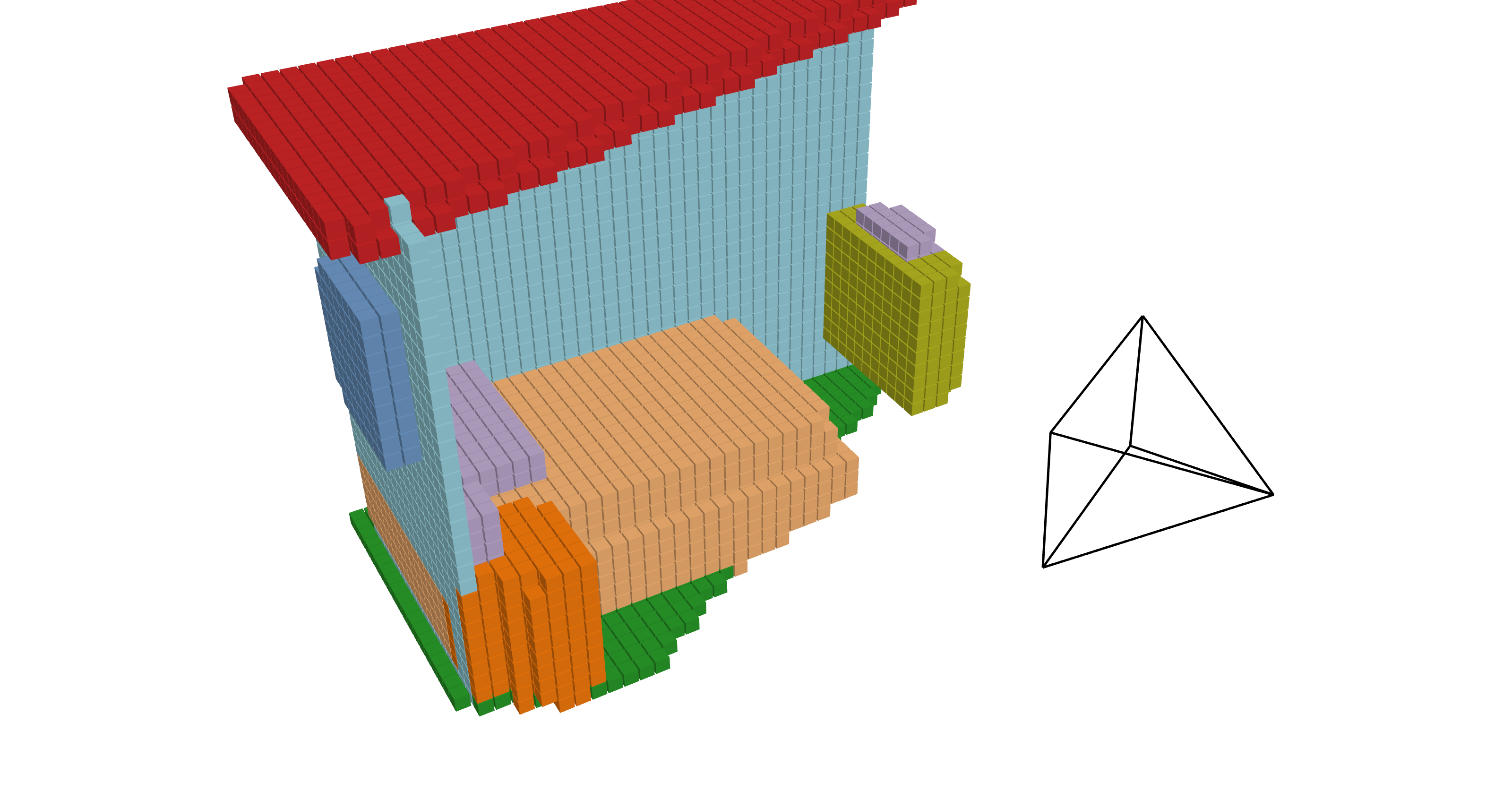}}
\end{minipage}

\begin{minipage}{0.23\linewidth}
    \vspace{3pt}
    \centerline{\includegraphics[width=\textwidth]{figure/nyu_exp/NYU1407_0000_color.jpg}}
\end{minipage}
\begin{minipage}{0.23\linewidth}
    \vspace{3pt}
    \centerline{\includegraphics[width=\textwidth, viewport=800 0 2400 1200,clip]{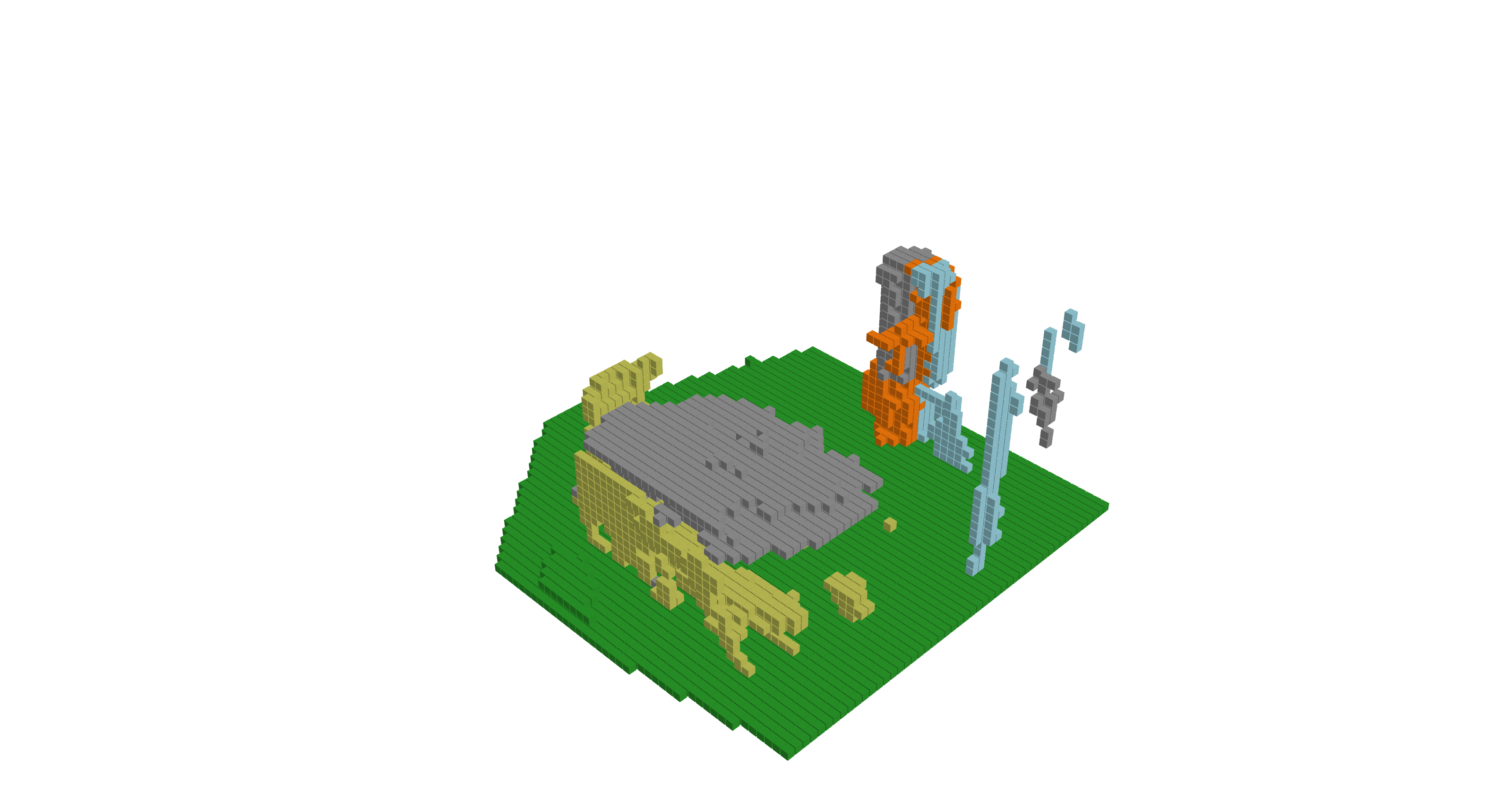}}
\end{minipage}
\begin{minipage}{0.23\linewidth}
    \vspace{3pt}
    \centerline{\includegraphics[width=\textwidth, viewport=800 0 2400 1200,clip]{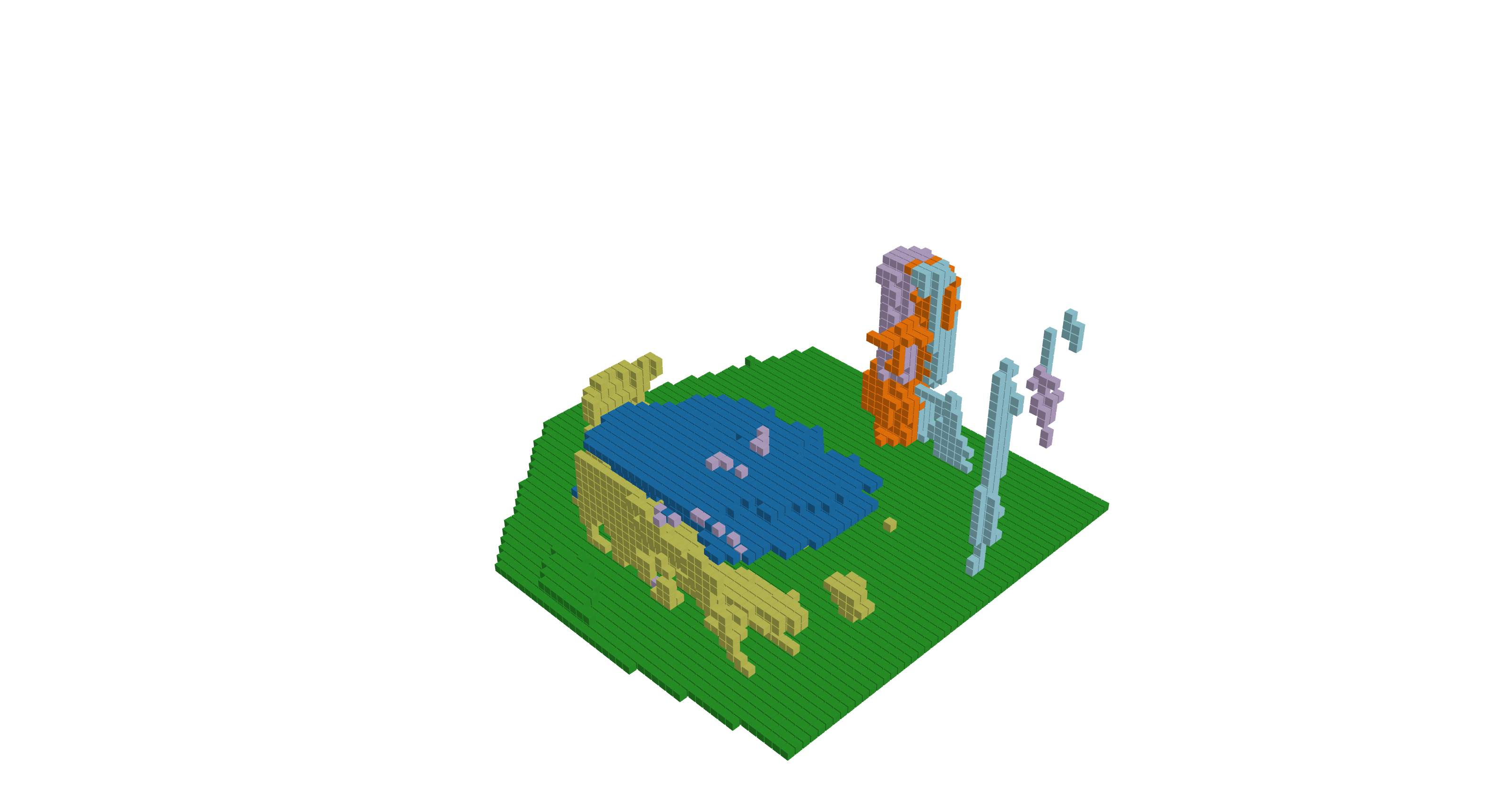}}
\end{minipage}
\begin{minipage}{0.23\linewidth}
    \vspace{3pt}
    \centerline{\includegraphics[width=\textwidth, viewport=800 0 2400 1200,clip]{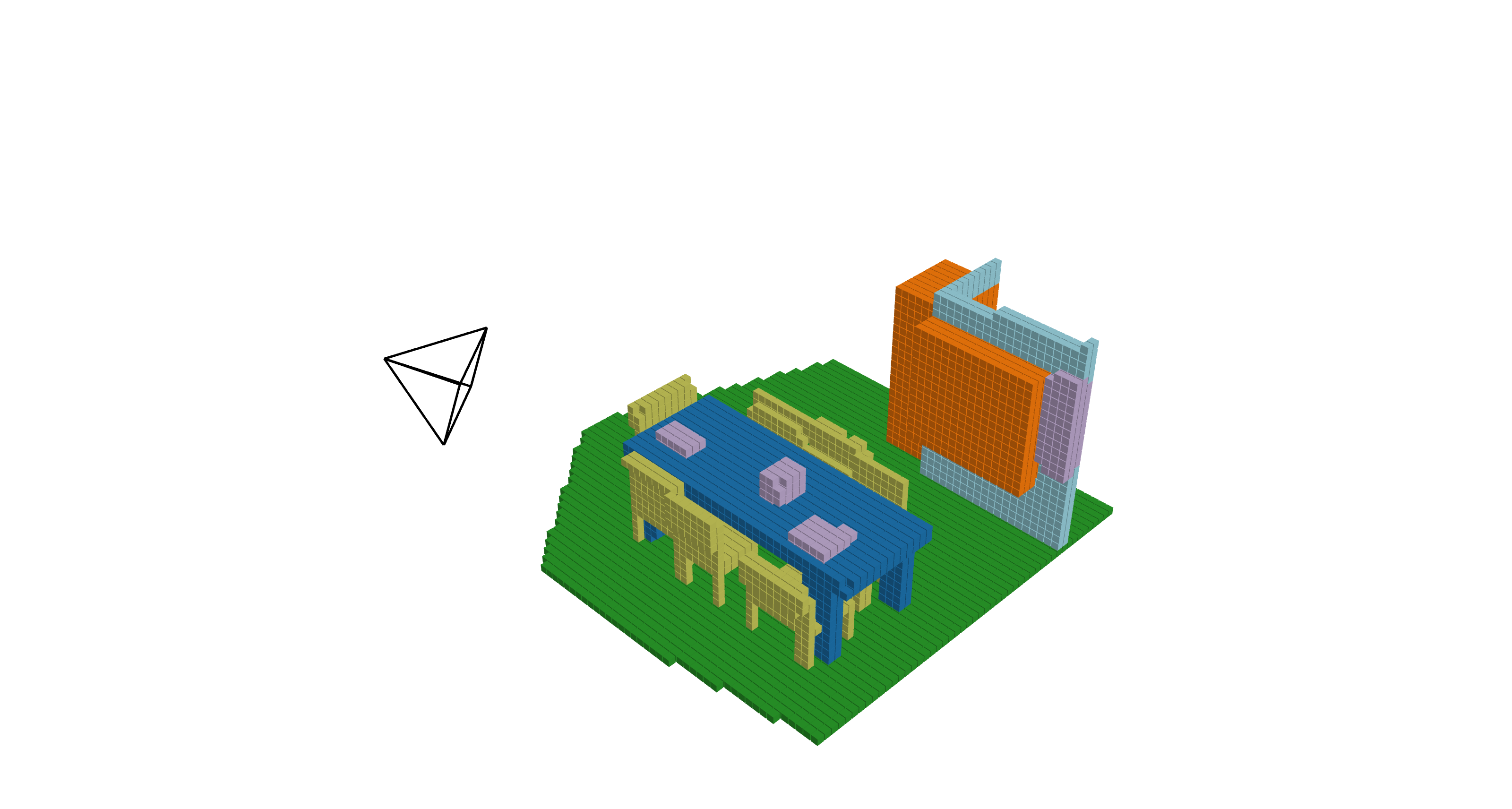}}
\end{minipage}

\begin{minipage}{0.23\linewidth}
    \centering
    RGB input
\end{minipage}
\begin{minipage}{0.23\linewidth}
    \centering
    MonoScene
\end{minipage}
\begin{minipage}{0.23\linewidth}
    \centering
    OVO (ours)
\end{minipage}
\begin{minipage}{0.23\linewidth}
    \centering
    GT
\end{minipage}

\begin{minipage}{\linewidth}
    \centering
    \raisebox{0.5ex}{\colorbox{ceiling}{}}~ceiling \raisebox{0.5ex}{\colorbox{floor}{}}~floor \raisebox{0.5ex}{\colorbox{wall}{}}~wall \raisebox{0.5ex}{\colorbox{window}{}}~window \raisebox{0.5ex}{\colorbox{chair}{}}~chair \raisebox{0.5ex}{\colorbox{bed}{}}~bed \raisebox{0.5ex}{\colorbox{sofa}{}}~sofa \raisebox{0.5ex}{\colorbox{table}{}}~table \raisebox{0.5ex}{\colorbox{tvs}{}}~tvs \raisebox{0.5ex}{\colorbox{furniture}{}}~furniture \raisebox{0.5ex}{\colorbox{other}{}}~other \raisebox{0.5ex}{\colorbox{unknown}{}}~unknown
\end{minipage}

\caption{\small \textbf{Qualitative visualization on NYUv2 dataset}~\cite{nyu}. The novel classes for NYUv2 dataset include  {\color{bed} “bed”}, {\color{table} “table”}, and {\color{other}“other”}. The gray voxels in the second column represent the instances of these novel classes that cannot be predicted by the vanilla MonoScene trained with supervised data. In the third column, gray voxels are painted according to the inference results of our OVO. We note that our OVO model can accurately predict the novel class even with low-quality input, as shown in the second row. Please see \app for more results.}
\label{fig:nyu_qualitative}
\end{figure}

%% file: table/figure_qual_kitti.tex
\begin{figure*}[htbp]
\centering     

\begin{minipage}{0.23\linewidth}
    \vspace{3pt}
    \centerline{\includegraphics[width=\textwidth]{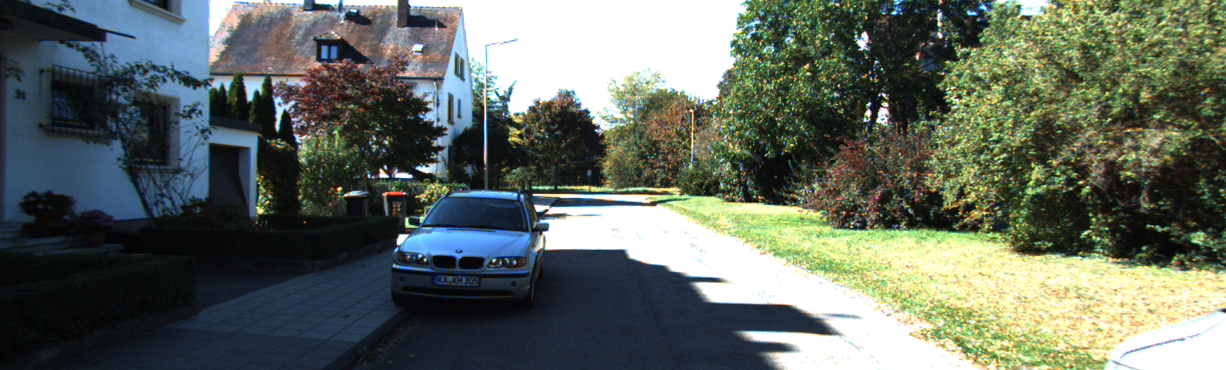}}
\end{minipage}
\begin{minipage}{0.23\linewidth}
    \vspace{3pt}
    \centerline{\includegraphics[width=\textwidth, viewport=200 250 2000 1300,clip]{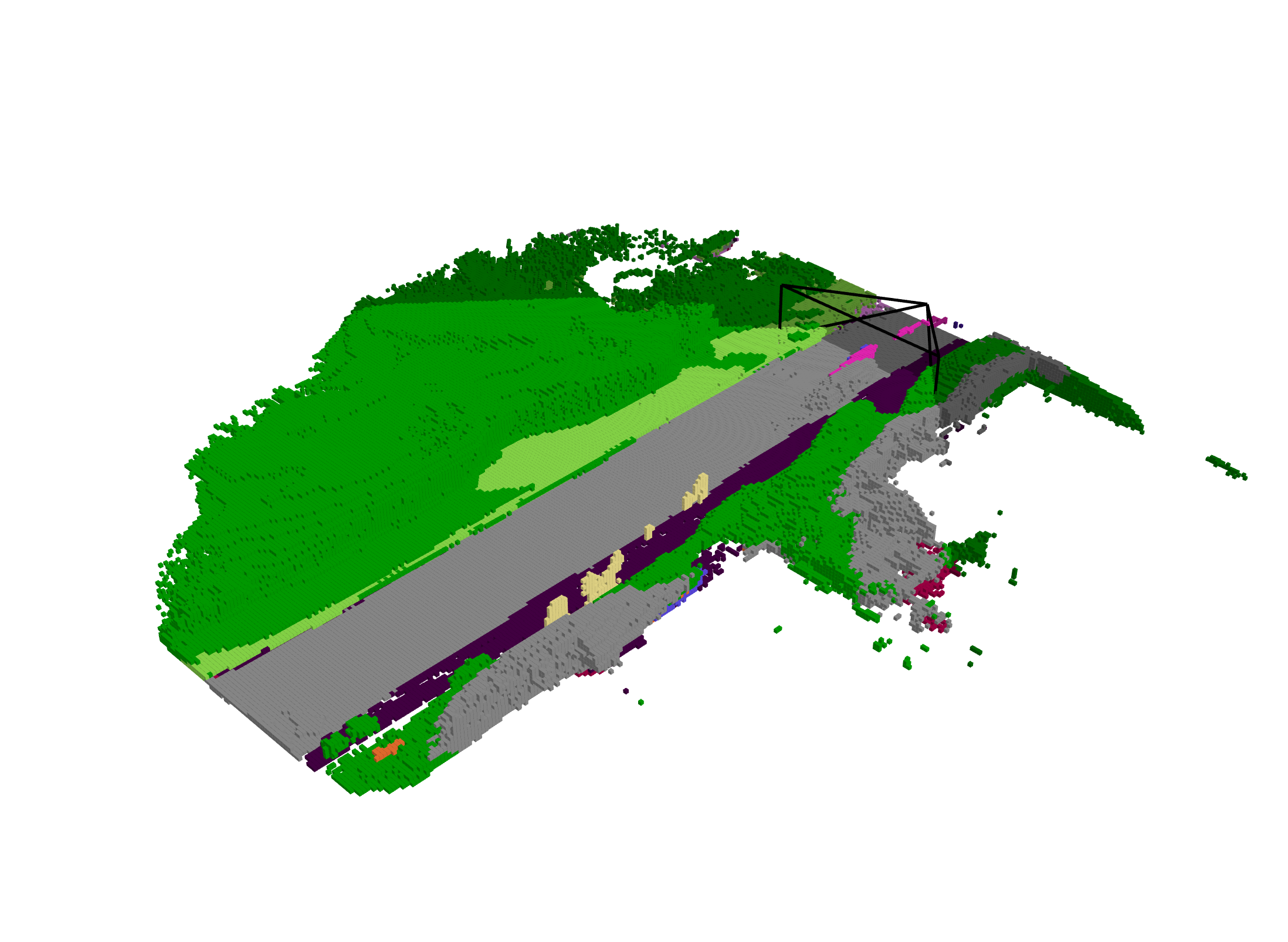}}
\end{minipage}
\begin{minipage}{0.23\linewidth}
    \vspace{3pt}
    \centerline{\includegraphics[width=\textwidth, viewport=200 250 2000 1300,clip]{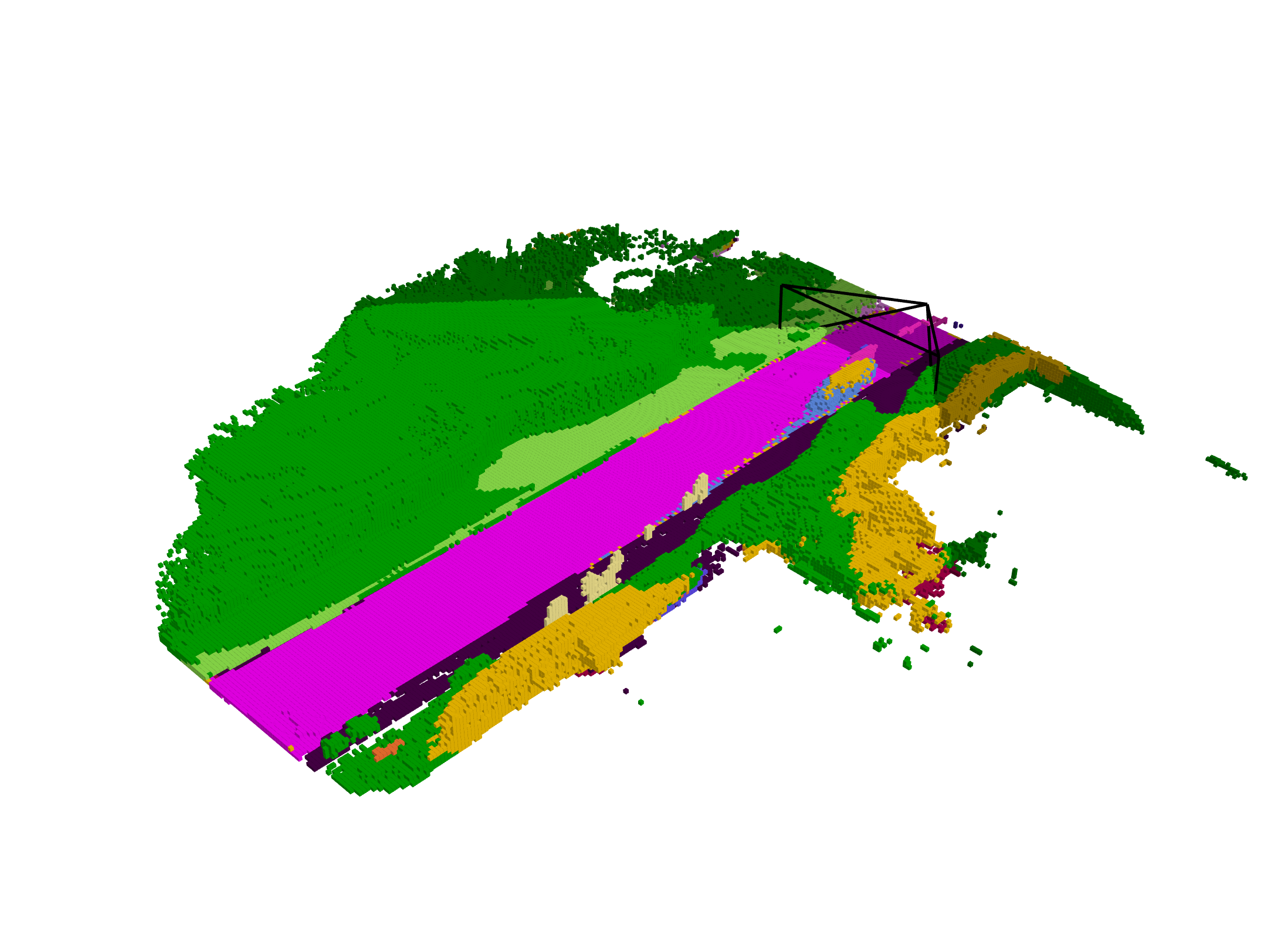}}
\end{minipage}
\begin{minipage}{0.23\linewidth}
    \vspace{3pt}
    \centerline{\includegraphics[width=\textwidth, viewport=200 250 2000 1300,clip]{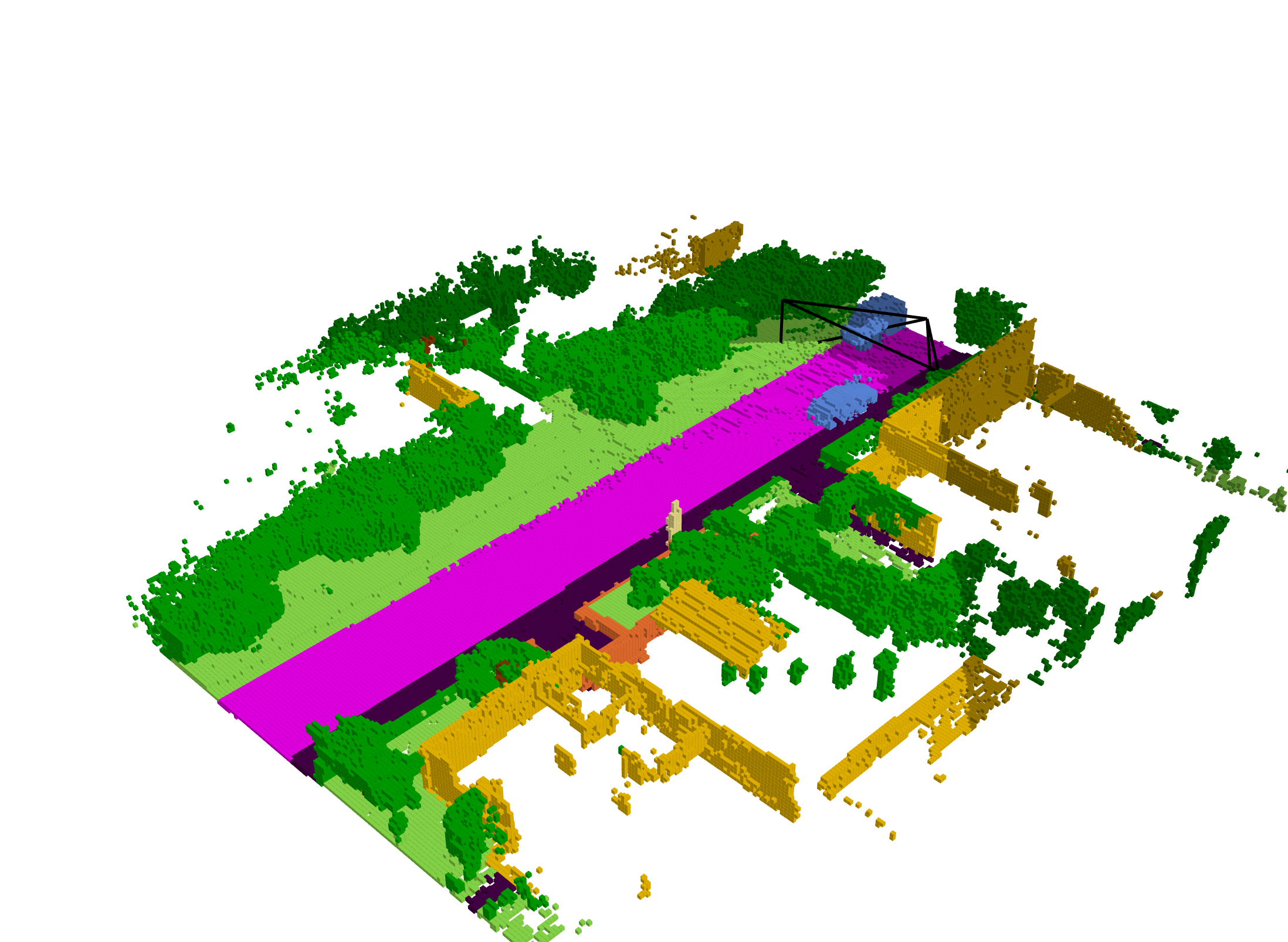}}
\end{minipage}

\begin{minipage}{0.23\linewidth}
    \vspace{3pt}
    \centerline{\includegraphics[width=\textwidth]{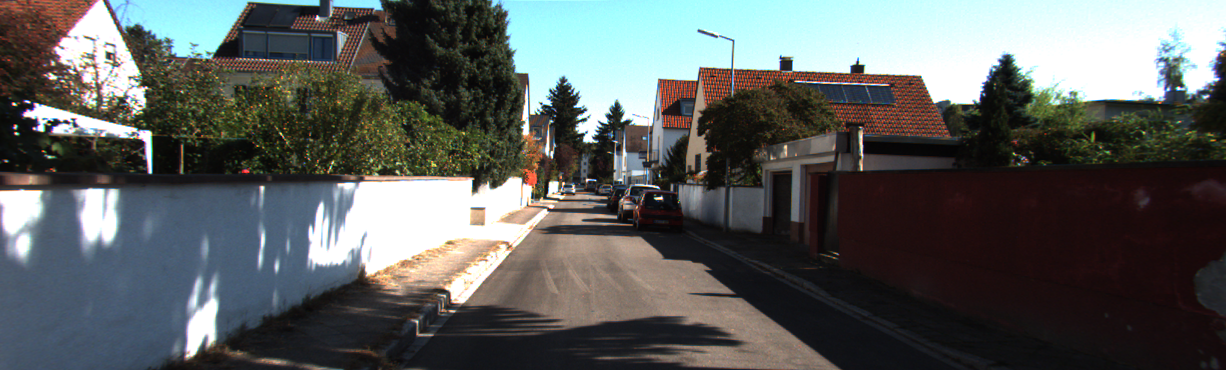}}
\end{minipage}
\begin{minipage}{0.23\linewidth}
    \vspace{3pt}
    \centerline{\includegraphics[width=\textwidth, viewport=200 250 2000 1300,clip]{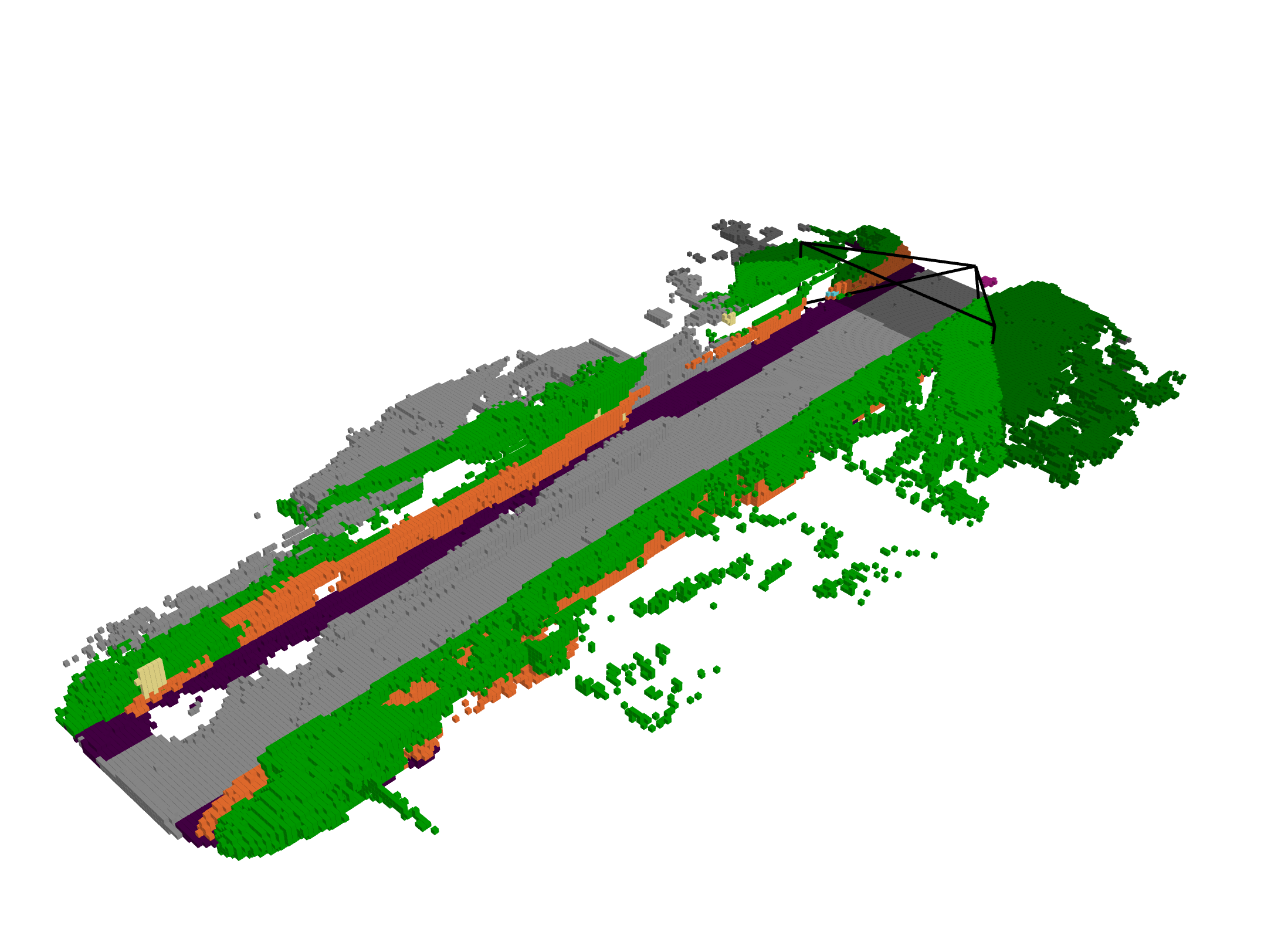}}
\end{minipage}
\begin{minipage}{0.23\linewidth}
    \vspace{3pt}
    \centerline{\includegraphics[width=\textwidth, viewport=200 250 2000 1300,clip]{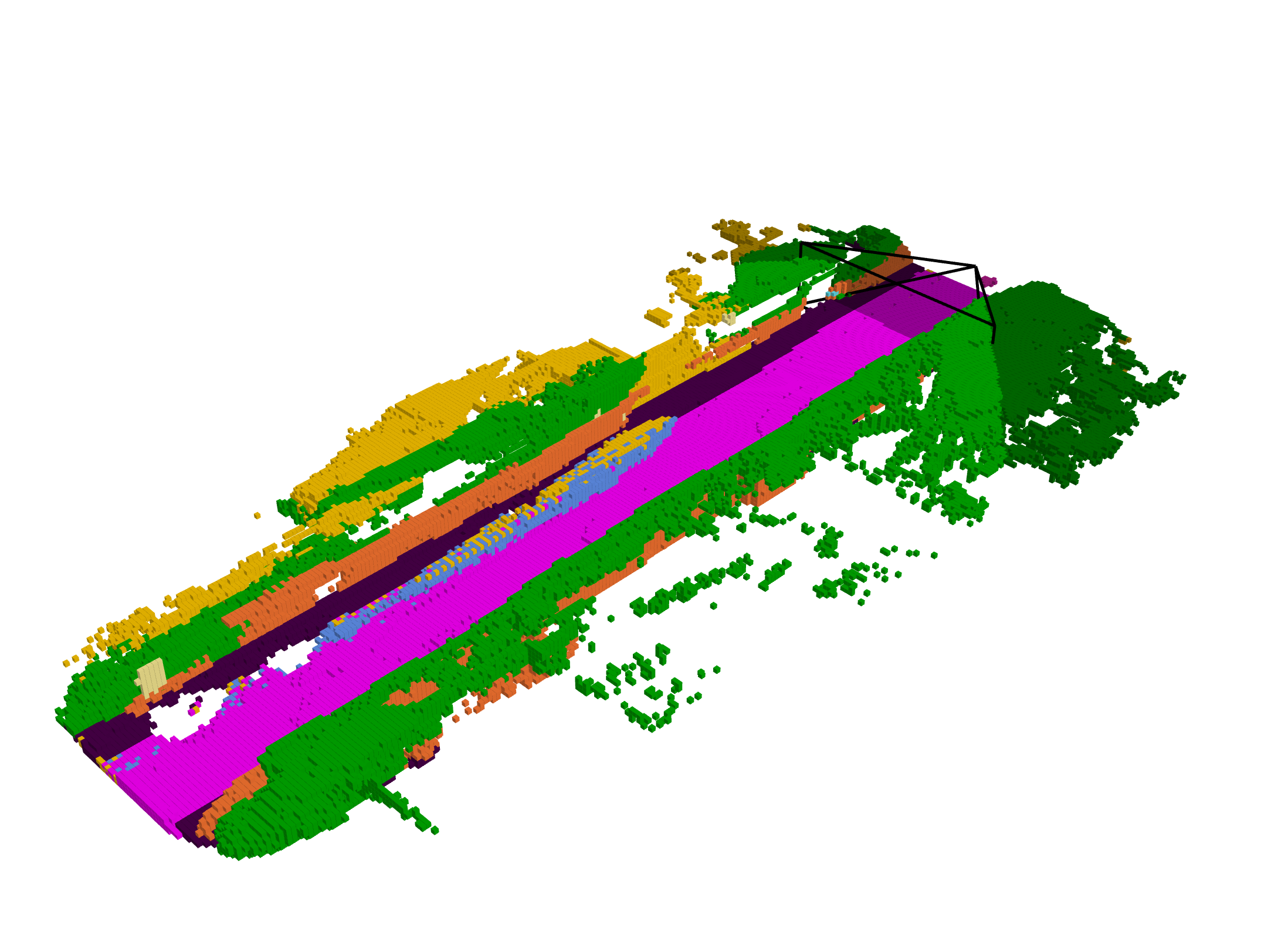}}
\end{minipage}
\begin{minipage}{0.23\linewidth}
    \vspace{3pt}
    \centerline{\includegraphics[width=\textwidth, viewport=200 250 2000 1300,clip]{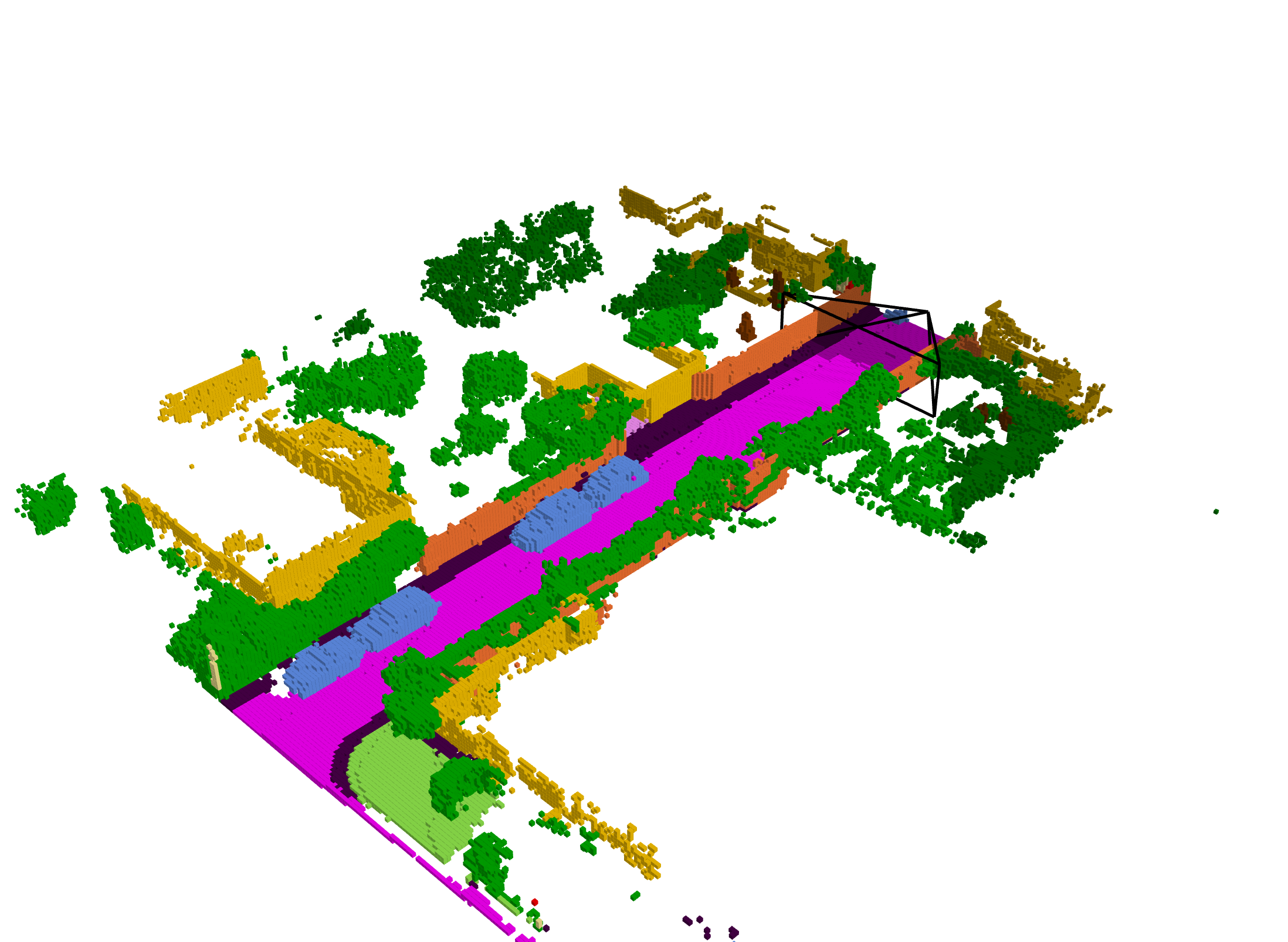}}
\end{minipage}

\begin{minipage}{0.23\linewidth}
    \vspace{3pt}
    \centerline{\includegraphics[width=\textwidth]{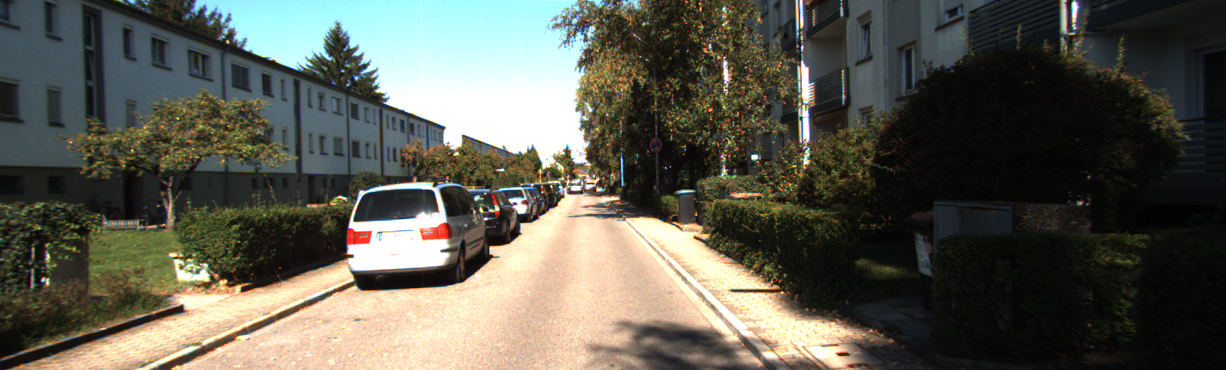}}
\end{minipage}
\begin{minipage}{0.23\linewidth}
    \vspace{3pt}
    \centerline{\includegraphics[width=\textwidth, viewport=200 250 2000 1300,clip]{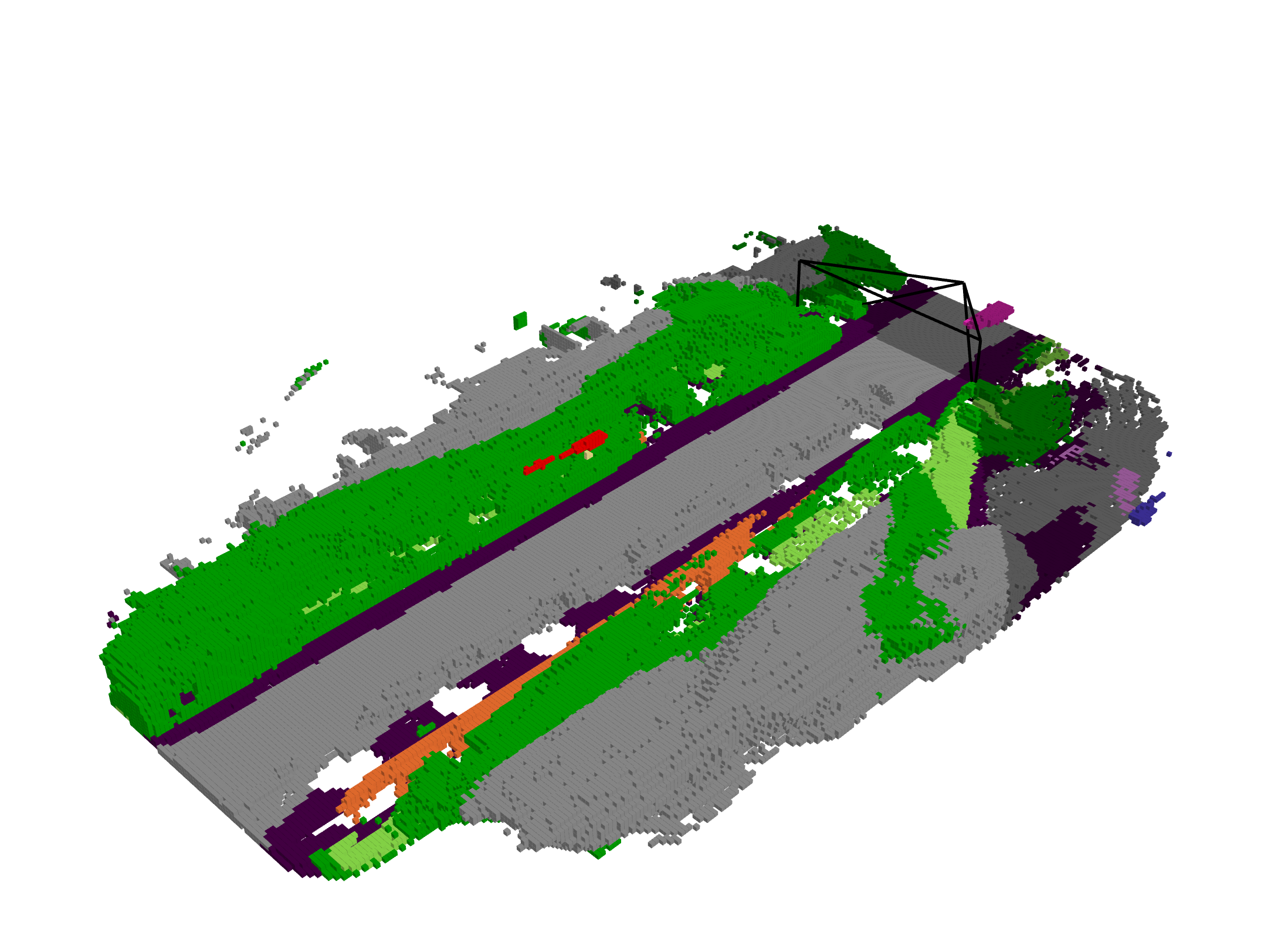}}
\end{minipage}
\begin{minipage}{0.23\linewidth}
    \vspace{3pt}
    \centerline{\includegraphics[width=\textwidth, viewport=200 250 2000 1300,clip]{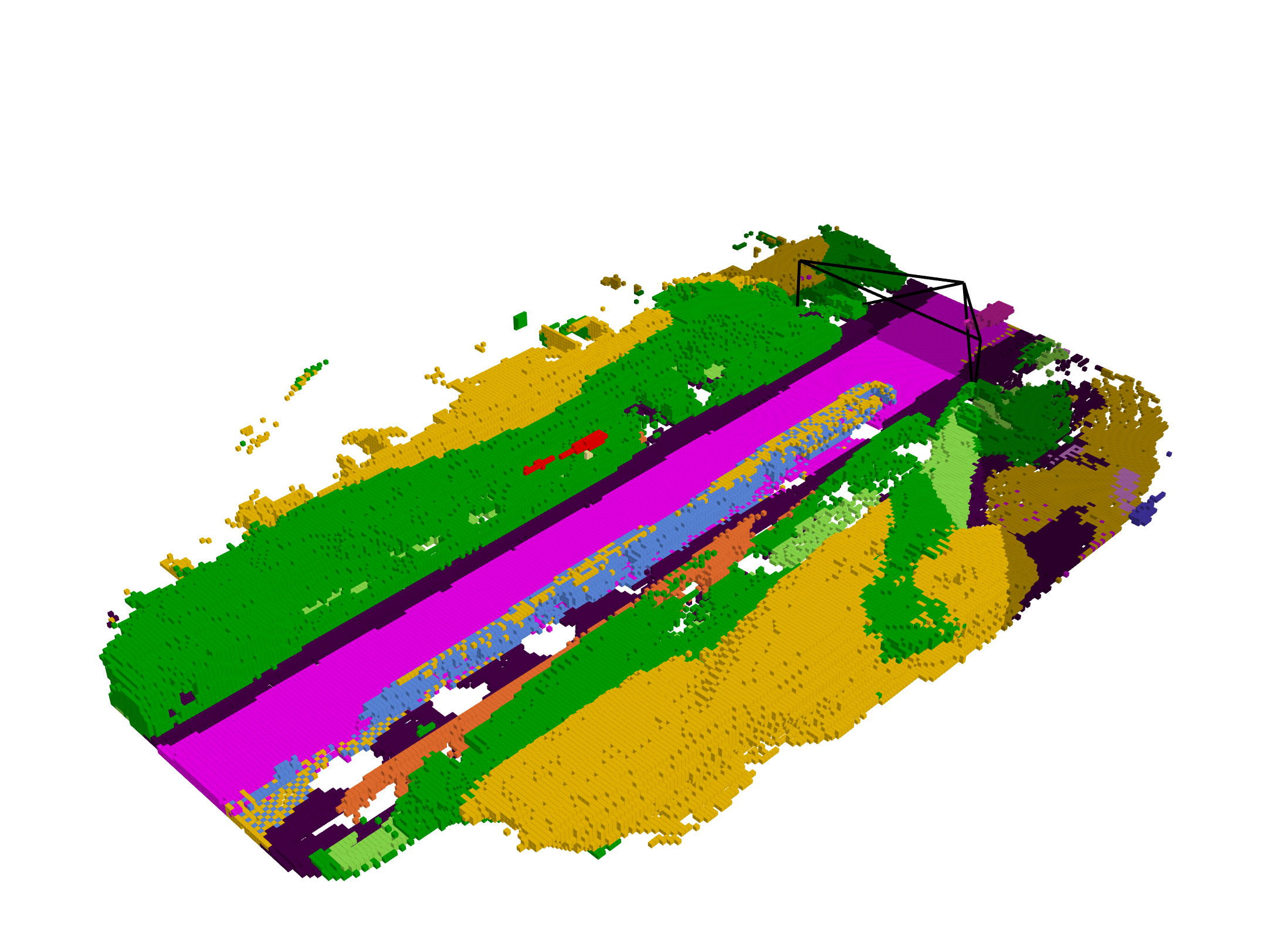}}
\end{minipage}
\begin{minipage}{0.23\linewidth}
    \vspace{3pt}
    \centerline{\includegraphics[width=\textwidth, viewport=200 250 2000 1300,clip]{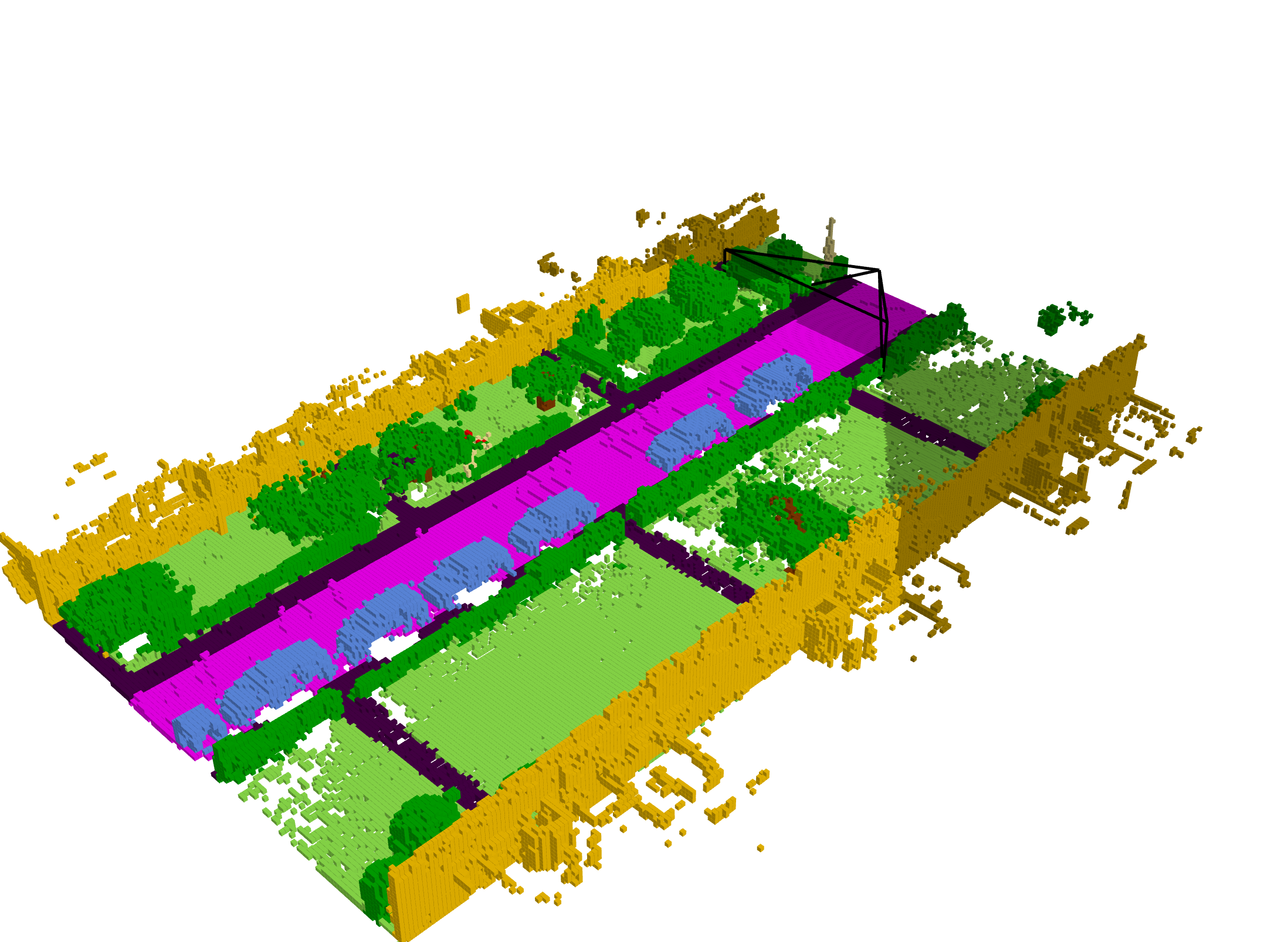}}
\end{minipage}

\begin{minipage}{0.23\linewidth}
    \vspace{3pt}
    \centerline{\includegraphics[width=\textwidth]{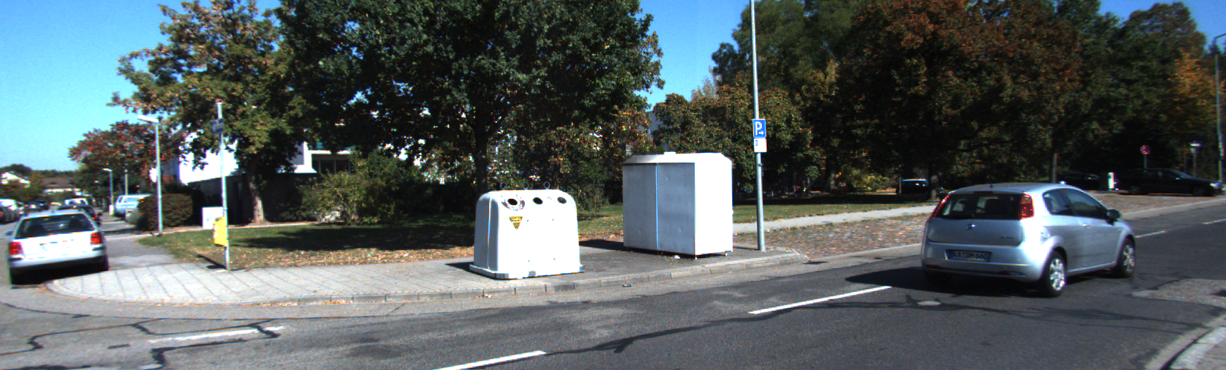}}
\end{minipage}
\begin{minipage}{0.23\linewidth}
    \vspace{3pt}
    \centerline{\includegraphics[width=\textwidth, viewport=200 250 2000 1300,clip]{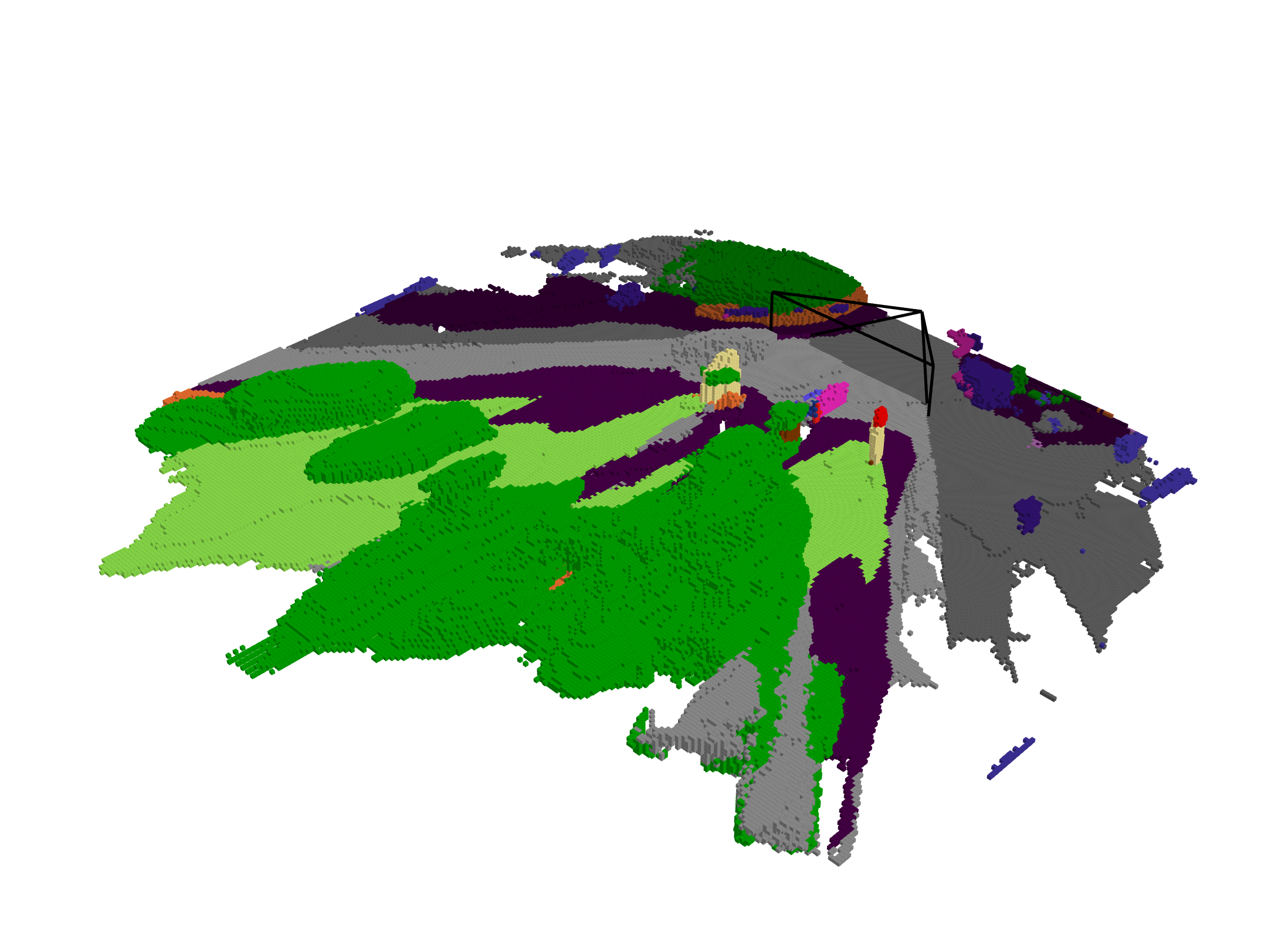}}
\end{minipage}
\begin{minipage}{0.23\linewidth}
    \vspace{3pt}
    \centerline{\includegraphics[width=\textwidth, viewport=200 250 2000 1300,clip]{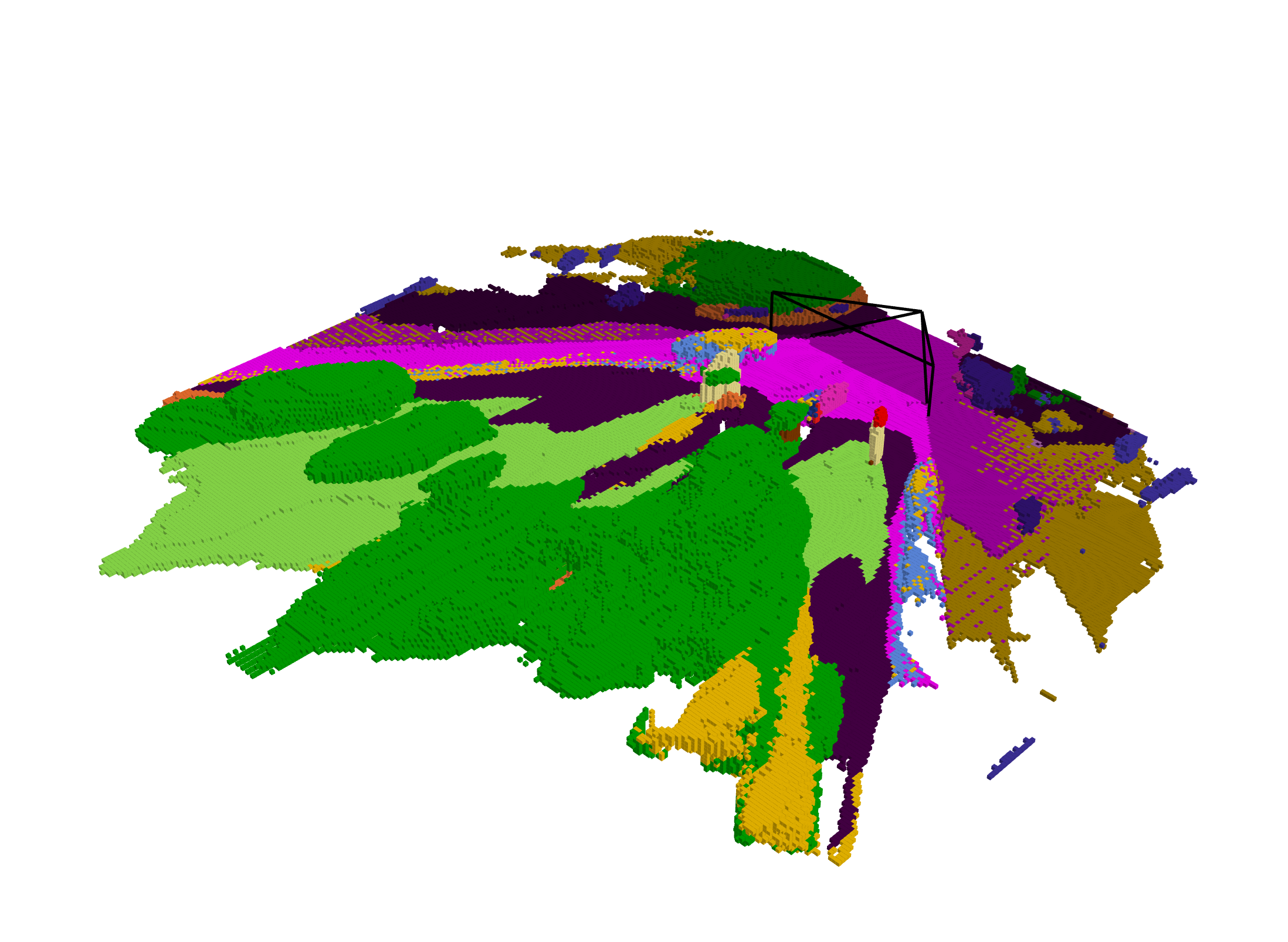}}
\end{minipage}
\begin{minipage}{0.23\linewidth}
    \vspace{3pt}
    \centerline{\includegraphics[width=\textwidth, viewport=200 250 2000 1300,clip]{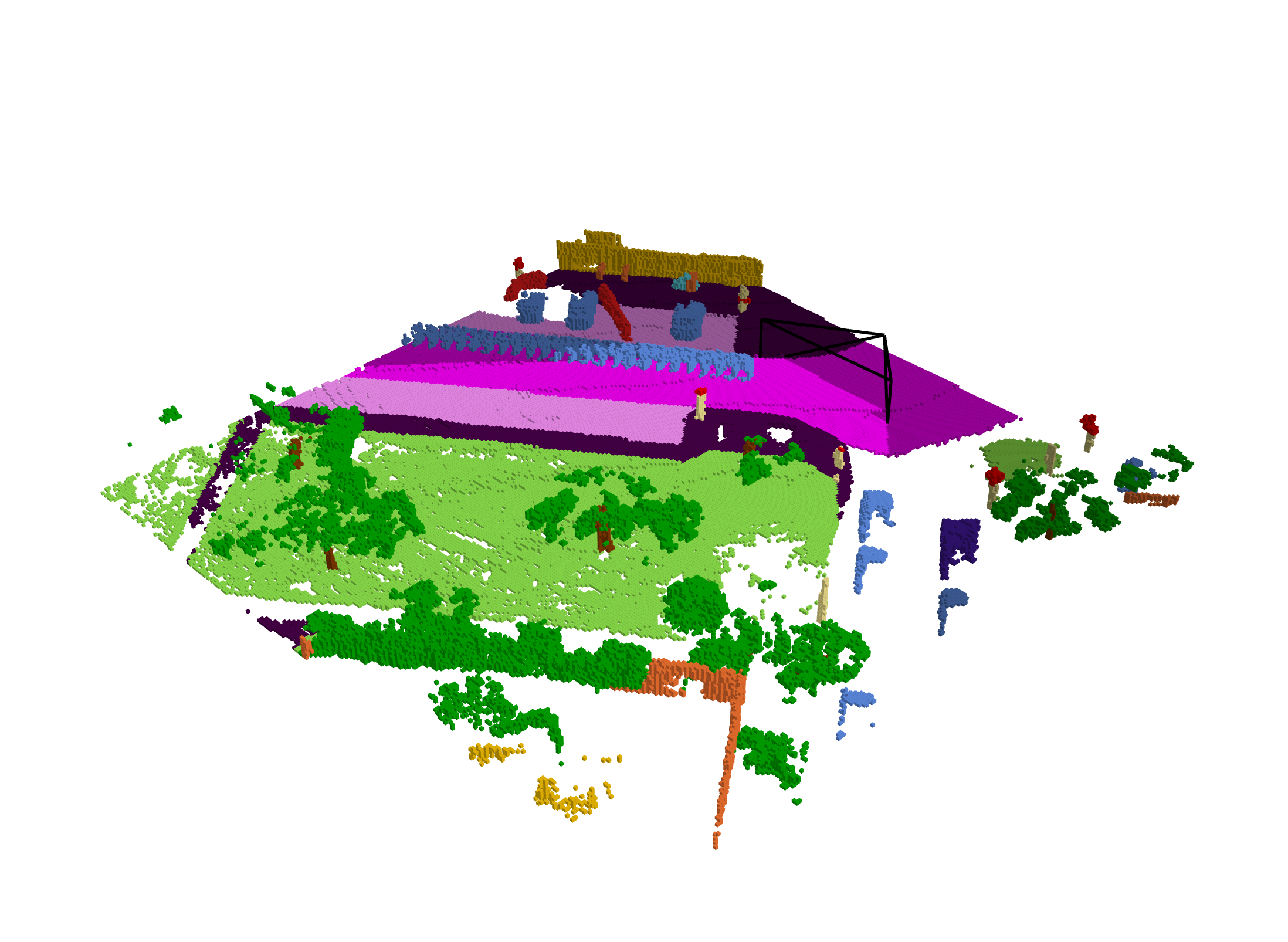}}
\end{minipage}

\begin{minipage}{0.23\linewidth}
    \centering
    RGB input
\end{minipage}
\begin{minipage}{0.23\linewidth}
    \centering
    MonoScene
\end{minipage}
\begin{minipage}{0.23\linewidth}
    \centering
    OVO (ours)
\end{minipage}
\begin{minipage}{0.23\linewidth}
    \centering
    GT
\end{minipage}

\begin{minipage}{\linewidth}
    \centering
    \raisebox{0.5ex}{\colorbox{bicycle}{}}~bicycle \raisebox{0.5ex}{\colorbox{car}{}}~car \raisebox{0.5ex}{\colorbox{motorcycle}{}}~motorcycle \raisebox{0.5ex}{\colorbox{truck}{}}~truck \raisebox{0.5ex}{\colorbox{otherVehicle}{}}~other vehicle \raisebox{0.5ex}{\colorbox{person}{}}~person \raisebox{0.5ex}{\colorbox{bicyclist}{}}~bicyclist \raisebox{0.5ex}{\colorbox{motorcyclist}{}}~motorcyclist \raisebox{0.5ex}{\colorbox{road}{}}~road \raisebox{0.5ex}{\colorbox{parking}{}}~parking \raisebox{0.5ex}{\colorbox{sidewalk}{}}~sidewalk \raisebox{0.5ex}{\colorbox{otherGround}{}}~other ground \raisebox{0.5ex}{\colorbox{building}{}}~building \raisebox{0.5ex}{\colorbox{fence}{}}~fence \raisebox{0.5ex}{\colorbox{vegetation}{}}~vegetation \raisebox{0.5ex}{\colorbox{trunk}{}}~trunk \raisebox{0.5ex}{\colorbox{terrain}{}}~terrain \raisebox{0.5ex}{\colorbox{pole}{}}~pole \raisebox{0.5ex}{\colorbox{trafficSign}{}}~traffic sign \raisebox{0.5ex}{\colorbox{unknown}{}}~unknown
\end{minipage}

\caption{\small \textbf{Qualitative results on SemanticKITTI.} We use the same visualization color code for novel class voxels following Figure~\ref{fig:nyu_qualitative}. The novel classes used for SemanticKITTI dataset include {\color{car} “car”}, {\color{road} “road”}, and {\color{building} “building”}. The visualization in the last row demonstrates that our model is capable of reasonable completion for ``road'' regions outside the field of view, showcasing the effectiveness of our model in handling scenes beyond the visible range. Please see \app for more results.}
\label{fig:kitti_qualitative}
\end{figure*}

%% file: section/disc.tex
\section{Conclusion and Discussion}
Most existing semantic occupancy networks require fully-annotated volumetric data for model training. In this paper, we introduce Open Vocabulary Occupancy (OVO), which allows semantic occupancy prediction of unseen object categories without the need for 3D annotations during training. Essentially, OVO is simple and model-agnostic. As the very first attempt in this direction, we hope OVO to serve as a baseline for future work to build upon and take advantage of. One limitation is that \ourmethod relies on the voxel-wise semantic prediction but without instance-level optimization, potentially resulting in occasional inconsistent prediction within one object. Future work will investigate voxel grouping techniques to improve the overall consistency of voxel labels at the instance level.

%% file: appendix/app_content.tex
\section{Additional Implementation Details}
\label{app_example}

As mentioned in Section 3.2 and Figure 2 of the main paper, we design a 2D distillation ($\phi_\text{2D}$) and a 3D distillation module ($\phi_\text{3D}$) in our Open-Vocabulary Occupancy (OVO) framework. The main purposes of the distillation modules are (1) aligning the dimensions of features and (2) bridging the representation gap for knowledge distillation. Here, we provide additional details.  

\begin{figure}[htbp]
\centering     

\begin{minipage}{0.55\linewidth}
    \begin{minipage}{\linewidth}
        \vspace{3pt}
        \centerline{\includegraphics[width=\textwidth]{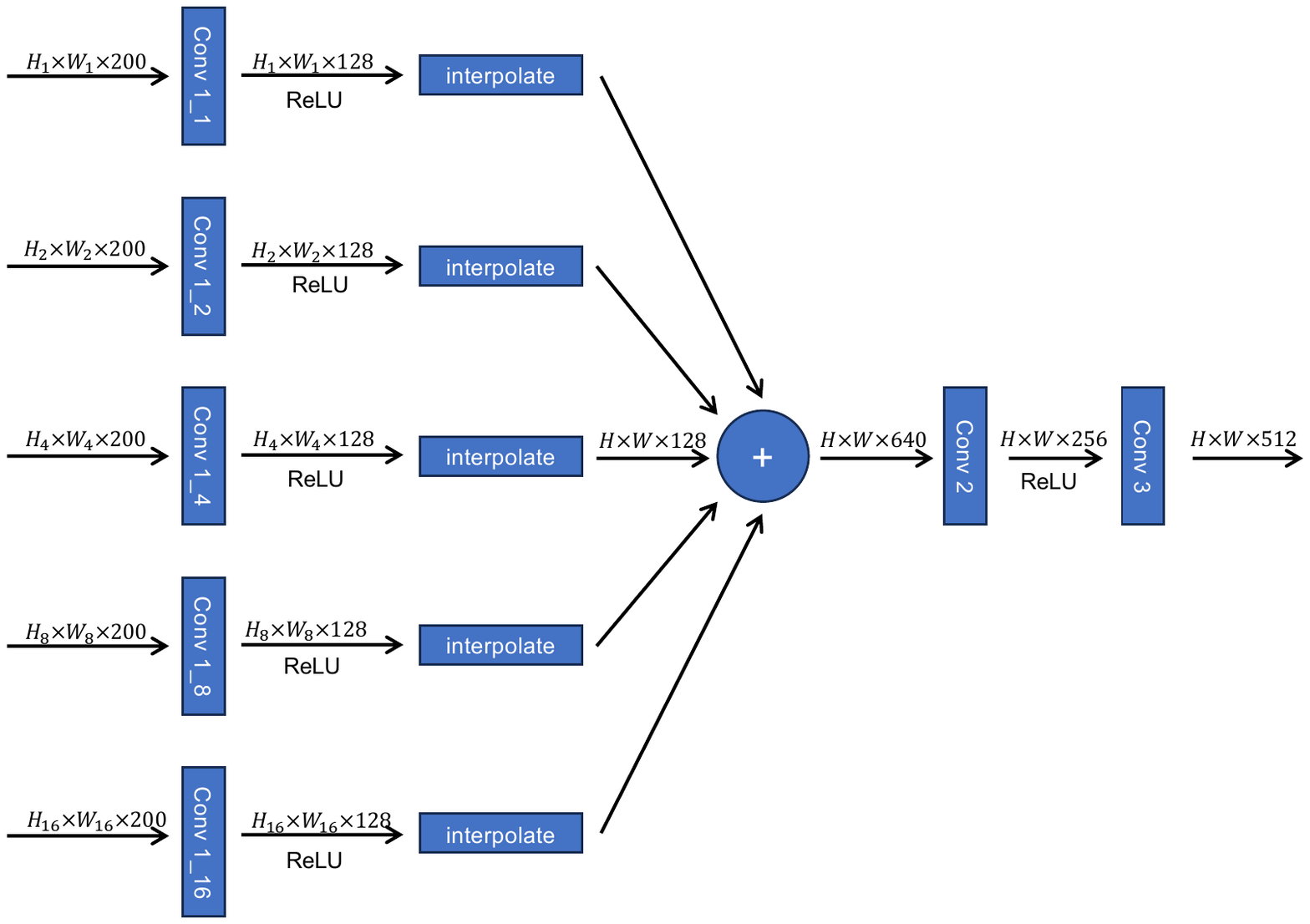}}
    \end{minipage}
    \begin{minipage}{\linewidth}
        \centering
        \small
        (a) 2D Distillation
    \end{minipage}
\end{minipage}
\begin{minipage}{0.44\linewidth}
    
    \begin{minipage}{\linewidth}
        \vspace{3pt}
        \centerline{\includegraphics[width=\textwidth]{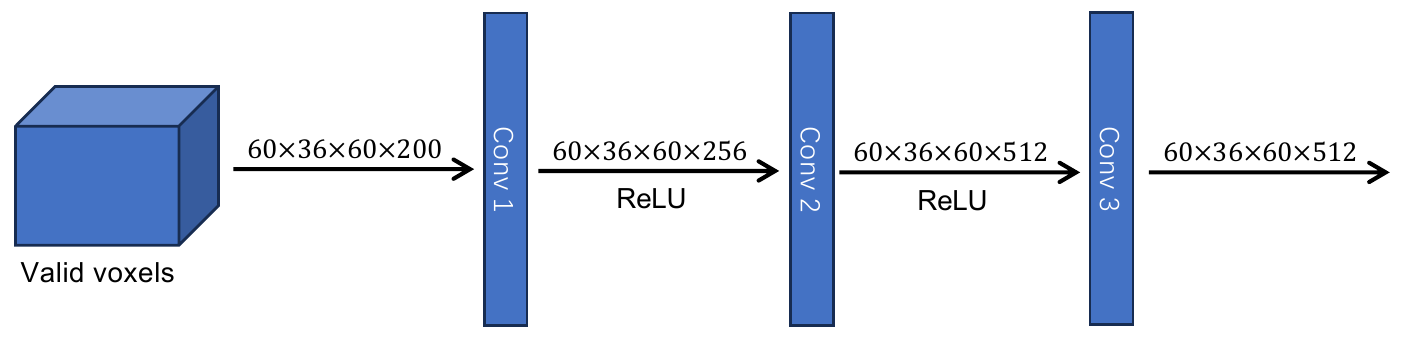}}
    \end{minipage}
    \vspace{25pt}
    \begin{minipage}{\linewidth}
        \centering
        \vspace{3mm}
        \small
        (b) 3D Distillation for NYUv2
    \end{minipage}

    \vspace{5mm}
    \begin{minipage}{\linewidth}
        \vspace{3pt}
        \centerline{\includegraphics[width=\textwidth]{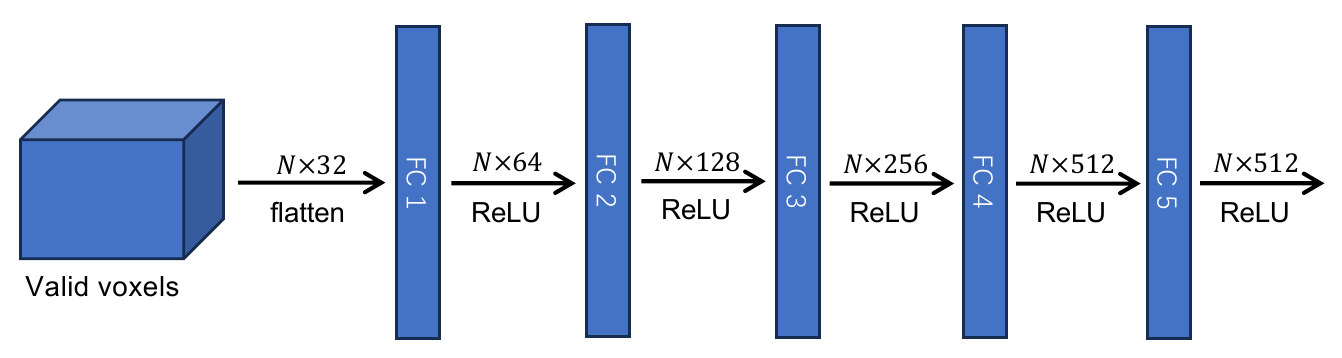}}
    \end{minipage}
    \vspace{10pt}
    \begin{minipage}{\linewidth}
        \centering
        \vspace{4mm}
        \small
        (c) 3D Distillation for SemanticKITTI
    \end{minipage}
\end{minipage}
\caption{\small \textbf{Architecture of the 2D distillation and 3D distillation.} ``$+$'' represents concatenation of all features.}
\label{fig:2d_3d}
\end{figure}

\mypara{2D Distillation.}
As depicted in Figure~\ref{fig:2d_3d}~(a), the 2D distillation module takes 2D features at five different resolutions as the input, which are derived from the output of the 2D backbone. The 200-dimensional features at various scales undergo convolution to yield 128-dimensional features. Next, the features at different scales are bilinearly interpolated to obtain five features with dimensions $H\times W \times 128$. These five features are concatenated, resulting in a final feature of dimension 640. This final feature then passes through two convolutional layers to produce a feature of dimension 512.

\mypara{3D Distillation.}
We design two networks for the 3D distillation module to address memory constraints. For the NYUv2 dataset, we adopt the network structure shown in Figure~\ref{fig:2d_3d}~(b). The voxel features with a size of $60\times 36\times 60\times 200$ are directly processed through a three-layer convolutional neural network (CNN), resulting in final features with a dimension of 512. However, due to the high voxel count in the SemanticKITTI dataset, we first select valid voxels and flatten them. Subsequently, these voxels are fed into a five-layer fully connected network to obtain the final features with a dimension of 512, as illustrated in Figure~\ref{fig:2d_3d}~(c).

%% file: appendix/app_ablation.tex
\vspace{3mm}
\section{Additional Ablation Studies}
\label{app_ablation}
\begin{figure}[t]
\small
\centering     
\begin{minipage}{0.32\linewidth}
    \vspace{3pt}
    \centerline{\includegraphics[width=\textwidth, viewport=150 0 450 350, clip]{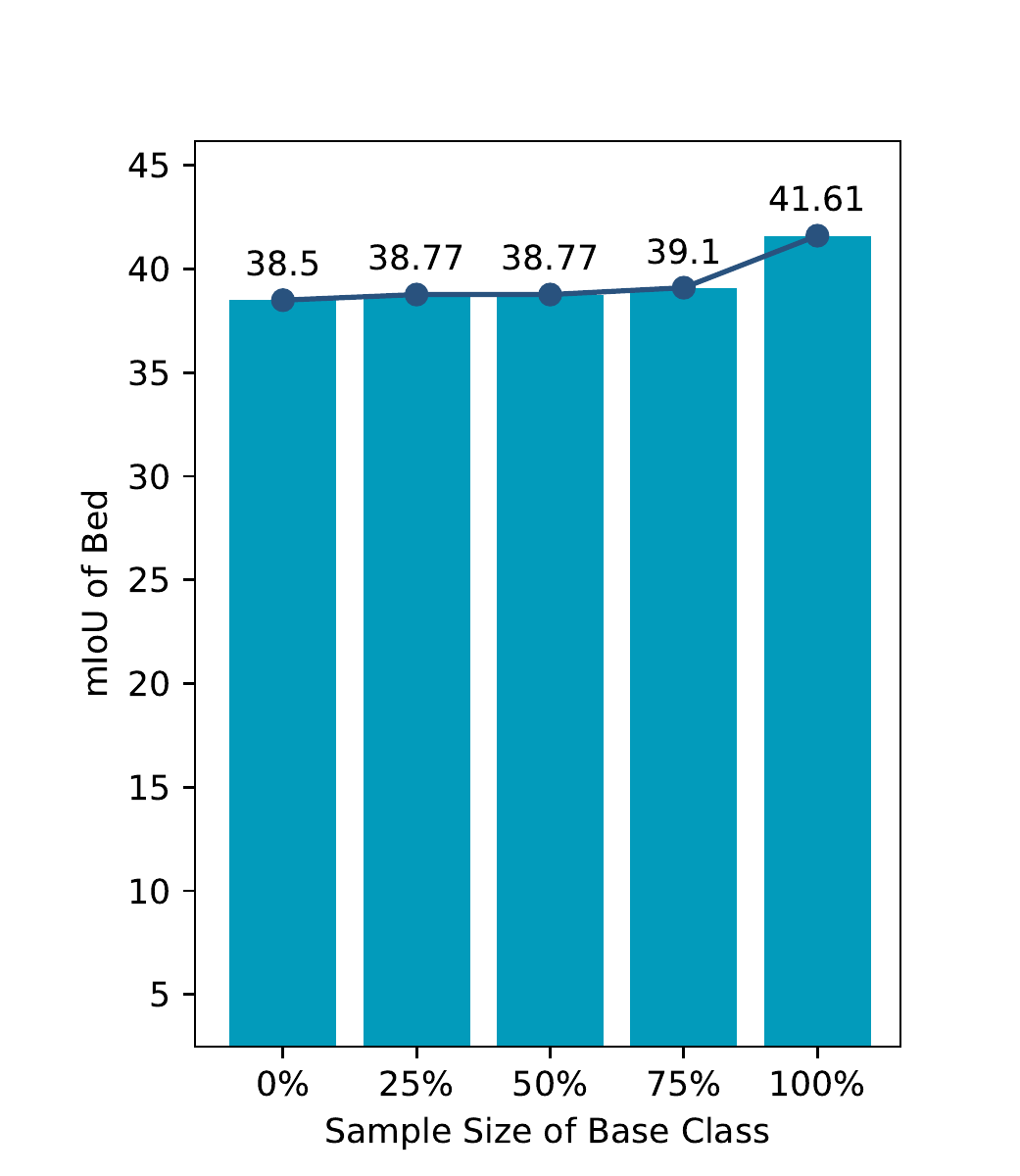}}
\end{minipage}
\begin{minipage}{0.32\linewidth}
    \vspace{3pt}
    \centerline{\includegraphics[width=\textwidth, viewport=150 0 450 350, clip]{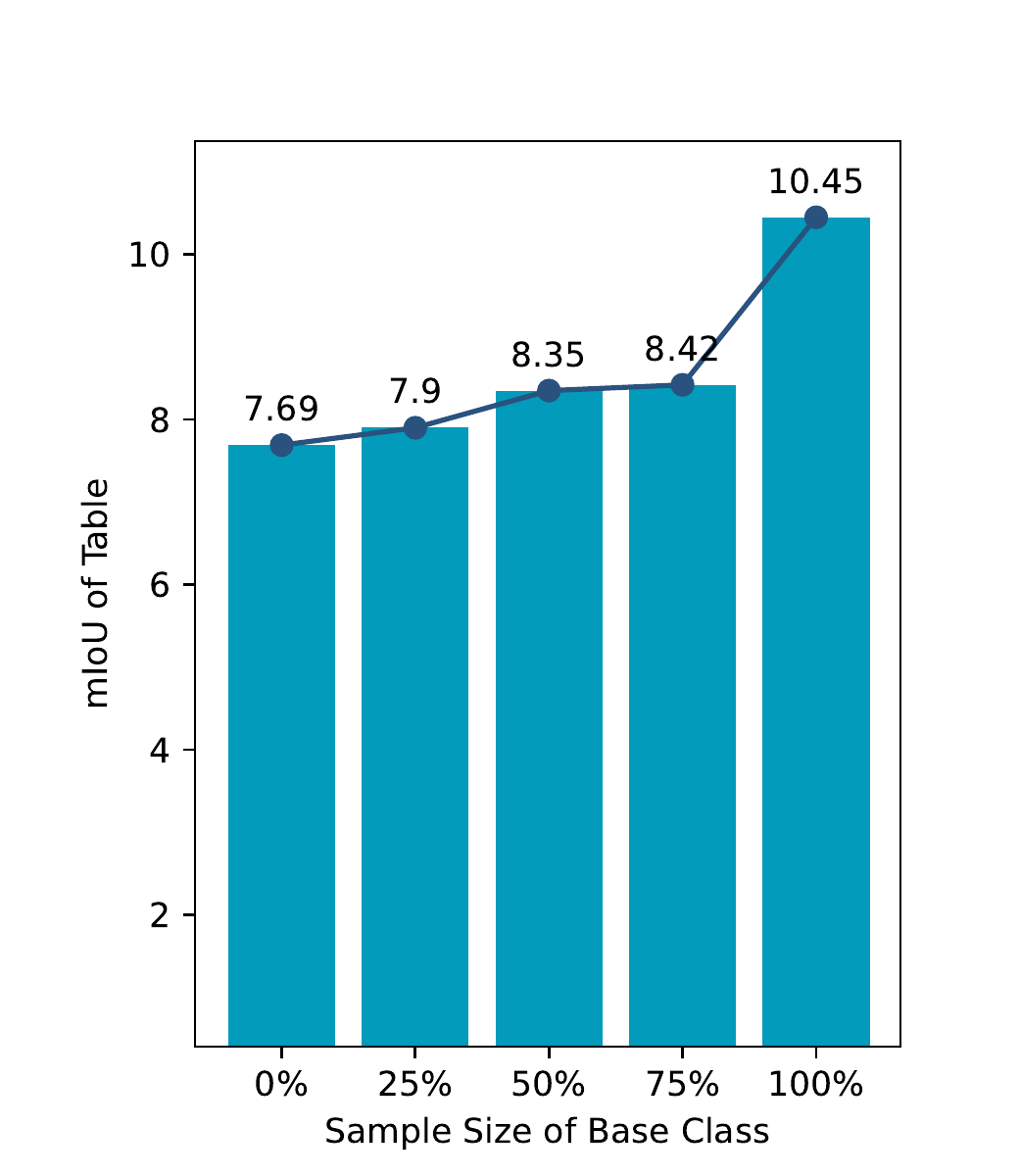}}
\end{minipage}
\begin{minipage}{0.32\linewidth}
    \vspace{3pt}
    \centerline{\includegraphics[width=\textwidth, viewport=150 0 450 350, clip]{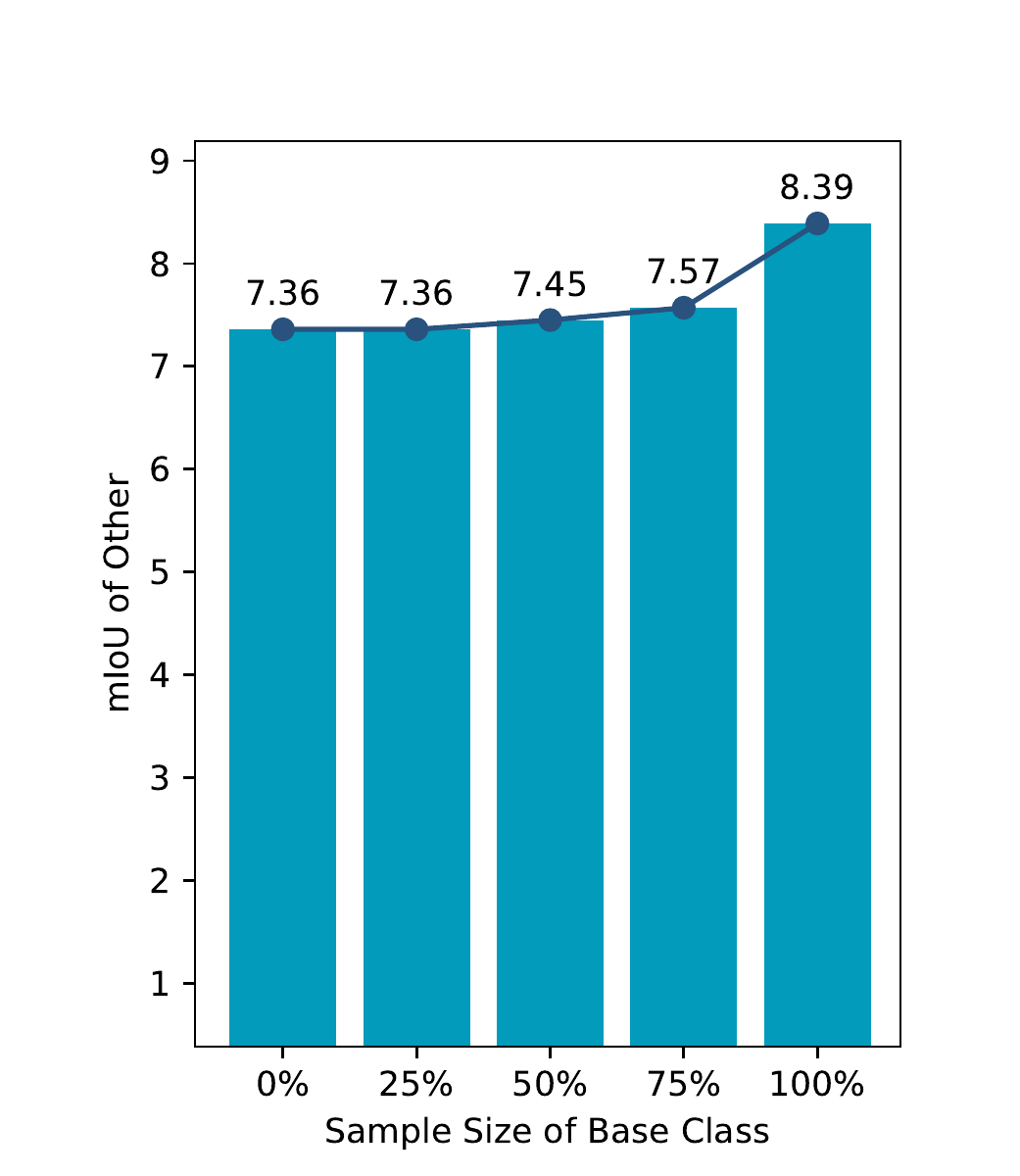}}
\end{minipage}

\begin{minipage}{0.32\linewidth}
    \centering
    (a) bed
\end{minipage}
\begin{minipage}{0.32\linewidth}
    \centering
    (b) table
\end{minipage}
\begin{minipage}{0.32\linewidth}
    \centering
    (c) other
\end{minipage}

\caption{\small \textbf{Ablation study on the sample size of base classes.} The information in the figure indicates that as the sample size of the base class increases, OVO performs better.}
\label{fig:sample_size_base}
\end{figure}

\mypara{Impact of the Sample Size of Base Classes.}
To evaluate the robustness of OVO with respect to the sample size of the base class, we randomly discard a portion of base class samples. The experimental results are shown in Figure~\ref{fig:sample_size_base}, where (a), (b), and (c) represent the mIoU of bed, table, and other, respectively, under different sample sizes of the base class. It can be observed that as the number of base class samples increases, the performance (mIoU) also increases. Furthermore, the experimental results demonstrate that OVO exhibits robustness to the sample size of the base class. Even when the base class sample size is 0, OVO still achieves reasonably good performance.

\mypara{Impact of the Choice of Base Class.}
In addition to evaluating the robustness of OVO with respect to the categories of the base class, we also completely discard certain base classes to examine the impact of the choice of base class on the experimental results. As shown in Table~\ref{table:choice_base}, we only use four base classes and OVO still achieves promising results. It is worth noting that these four base classes were easy to annotate (\ie, ``ceiling'', ``floor'', ``wall'', ``window''), which further demonstrates that OVO can be trained using easily annotated data to achieve satisfactory performance.
\begin{table}[t]
\small
\tabcolsep 6pt
\centering
\caption{\small \textbf{Ablation study on the choice of base class.} }
\label{table:choice_base}
\vspace{1mm}
\begin{tabular}{cccccccc|cccc} \toprule
\multicolumn{8}{c|}{Base Class} &\multicolumn{4}{c}{Novel Class (mIoU$\uparrow$)}  \\  ceiling & floor & wall & window & chair & sofa & tvs & furniture  & \multicolumn{1}{c}{bed} & \multicolumn{1}{c}{table} & \multicolumn{1}{c}{other} & \multicolumn{1}{c}{mean} \\ \midrule

\cmark & \cmark & \cmark & \cmark &  &  &  &  & 38.15        &  8.14          &  7.29          & 17.86 \\
\cmark & \cmark & \cmark & \cmark & \cmark & \cmark & \cmark & \cmark & \textbf{41.61}        & \textbf{10.45}          &  \textbf{8.39}          & \textbf{20.15} \\ \bottomrule

\end{tabular} 
\end{table}

\input{table/app_new_novel}

\mypara{Impact of the Choice of Novel Classes.}
To validate the robustness of OVO with respect to the choice of novel classes, we pick ``sofa'', ``furniture'', and ``other'' as the novel classes and evaluate the performance of OVO. The experimental results are presented in Table~\ref{table:new_novel}, indicating that even with the change in novel classes, OVO can achieve good performance and outperforms the fully supervised method AICNet~\cite{AICNet}. This shows the robustness of OVO in handling the choice of novel classes.

%% file: table/app_new_novel.tex
\begin{table}[t]
\caption{\small \textbf{Performance (mIoU$\uparrow$) on NYUv2}~\cite{nyu} \textbf{with different novel Class}. \textbf{C}: camera; \textbf{D}: depth; $\dagger$: TSDF}
\label{table:new_novel}
\vspace{1mm}
\footnotesize
\centering
\tabcolsep 1.4pt
\begin{tabular}{l|c|cccc|ccccccccc}
\toprule
 & & \multicolumn{4}{c|}{{\textbf{(a) Novel Class}}}  & \multicolumn{9}{c}{\textbf{(b) Base Class}}\\
\multicolumn{1}{l|}{Method} & Input & \rotatebox{90}{\raisebox{0.5ex}{\colorbox{sofa}{}}~{sofa}} & \rotatebox{90}{\raisebox{0.5ex}{\colorbox{furniture}{}}~{furniture}} & \rotatebox{90}{\raisebox{0.5ex}{\colorbox{other}{}}~{other}} & {mean} & \rotatebox{90}{\raisebox{0.5ex}{\colorbox{ceiling}{}}~{ceiling}} & \rotatebox{90}{\raisebox{0.5ex}{\colorbox{floor}{}}~{floor}} & \rotatebox{90}{\raisebox{0.5ex}{\colorbox{wall}{}}~{wall}} & \rotatebox{90}{\raisebox{0.5ex}{\colorbox{window}{}}~{window}} & \rotatebox{90}{\raisebox{0.5ex}{\colorbox{chair}{}}~{chair}} & \rotatebox{90}{\raisebox{0.5ex}{\colorbox{bed}{}}~{bed}} & \rotatebox{90}{\raisebox{0.5ex}{\colorbox{table}{}}~{table}} & \rotatebox{90}{\raisebox{0.5ex}{\colorbox{tvs}{}}~{tvs}} & {mean} \\ 
\midrule

\multicolumn{14}{l}{\emph{\textbf{Fully-supervised}}}\\
\rowcolor{supervised} (1) AICNet~\cite{AICNet}   & C, D       & 22.92               & 15.90          & 6.45           & 15.09                      & 7.58             & 82.97          & 9.15          & 0.05            & 6.93        & 35.87        & 11.11         & 0.71          & 19.30                  \\
\rowcolor{supervised}(2) SSCNet~\cite{sscnet}   &C, D    & 35.00                 & 27.10          & 10.10           & 24.07                        & 15.10             & 94.70          & 24.40          & 0.00            & 12.60     & 32.10        & 13.00           & 7.80        & 24.96                  \\
\rowcolor{supervised}(3) 3DSketch~\cite{3DSketch}   &C$\dagger$    & 29.21               & 23.83           & 8.19           & 20.41                        & 8.53             & 90.45          & 9.94          & 5.67            & 10.64   & 42.29        & 13.88             & 9.38         & 23.85                  \\
\rowcolor{supervised}(4) MonoScene~\cite{cao2022monoscene}   & C    & 36.11               & 27.96             & 12.94          & 25.67                        & 8.89             & 93.50          & 12.06         & 12.57           & 13.72             & 48.19        & 15.13    & 15.22        & 27.41              \\ \midrule
\multicolumn{14}{l}{\emph{\textbf{Zero-shot}}}  \\
(5) \ourmethodbold (ours)   &C        & 16.12        & 23.36           & 6.76           & 15.41                     & 7.31             & 93.31          & 7.60          & 7.31            & 9.46          & 45.35         & 10.75         & 6.43              & 23.44                 \\
\bottomrule
\end{tabular}\vspace{-3mm}
\end{table}

%% file: appendix/app_qualitative.tex
\section{Additional Qualitative Visualization}
\label{app_qualitative}

\begin{figure}[htbp]
\centering     

\begin{minipage}{0.3\linewidth}
    \vspace{3pt}
    \centerline{\includegraphics[width=\textwidth]{figure/appendix_v2/NYU1449_0000_color.jpg}}
\end{minipage}
\begin{minipage}{0.3\linewidth}
    \vspace{3pt}
    \centerline{\includegraphics[width=\textwidth, viewport=400 0 1400 900,clip]{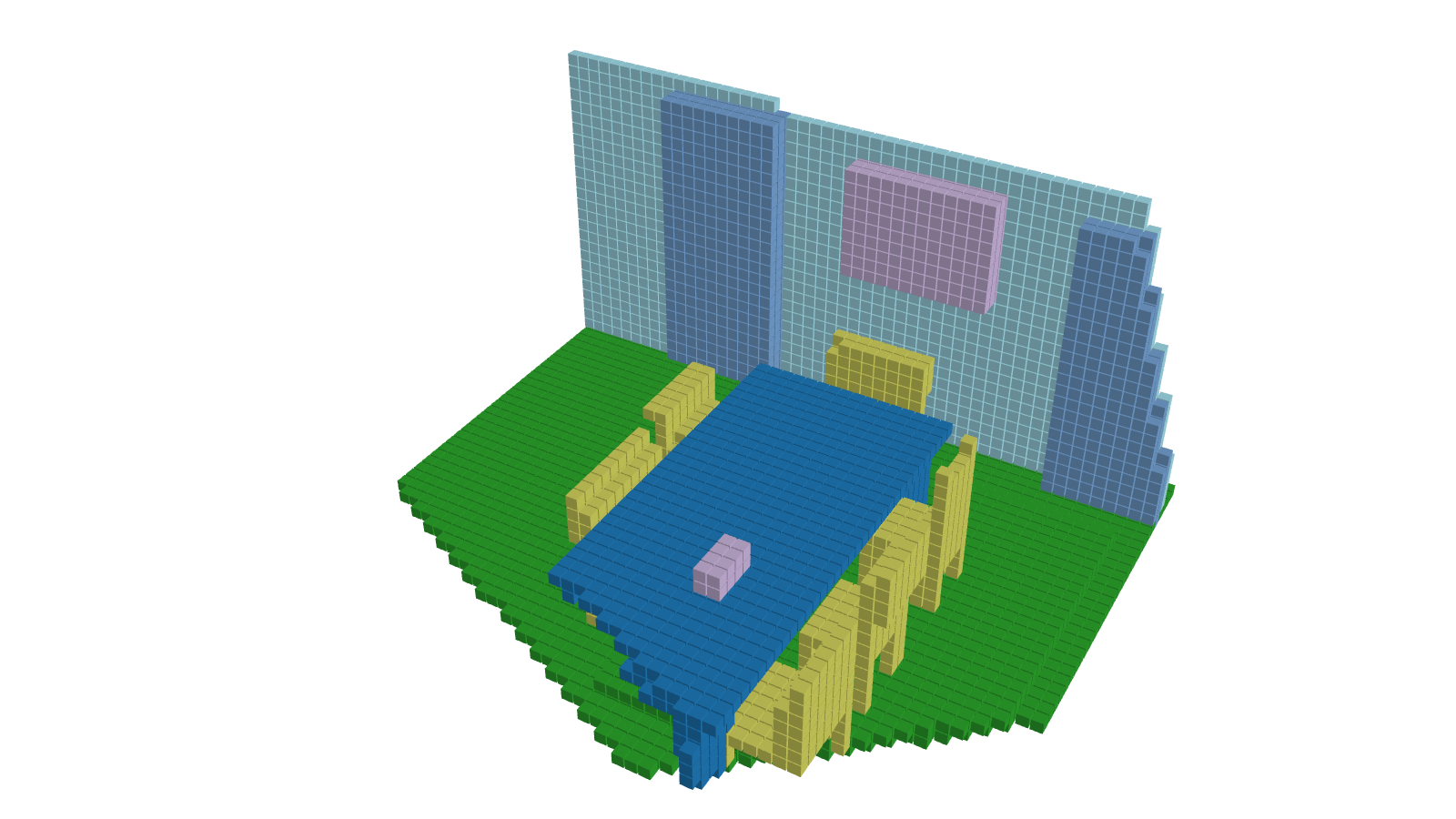}}
\end{minipage}
\begin{minipage}{0.3\linewidth}
    \vspace{3pt}
    \centerline{\includegraphics[width=\textwidth, viewport=400 0 1400 900,clip]{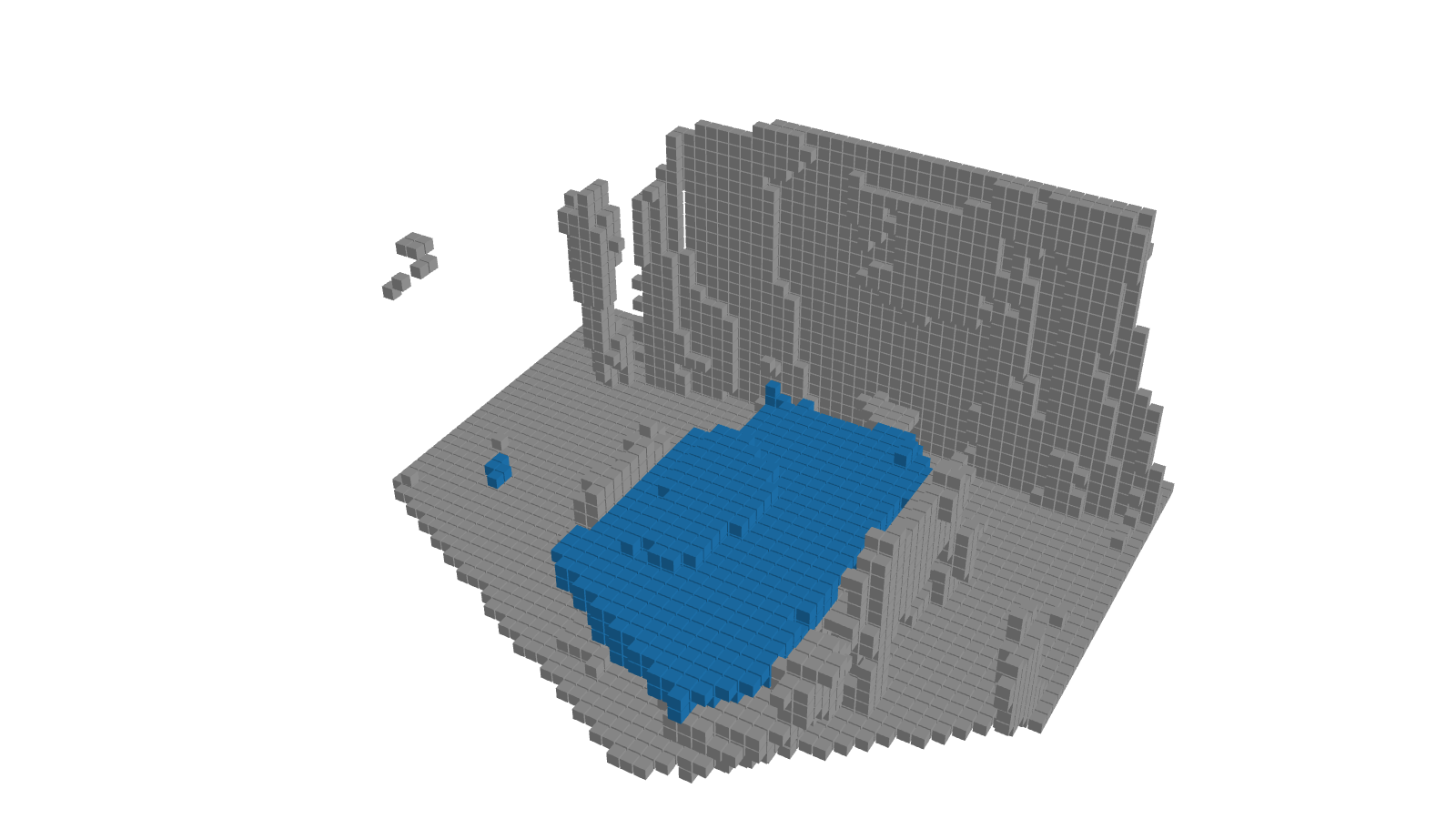}}
\end{minipage}

\begin{minipage}{0.3\linewidth}
    \centering
    RGB input
\end{minipage}
\begin{minipage}{0.3\linewidth}
    \centering
    GT
\end{minipage}
\begin{minipage}{0.3\linewidth}
    \centering
    ``table'' (default) \\ mIoU = 24.04
\end{minipage}

\begin{minipage}{0.3\linewidth}
    \vspace{3pt}
    \centerline{\includegraphics[width=\textwidth, viewport=400 0 1400 900,clip]{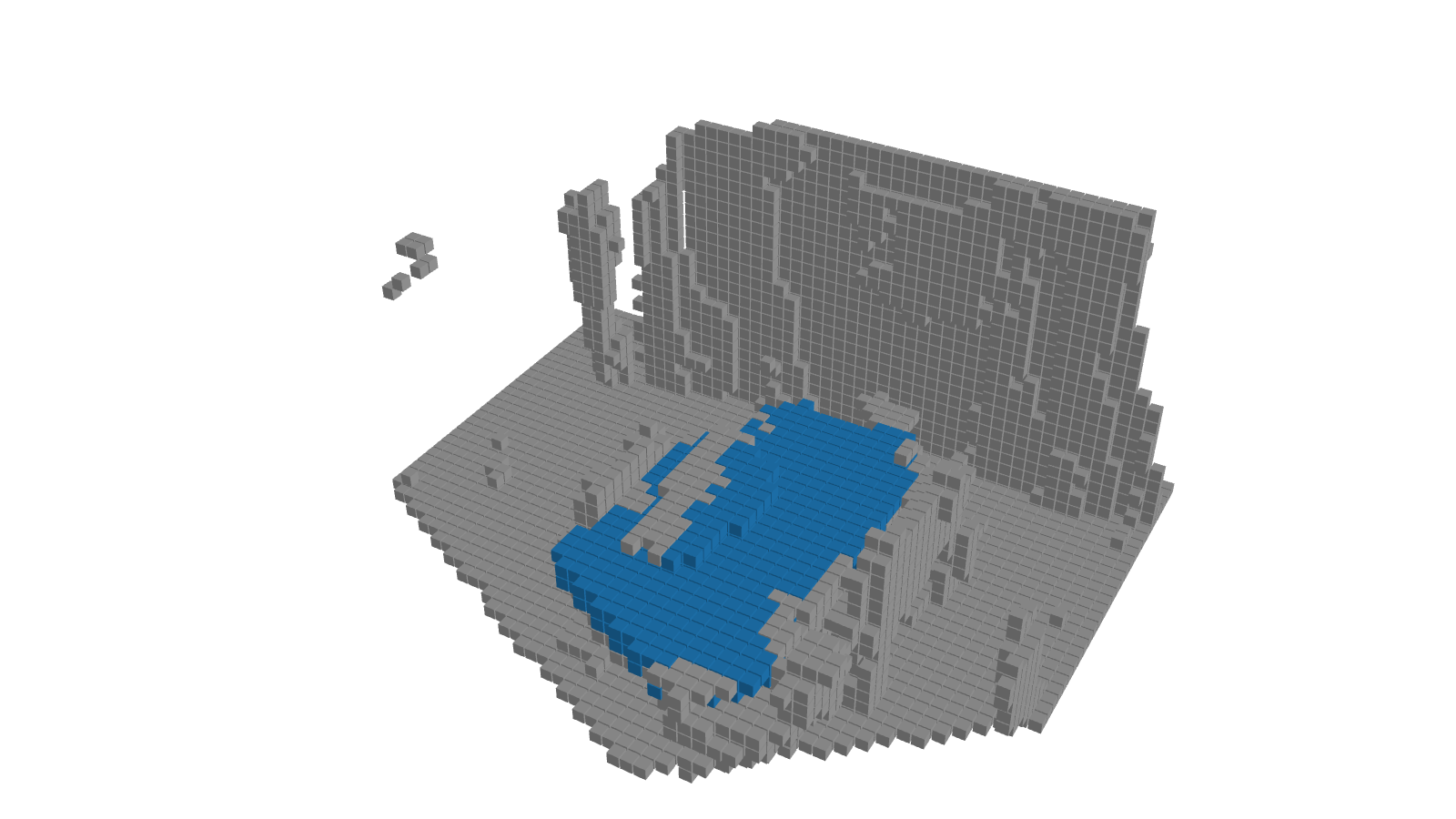}}
\end{minipage}
\begin{minipage}{0.3\linewidth}
    \vspace{3pt}
    \centerline{\includegraphics[width=\textwidth, viewport=400 0 1400 900,clip]{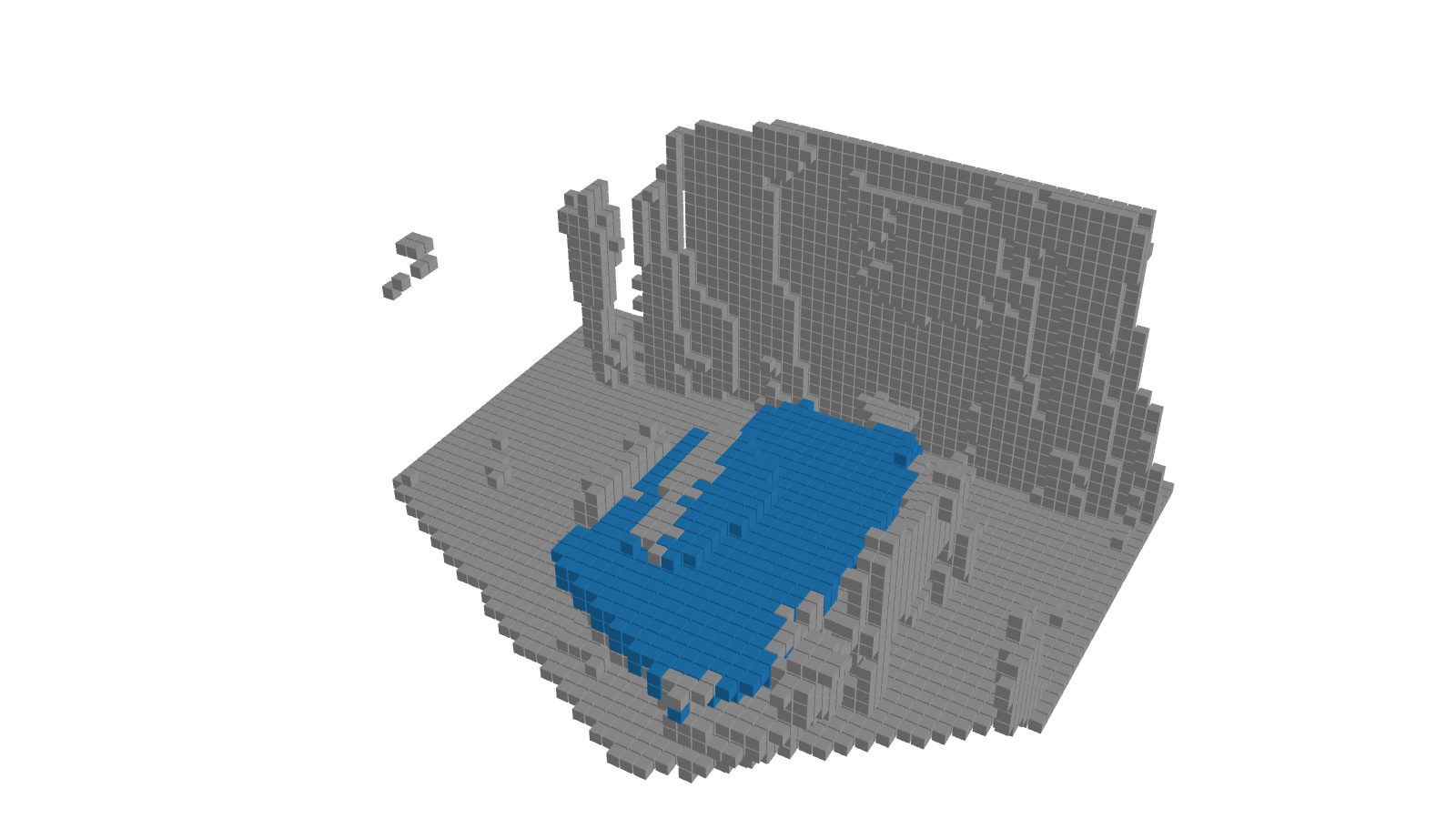}}
\end{minipage}
\begin{minipage}{0.3\linewidth}
    \vspace{3pt}
    \centerline{\includegraphics[width=\textwidth, viewport=400 0 1400 900,clip]{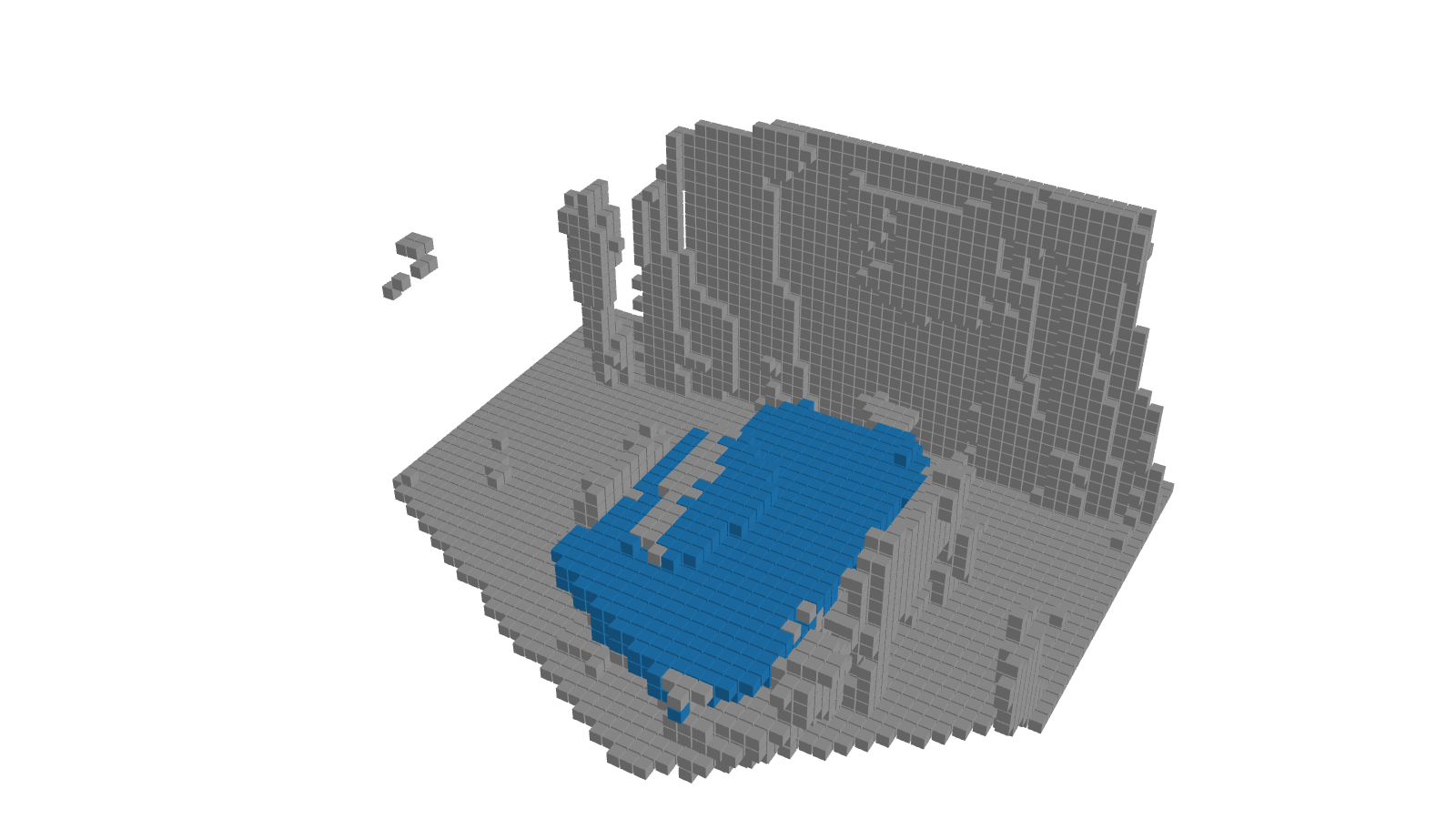}}
\end{minipage}

\begin{minipage}{0.3\linewidth}
    \centering
    ``brown table'' \\ mIoU = \textbf{25.32}
\end{minipage}
\begin{minipage}{0.3\linewidth}
    \centering
    ``conference table'' \\ mIoU = \underline{25.00}
\end{minipage}
\begin{minipage}{0.3\linewidth}
    \centering
    ``the table near the chair'' \\ mIoU = 24.74
\end{minipage}

\begin{minipage}{\linewidth}
    \vspace{5pt}
    \centering
    \raisebox{0.5ex}{\colorbox{ceiling}{}}~ceiling \raisebox{0.5ex}{\colorbox{floor}{}}~floor \raisebox{0.5ex}{\colorbox{wall}{}}~wall \raisebox{0.5ex}{\colorbox{window}{}}~window \raisebox{0.5ex}{\colorbox{chair}{}}~chair \raisebox{0.5ex}{\colorbox{bed}{}}~bed \raisebox{0.5ex}{\colorbox{sofa}{}}~sofa \raisebox{0.5ex}{\colorbox{table}{}}~table \raisebox{0.5ex}{\colorbox{tvs}{}}~tvs \raisebox{0.5ex}{\colorbox{furniture}{}}~furniture \raisebox{0.5ex}{\colorbox{other}{}}~other \raisebox{0.5ex}{\colorbox{unknown}{}}~unknown
\end{minipage}

\caption{\small \textbf{Qualitative results of ablation study on prompts for ``table'' on \textit{NYU1449\_0000}.} The information in the figure indicates that using detailed queries for prompting leads to better performance of OVO.}
\label{fig:nyu_attr}
\end{figure}

\mypara{Qualitative Results of Different Text Queries.}
To demonstrate the open vocabulary capability of the OVO model, we utilize different queries with various attributes to prompt the presence of ``table'' in the \textit{NYU1449\_0000} scene. The qualitative results are shown in Figure~\ref{fig:nyu_attr}. The mIoU values indicate that prompts with attributes lead to improved performance. The experimental results show that employing more detailed queries for prompting enables OVO to achieve better performance.

\mypara{Additional Qualitative Visualization Results on NYUv2~\cite{nyu} and SemanticKITTI~\cite{semantickitti}.}
Figure~\ref{fig:app_nyu_qualitative} and Figure~\ref{fig:app_kitti_qualitative} shows additional qualitative results.

\input{table/app_qual_nyu}

\input{table/app_qual_kitti}

%% file: table/app_qual_nyu.tex
\begin{figure}[htbp]
\centering

\begin{minipage}{0.23\linewidth}
    \vspace{3pt}
    \centerline{\includegraphics[width=\textwidth]{figure/app_nyu_exp/NYU0271_0000_color.jpg}}
\end{minipage}
\begin{minipage}{0.23\linewidth}
    \vspace{3pt}
    \centerline{\includegraphics[width=\textwidth, viewport=500 0 2800 1700,clip]{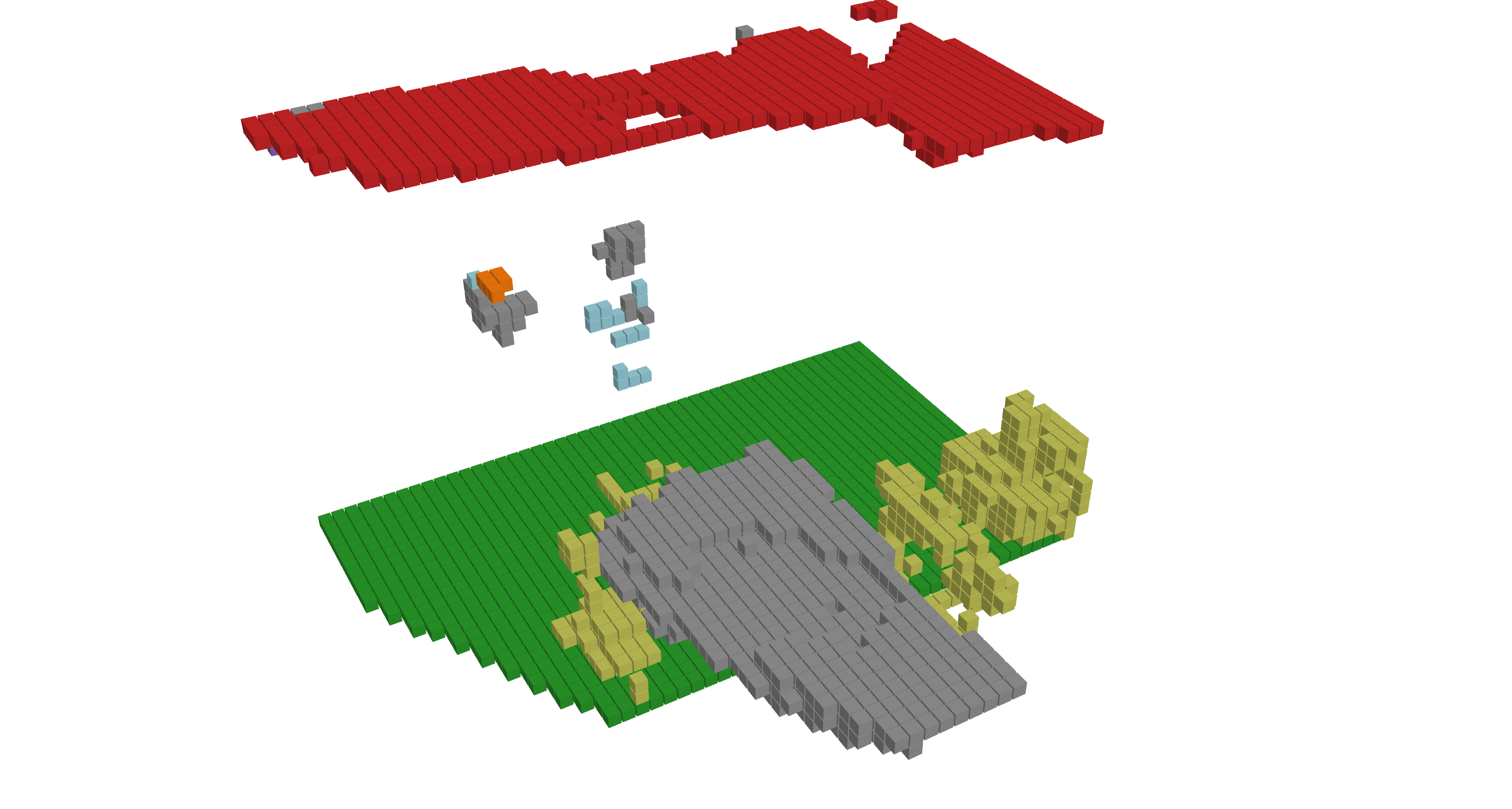}}
\end{minipage}
\begin{minipage}{0.23\linewidth}
    \vspace{3pt}
    \centerline{\includegraphics[width=\textwidth, viewport=500 0 2800 1700,clip]{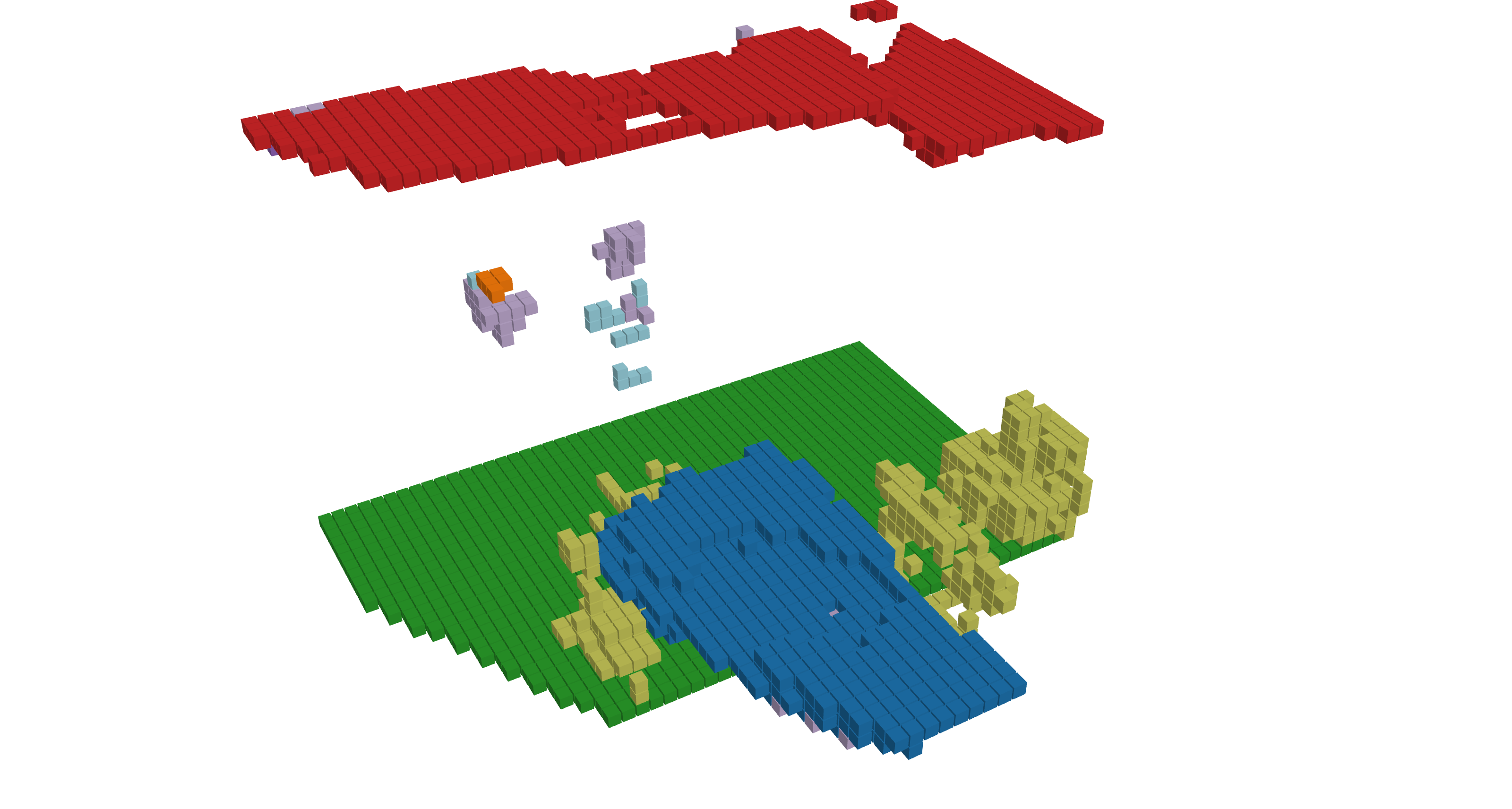}}
\end{minipage}
\begin{minipage}{0.23\linewidth}
    \vspace{3pt}
    \centerline{\includegraphics[width=\textwidth, viewport=500 0 2800 1700,clip]{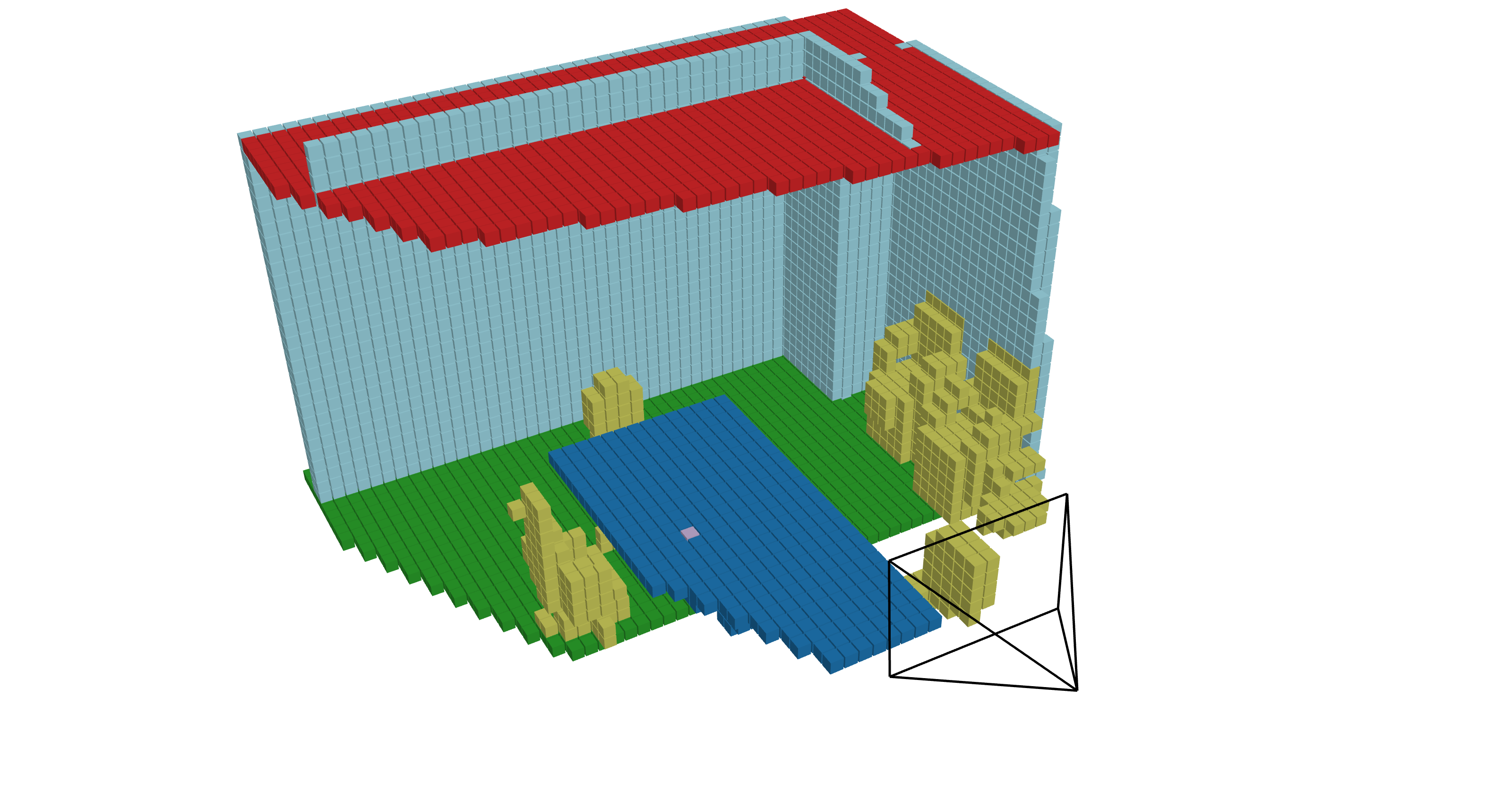}}
\end{minipage}

\begin{minipage}{0.23\linewidth}
    \vspace{3pt}
    \centerline{\includegraphics[width=\textwidth]{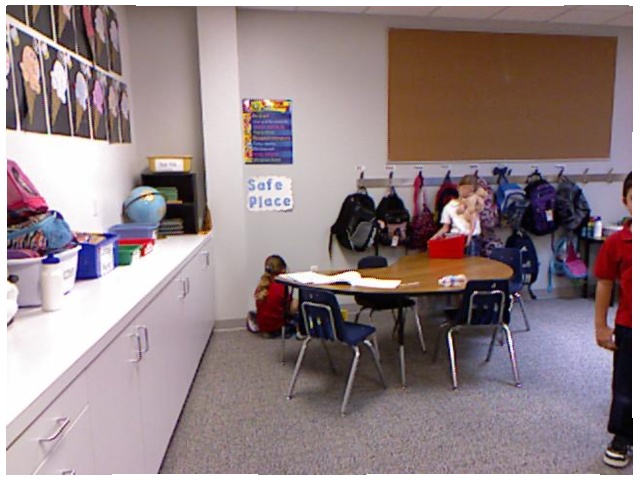}}
\end{minipage}
\begin{minipage}{0.23\linewidth}
    \vspace{3pt}
    \centerline{\includegraphics[width=\textwidth, viewport=500 0 1800 1500,clip]{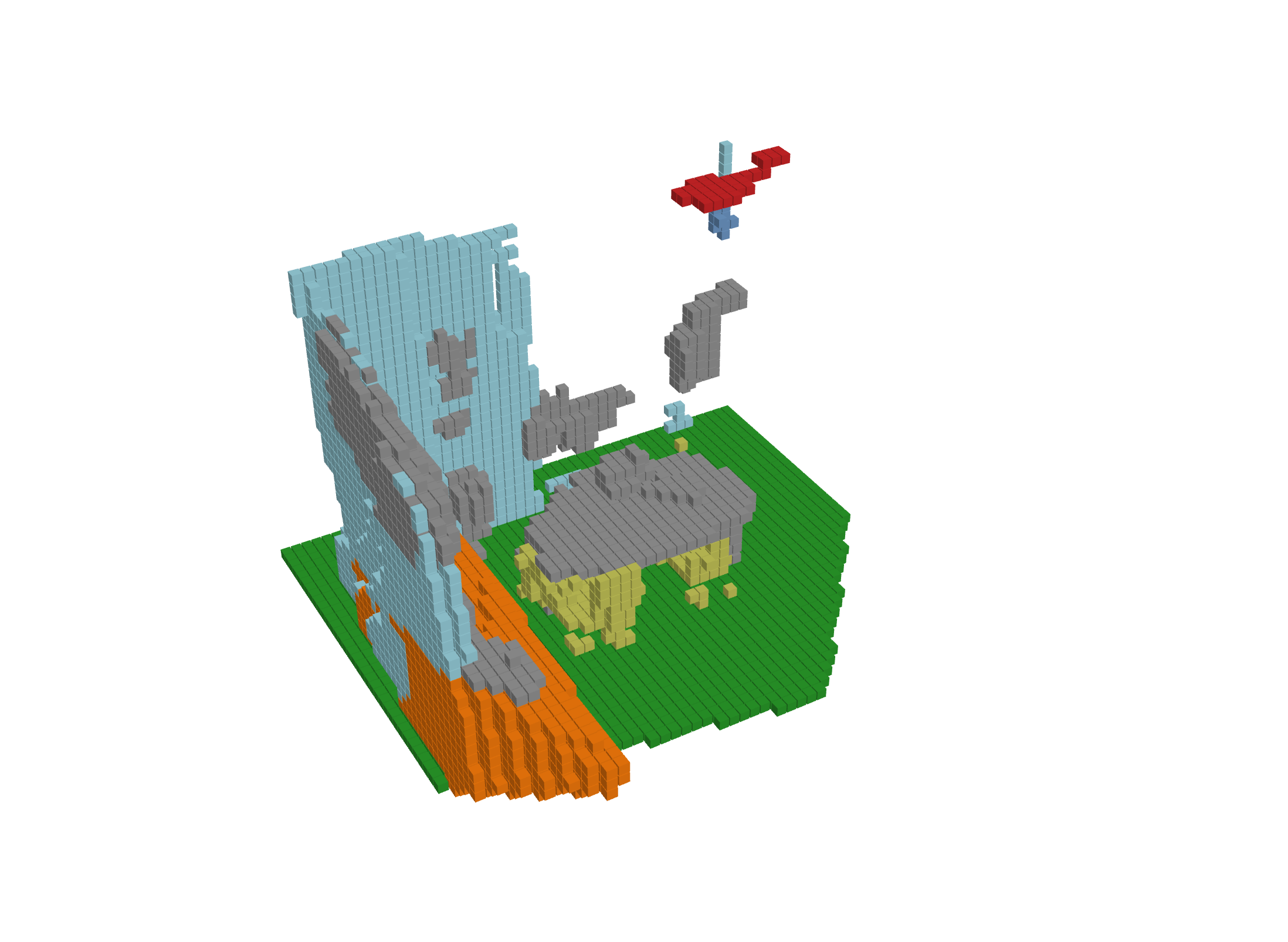}}
\end{minipage}
\begin{minipage}{0.23\linewidth}
    \vspace{3pt}
    \centerline{\includegraphics[width=\textwidth, viewport=500 0 1800 1500,clip]{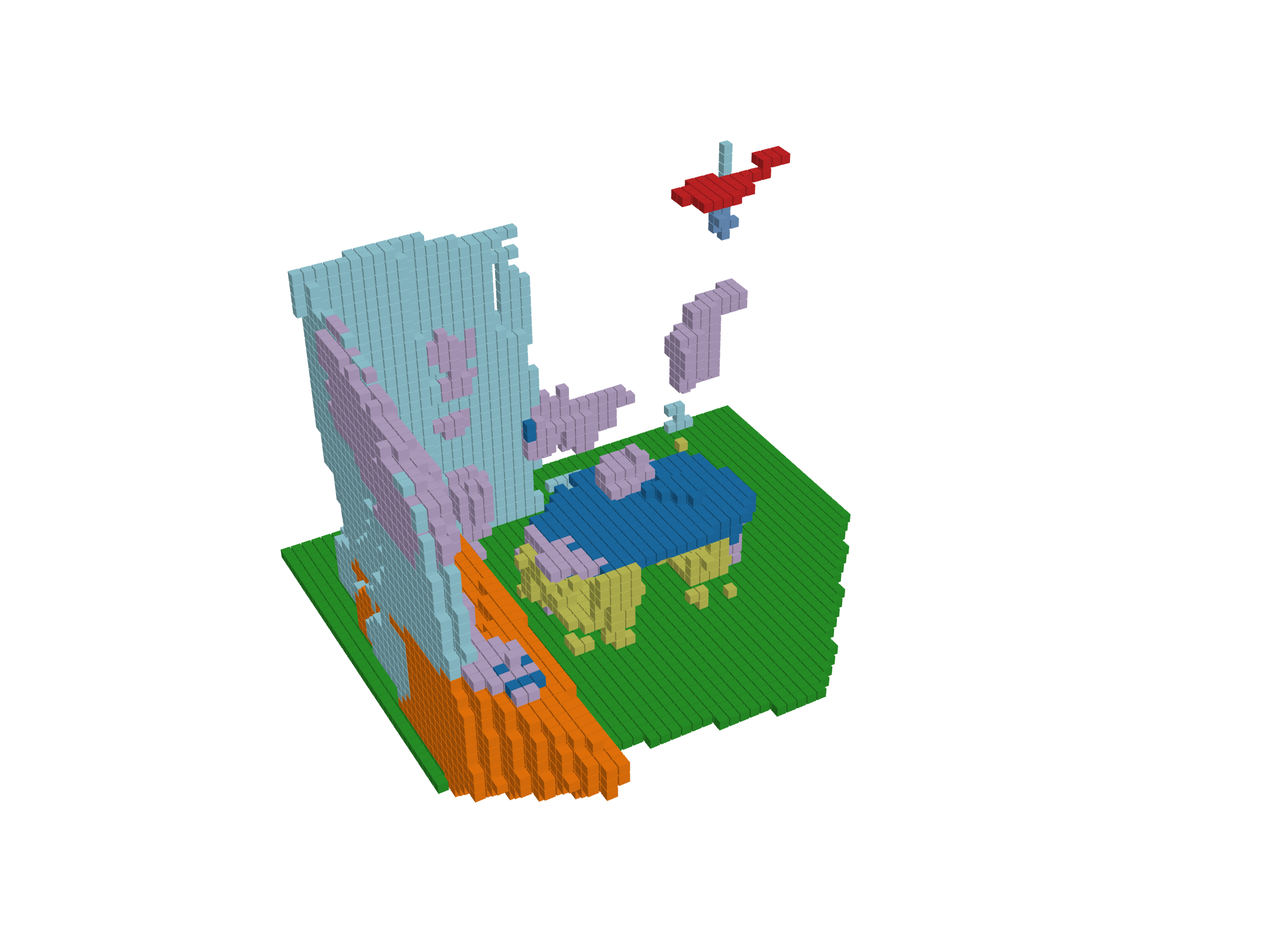}}
\end{minipage}
\begin{minipage}{0.23\linewidth}
    \vspace{3pt}
    \centerline{\includegraphics[width=\textwidth, viewport=250 0 2000 1700,clip]{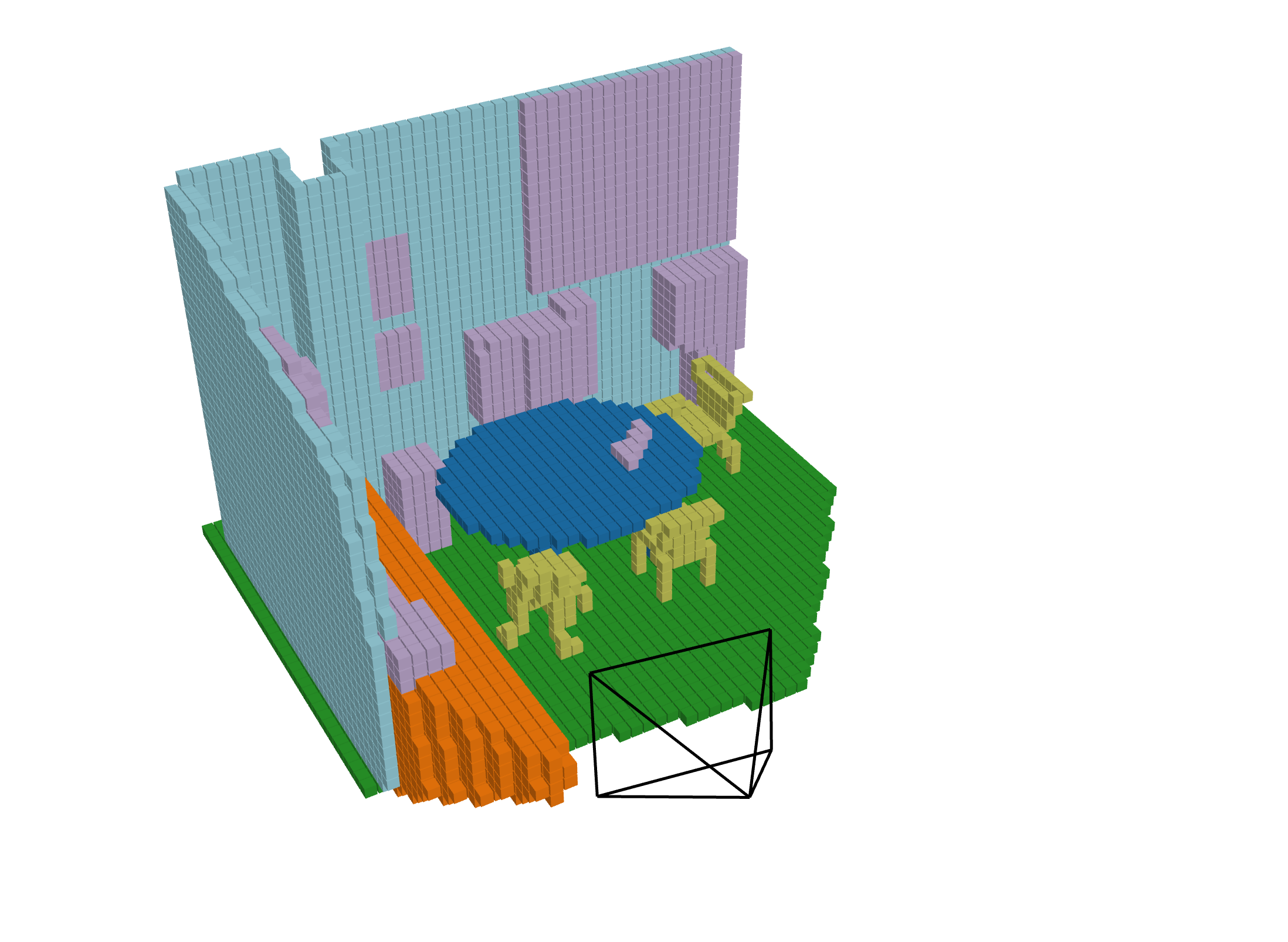}}
\end{minipage}

\begin{minipage}{0.23\linewidth}
    \vspace{3pt}
    \centerline{\includegraphics[width=\textwidth]{figure/nyu_exp/NYU1057_0000_color.jpg}}
\end{minipage}
\begin{minipage}{0.23\linewidth}
    \vspace{3pt}
    \centerline{\includegraphics[width=\textwidth, viewport=500 0 2800 1700,clip]{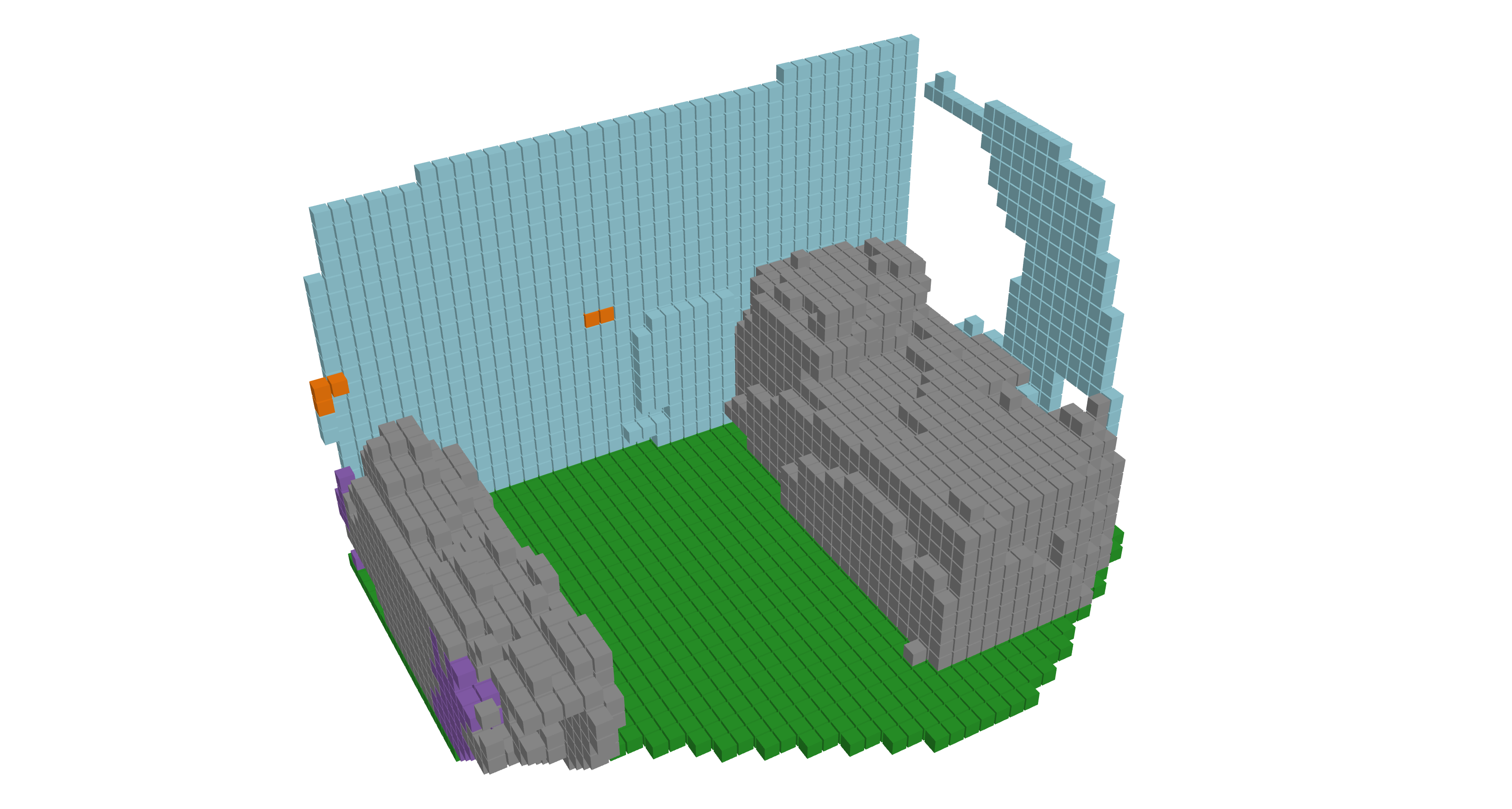}}
\end{minipage}
\begin{minipage}{0.23\linewidth}
    \vspace{3pt}
    \centerline{\includegraphics[width=\textwidth, viewport=500 0 2800 1700,clip]{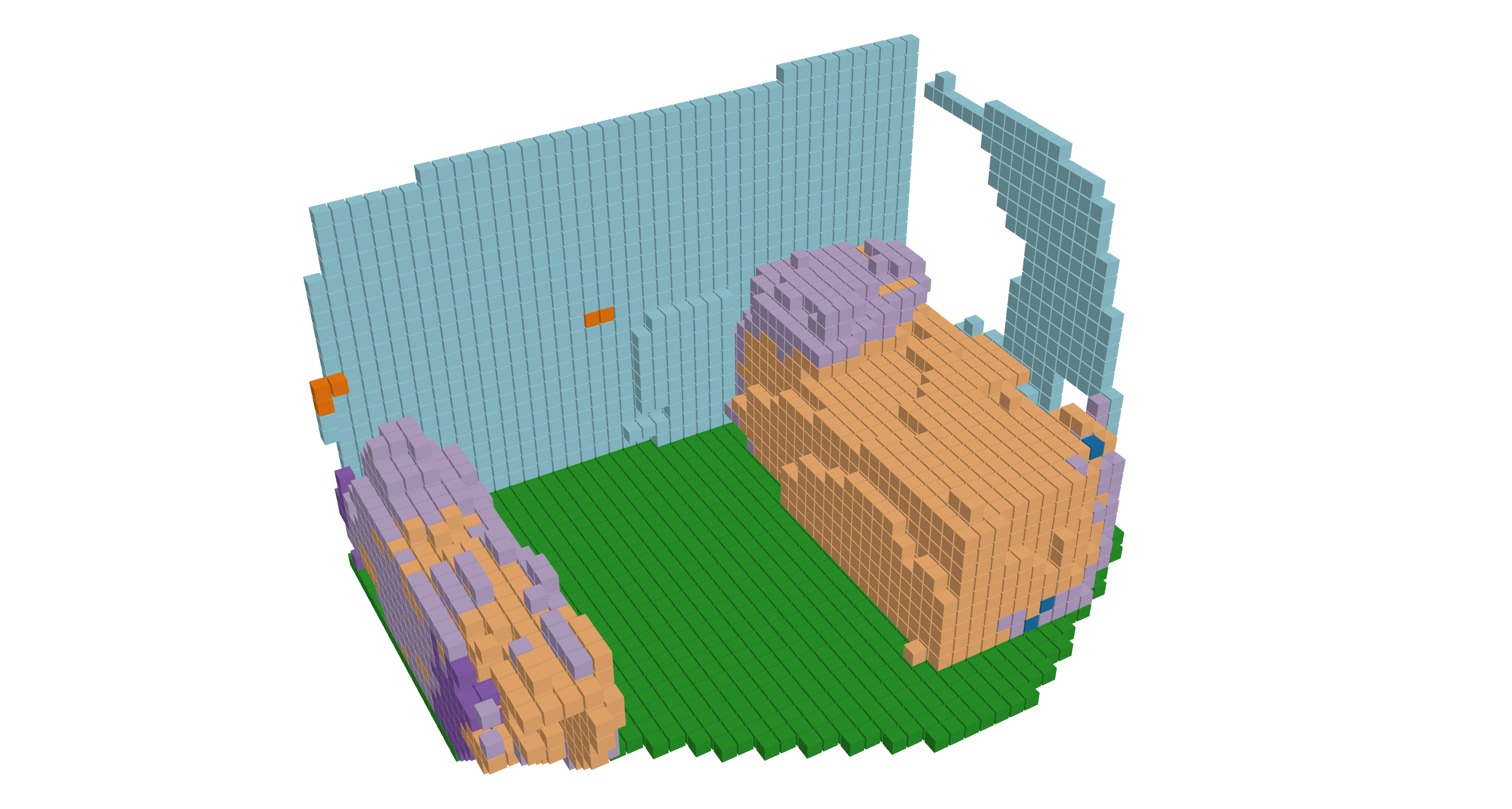}}
\end{minipage}
\begin{minipage}{0.23\linewidth}
    \vspace{3pt}
    \centerline{\includegraphics[width=\textwidth, viewport=500 0 2800 1700,clip]{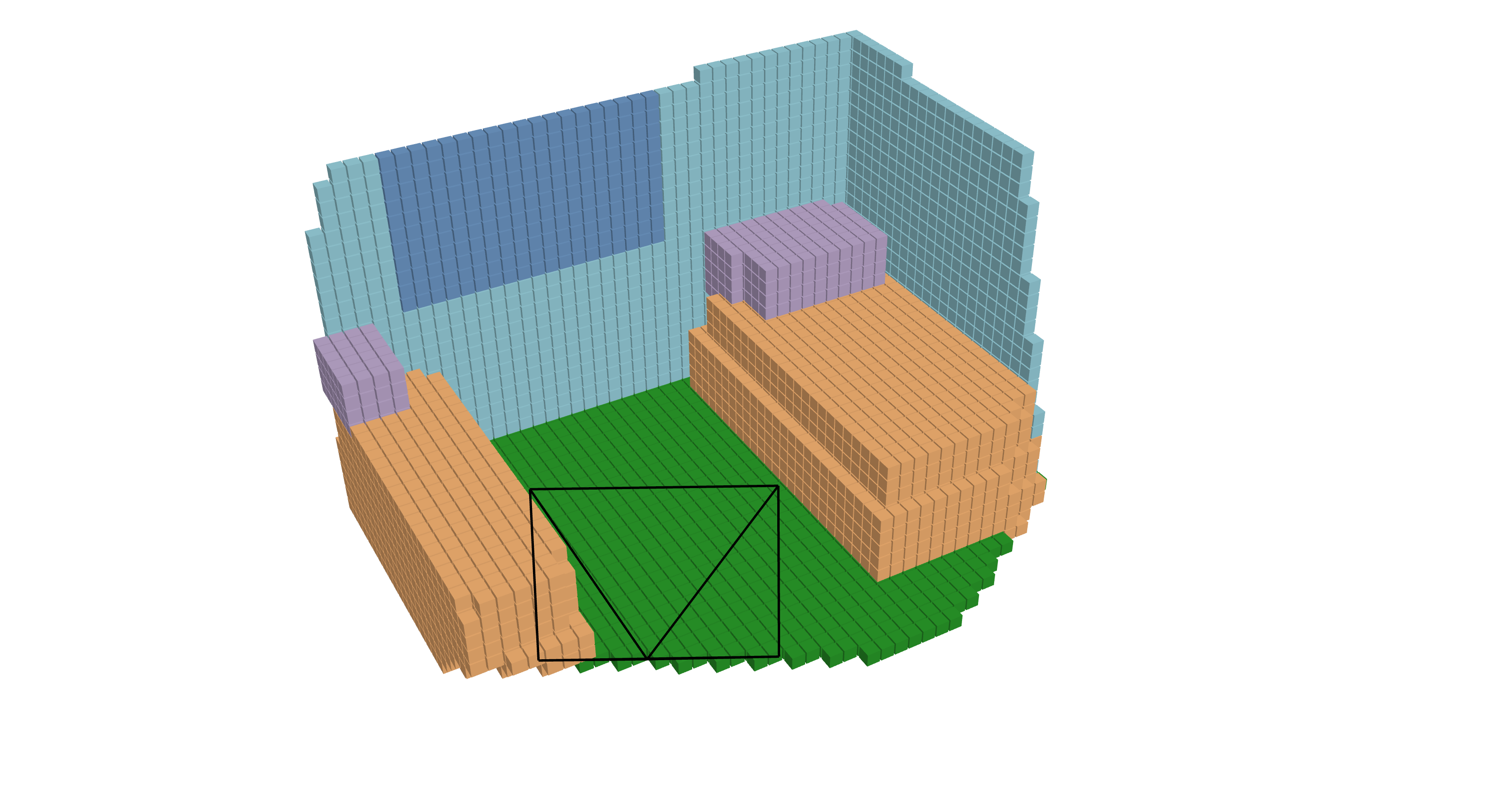}}
\end{minipage}

\begin{minipage}{0.23\linewidth}
    \vspace{3pt}
    \centerline{\includegraphics[width=\textwidth]{figure/app_nyu_exp/NYU0474_0000_color.jpg}}
\end{minipage}
\begin{minipage}{0.23\linewidth}
    \vspace{3pt}
    \centerline{\includegraphics[width=\textwidth, viewport=300 0 2500 1700,clip]{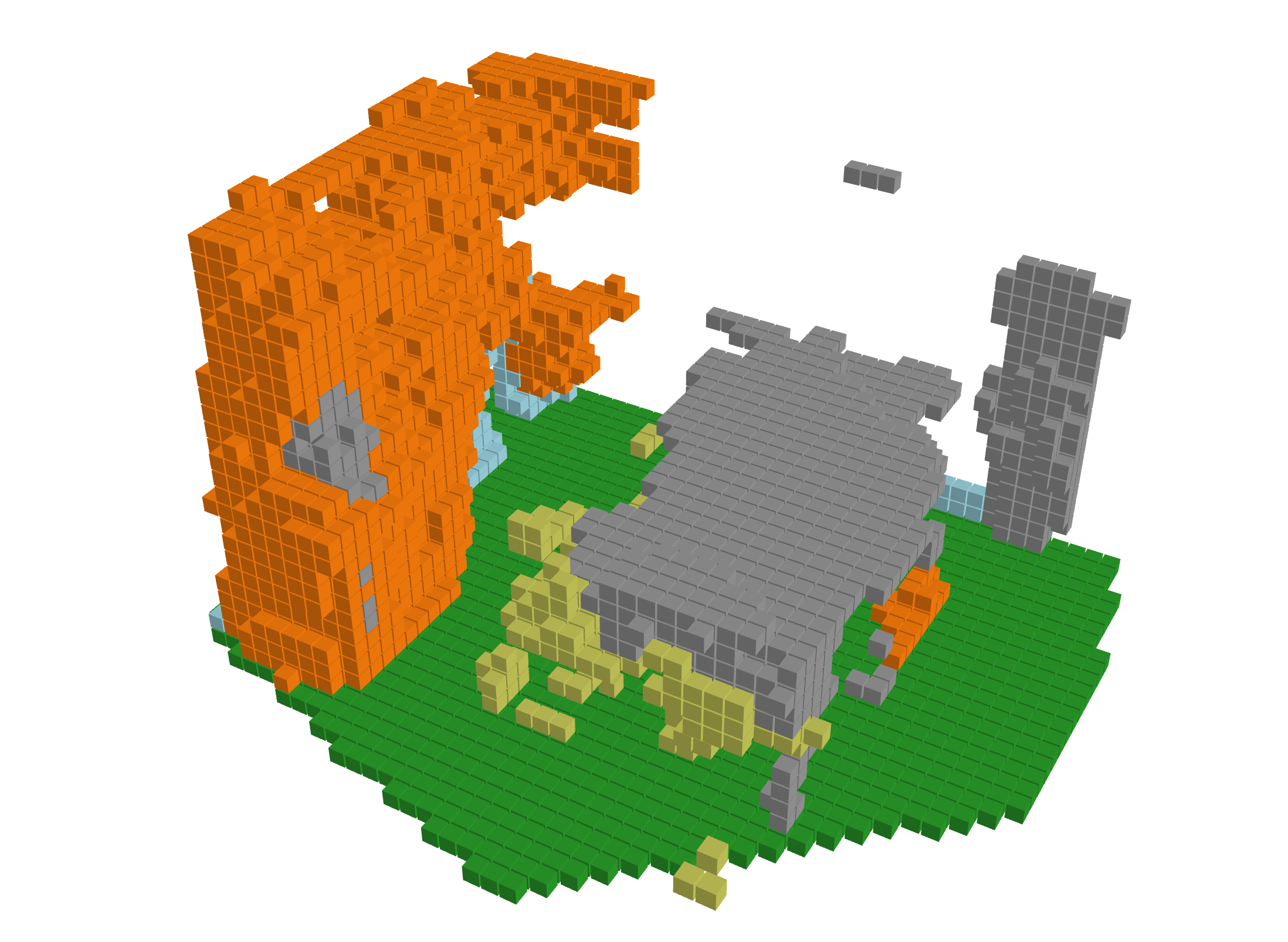}}
\end{minipage}
\begin{minipage}{0.23\linewidth}
    \vspace{3pt}
    \centerline{\includegraphics[width=\textwidth, viewport=300 0 2500 1700,clip]{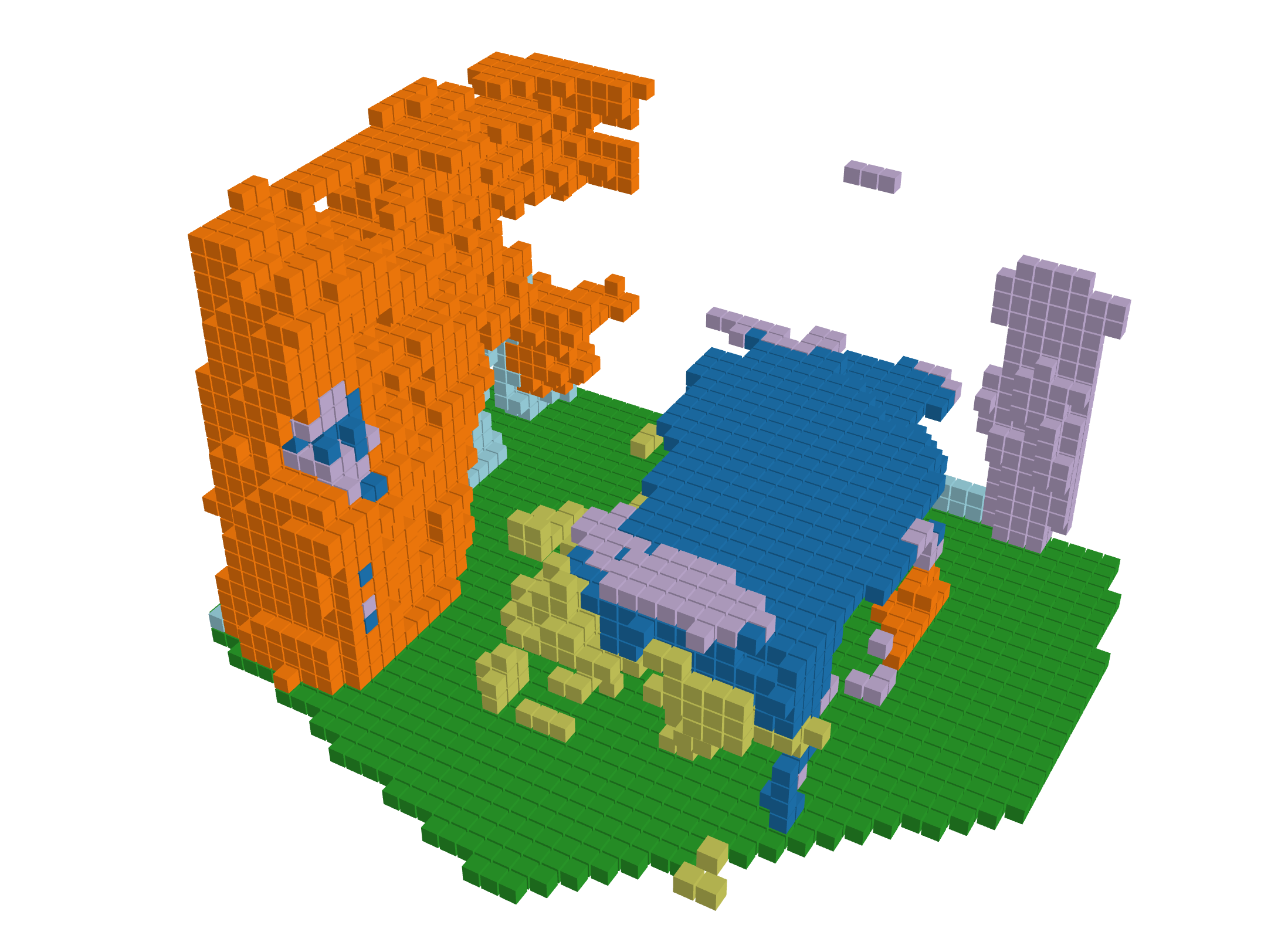}}
\end{minipage}
\begin{minipage}{0.23\linewidth}
    \vspace{3pt}
    \centerline{\includegraphics[width=\textwidth, viewport=250 0 2300 1700,clip]{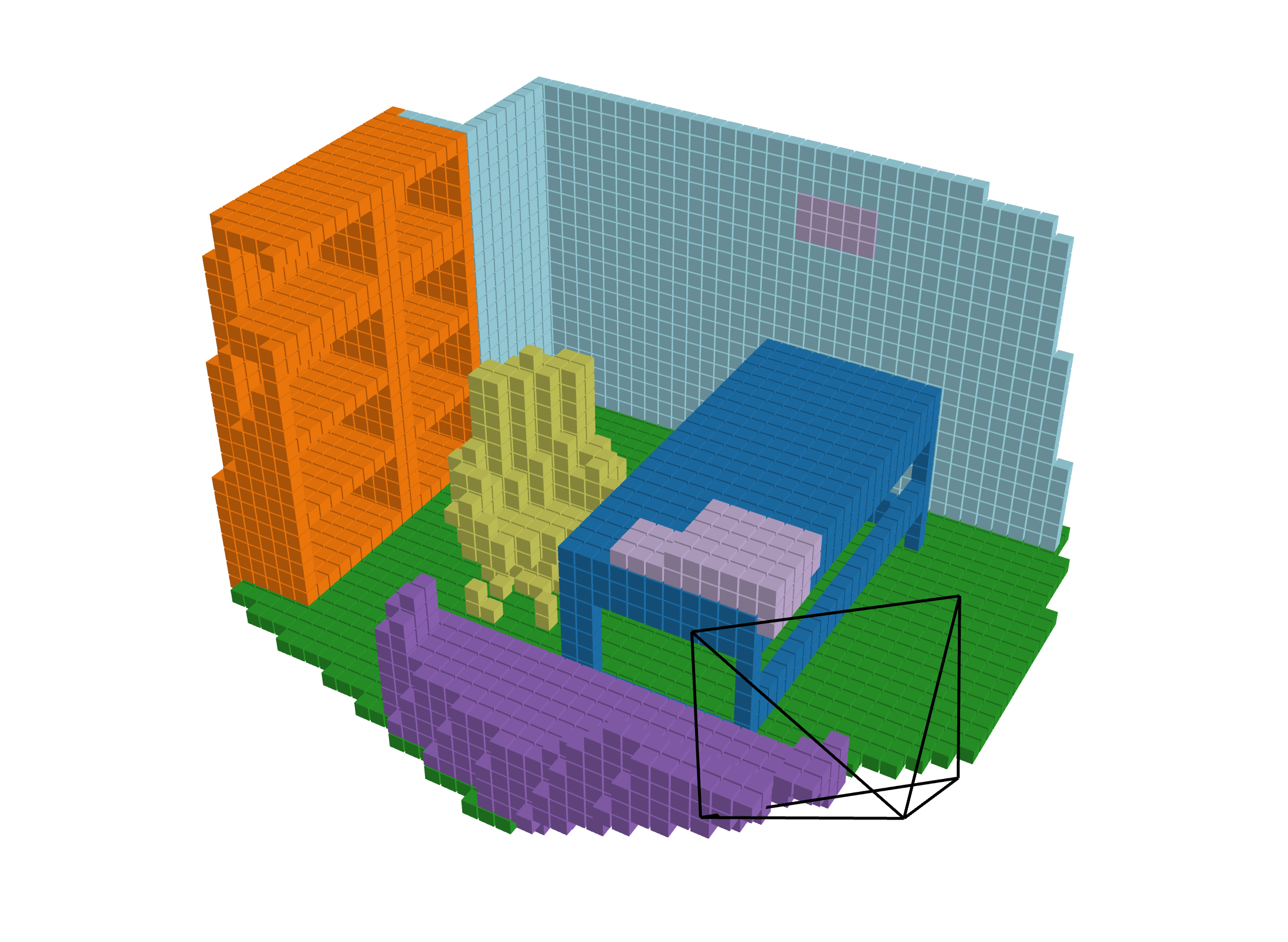}}
\end{minipage}

\begin{minipage}{0.23\linewidth}
    \vspace{3pt}
    \centerline{\includegraphics[width=\textwidth]{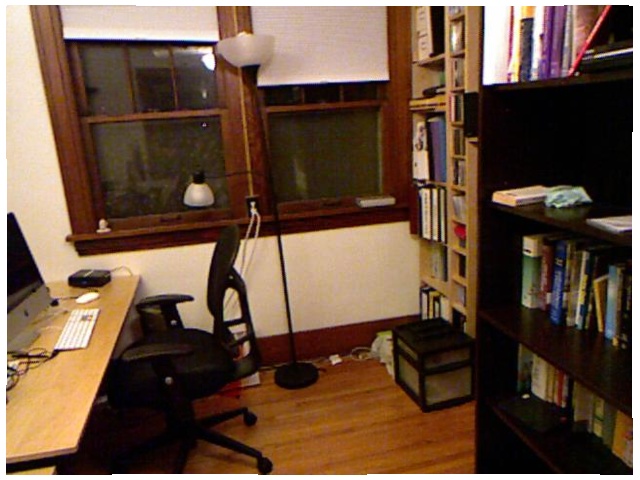}}
\end{minipage}
\begin{minipage}{0.23\linewidth}
    \vspace{3pt}
    \centerline{\includegraphics[width=\textwidth, viewport=800 0 2500 1700,clip]{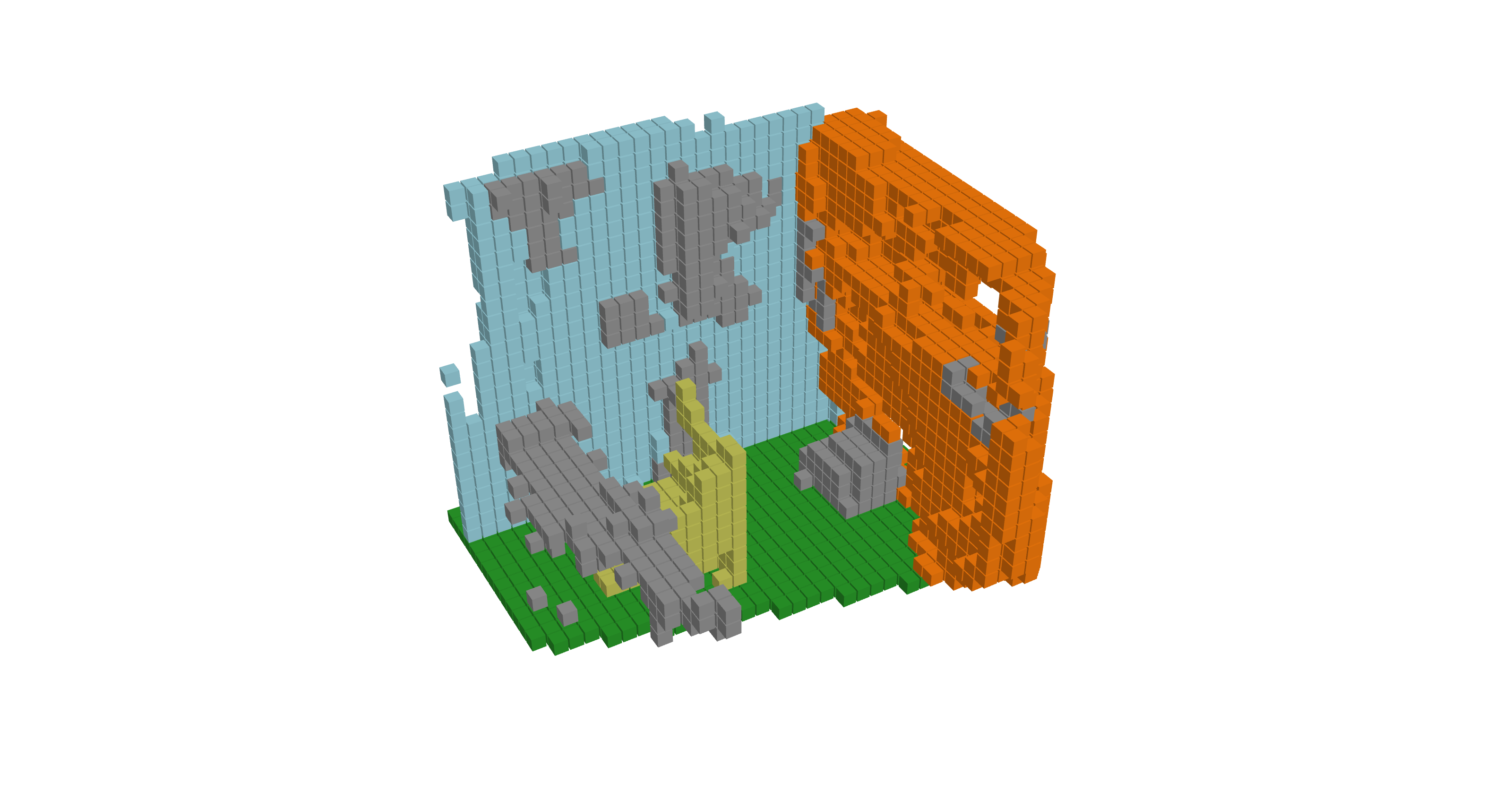}}
\end{minipage}
\begin{minipage}{0.23\linewidth}
    \vspace{3pt}
    \centerline{\includegraphics[width=\textwidth, viewport=800 0 2500 1700,clip]{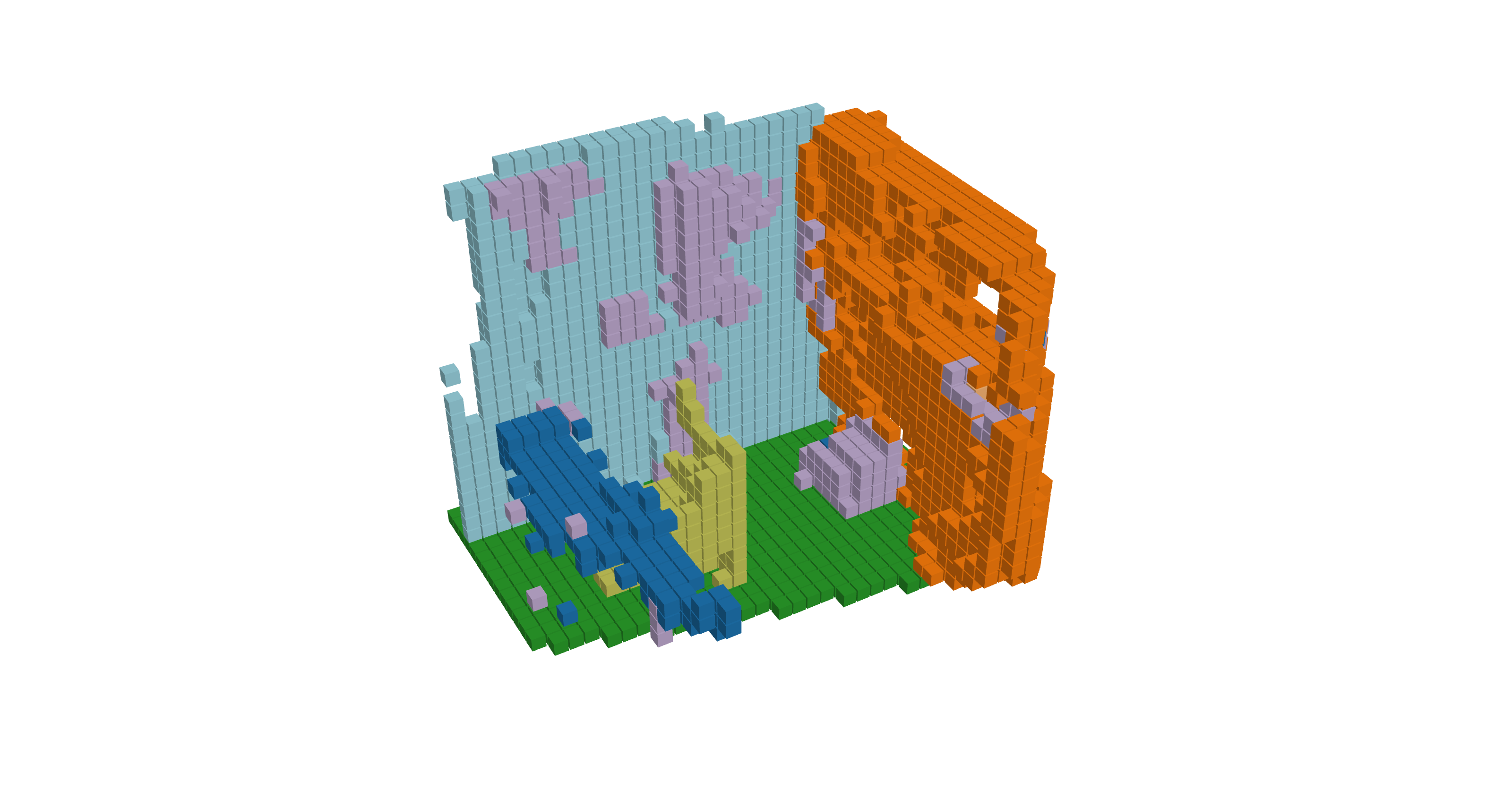}}
\end{minipage}
\begin{minipage}{0.23\linewidth}
    \vspace{3pt}
    \centerline{\includegraphics[width=\textwidth, viewport=600 0 2500 1700,clip]{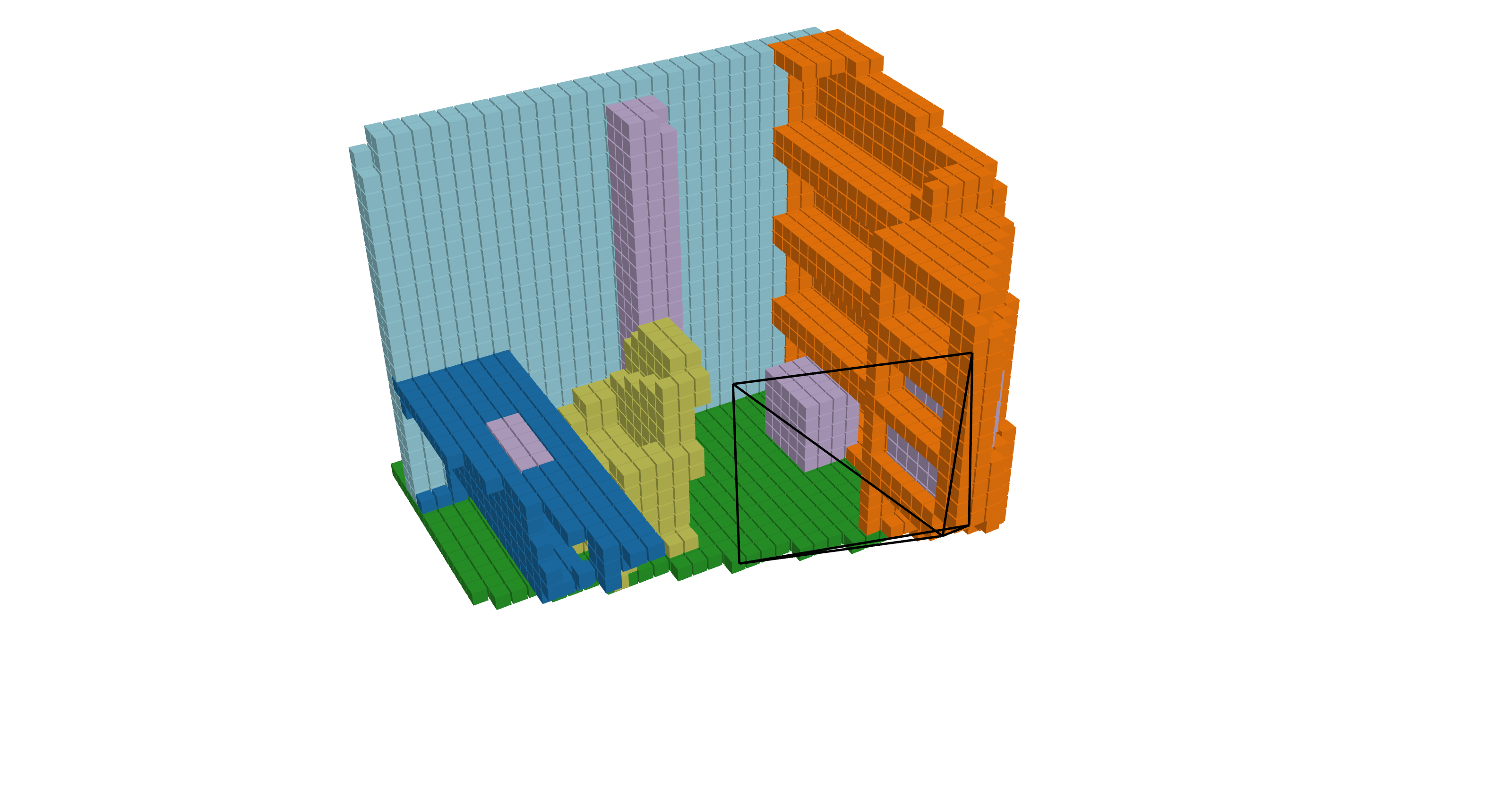}}
\end{minipage}

\begin{minipage}{0.23\linewidth}
    \centering
    RGB input
\end{minipage}
\begin{minipage}{0.23\linewidth}
    \centering
    MonoScene
\end{minipage}
\begin{minipage}{0.23\linewidth}
    \centering
    OVO (ours)
\end{minipage}
\begin{minipage}{0.23\linewidth}
    \centering
    GT
\end{minipage}

\begin{minipage}{\linewidth}
    \centering
    \raisebox{0.5ex}{\colorbox{ceiling}{}}~ceiling \raisebox{0.5ex}{\colorbox{floor}{}}~floor \raisebox{0.5ex}{\colorbox{wall}{}}~wall \raisebox{0.5ex}{\colorbox{window}{}}~window \raisebox{0.5ex}{\colorbox{chair}{}}~chair \raisebox{0.5ex}{\colorbox{bed}{}}~bed \raisebox{0.5ex}{\colorbox{sofa}{}}~sofa \raisebox{0.5ex}{\colorbox{table}{}}~table \raisebox{0.5ex}{\colorbox{tvs}{}}~tvs \raisebox{0.5ex}{\colorbox{furniture}{}}~furniture \raisebox{0.5ex}{\colorbox{other}{}}~other \raisebox{0.5ex}{\colorbox{unknown}{}}~unknown
\end{minipage}

\caption{\small \textbf{Qualitative visualization on NYUv2 dataset}~\cite{nyu}. The novel classes for NYUv2 dataset include  {\color{bed} “bed”}, {\color{table} “table”}, and {\color{other}“other”}. The gray voxels in the second column represent the instances of these novel classes that cannot be predicted by the vanilla MonoScene trained with supervised data. In the third column, gray voxels are painted according to the inference results of our OVO.}
\label{fig:app_nyu_qualitative}
\end{figure}

%% file: table/app_qual_kitti.tex
\begin{figure*}[htbp]
\centering     

\begin{minipage}{0.23\linewidth}
    \vspace{3pt}
    \centerline{\includegraphics[width=\textwidth]{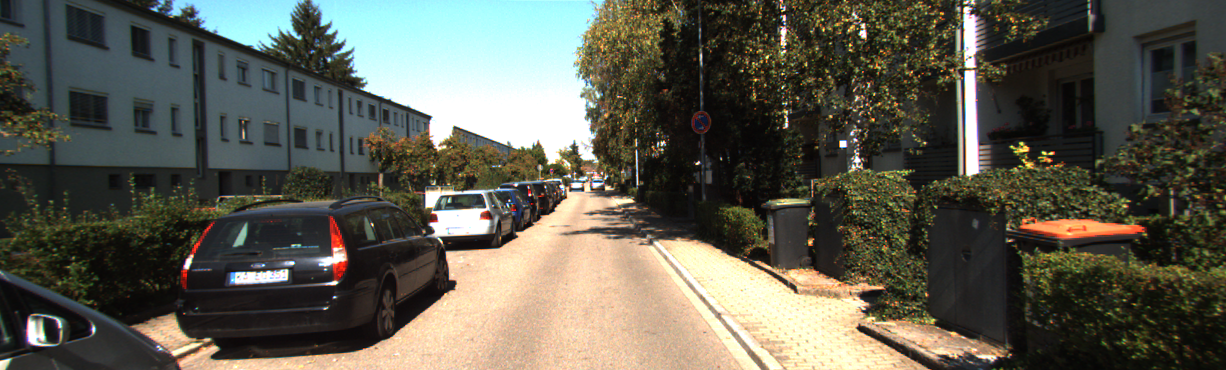}}
\end{minipage}
\begin{minipage}{0.23\linewidth}
    \vspace{3pt}
    \centerline{\includegraphics[width=\textwidth, viewport=200 250 2000 1300,clip]{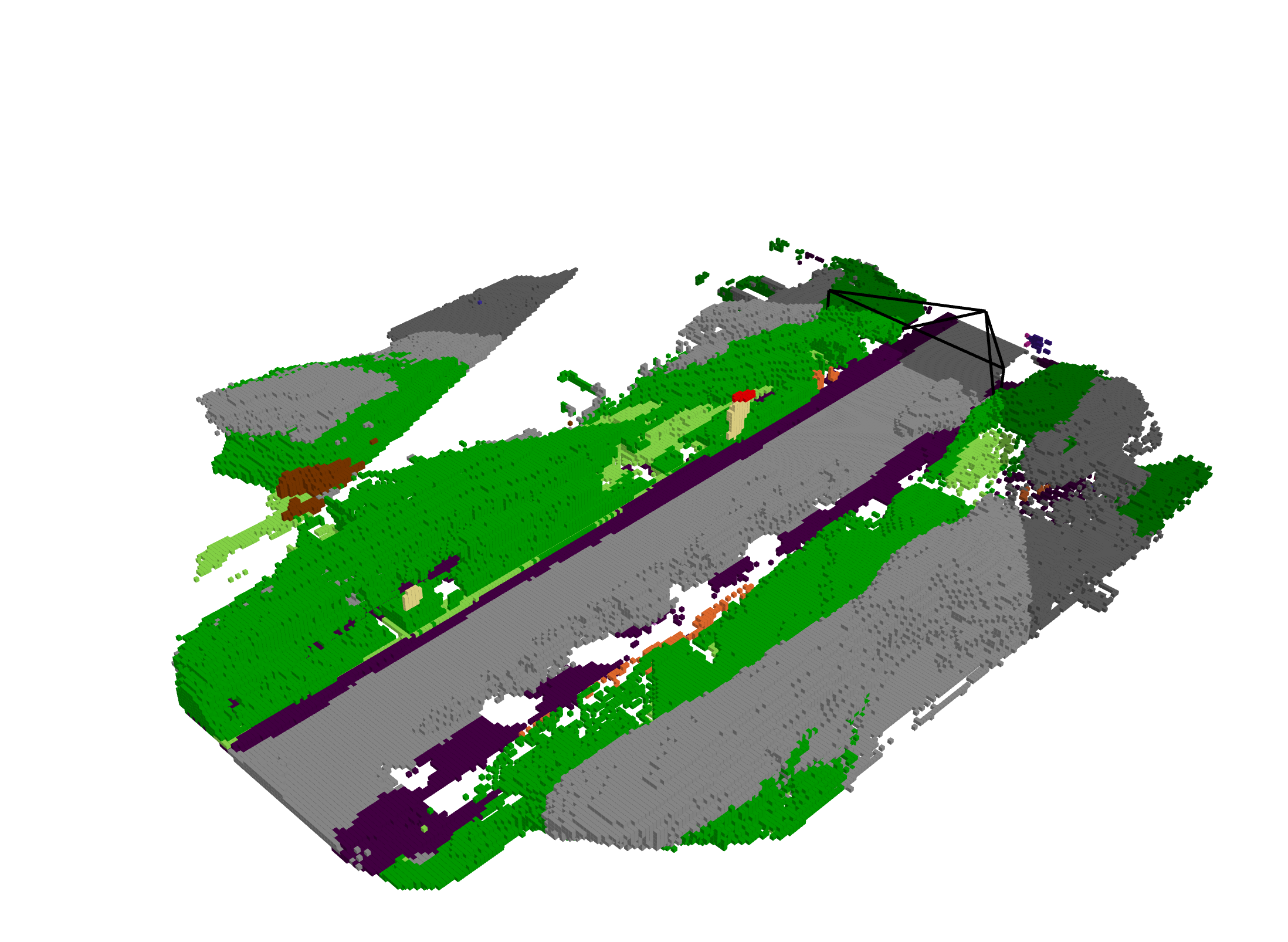}}
\end{minipage}
\begin{minipage}{0.23\linewidth}
    \vspace{3pt}
    \centerline{\includegraphics[width=\textwidth, viewport=200 250 2000 1300,clip]{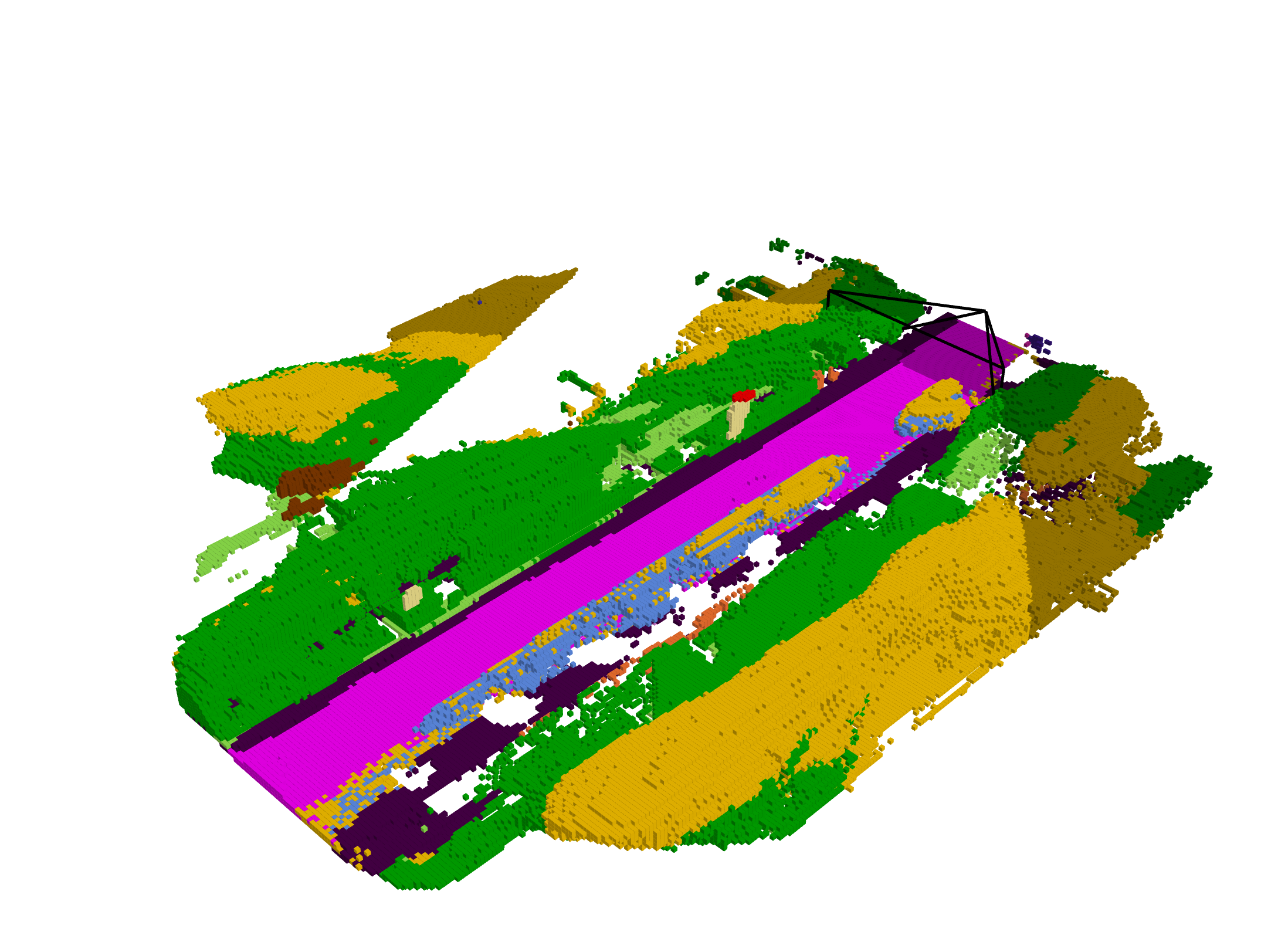}}
\end{minipage}
\begin{minipage}{0.23\linewidth}
    \vspace{3pt}
    \centerline{\includegraphics[width=\textwidth, viewport=200 250 2000 1300,clip]{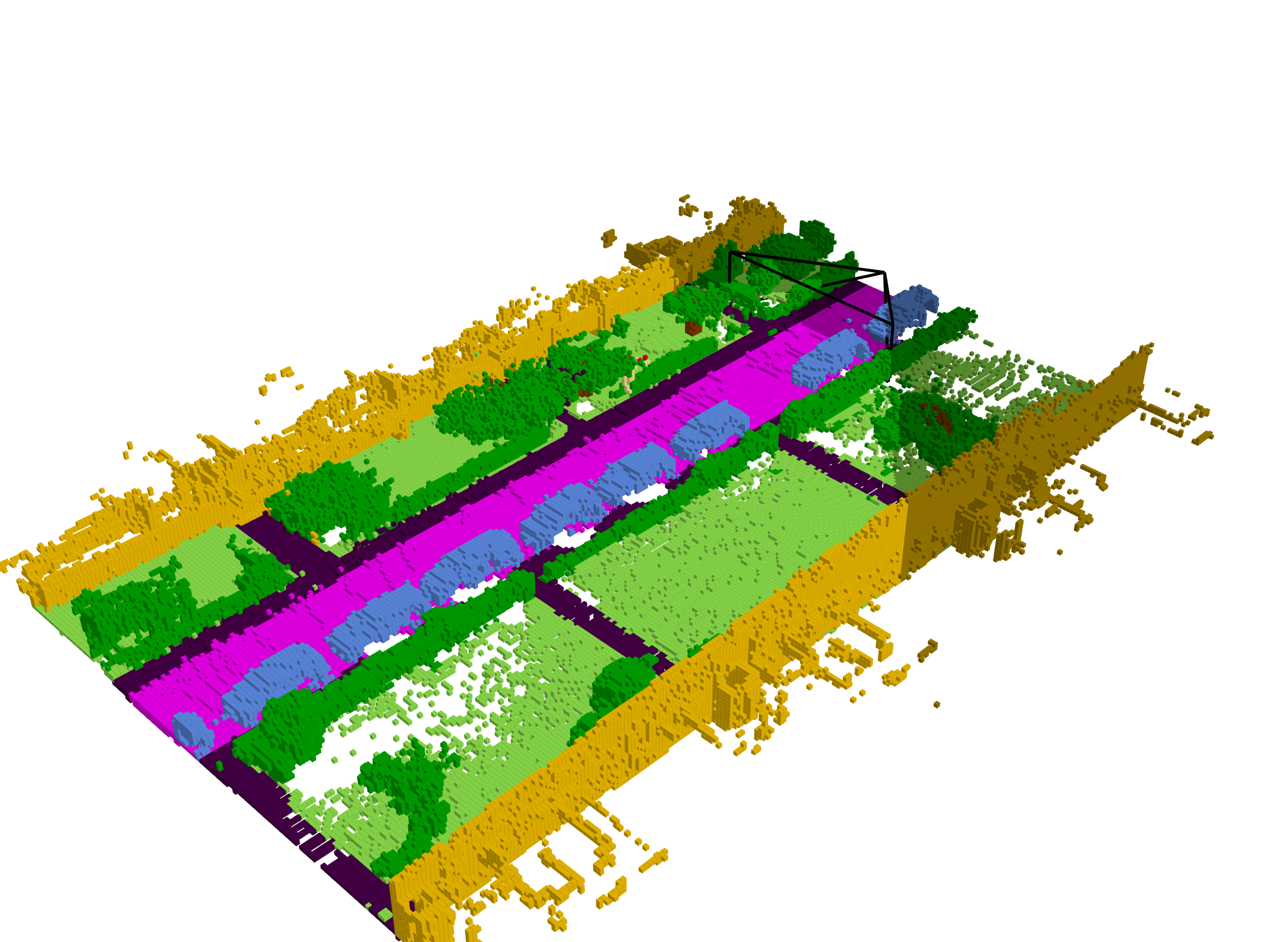}}
\end{minipage}

\begin{minipage}{0.23\linewidth}
    \vspace{3pt}
    \centerline{\includegraphics[width=\textwidth]{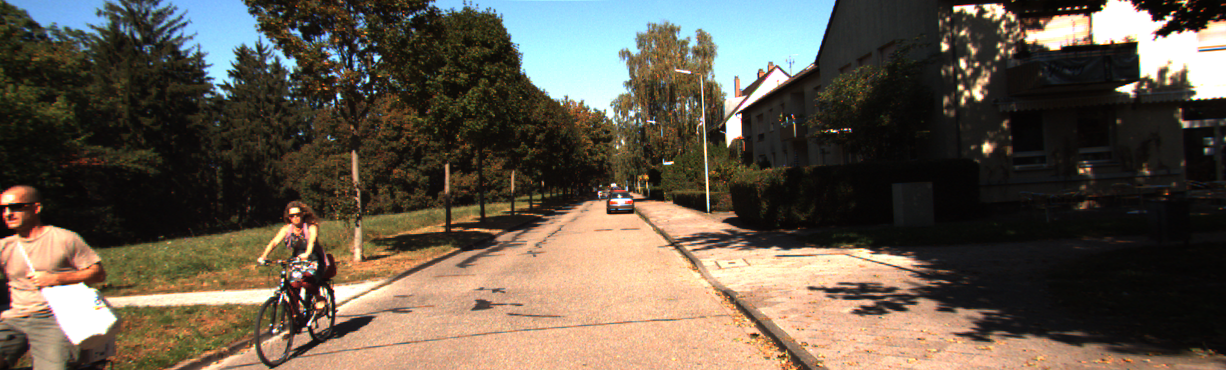}}
\end{minipage}
\begin{minipage}{0.23\linewidth}
    \vspace{3pt}
    \centerline{\includegraphics[width=\textwidth, viewport=200 250 2000 1300,clip]{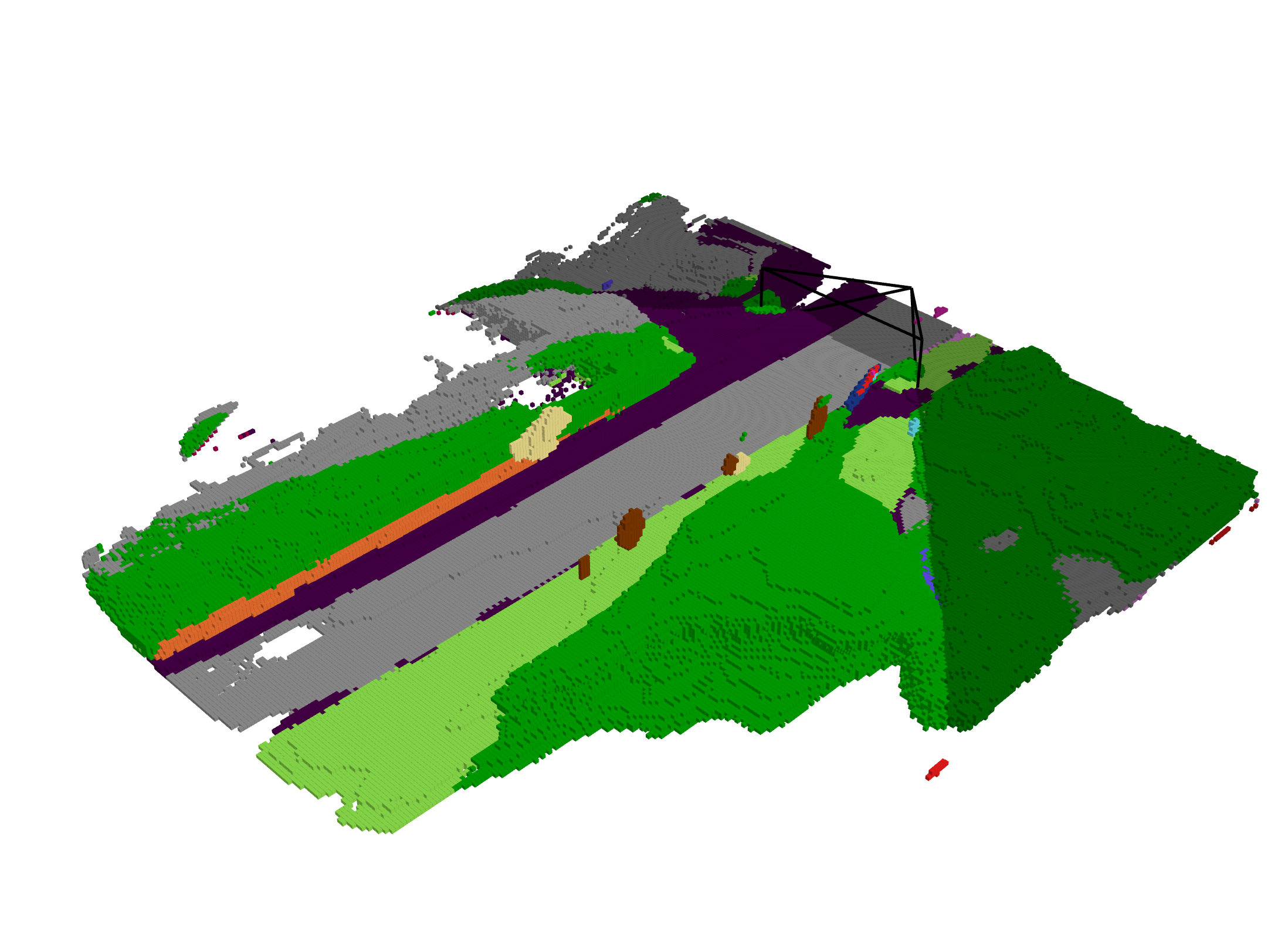}}
\end{minipage}
\begin{minipage}{0.23\linewidth}
    \vspace{3pt}
    \centerline{\includegraphics[width=\textwidth, viewport=200 250 2000 1300,clip]{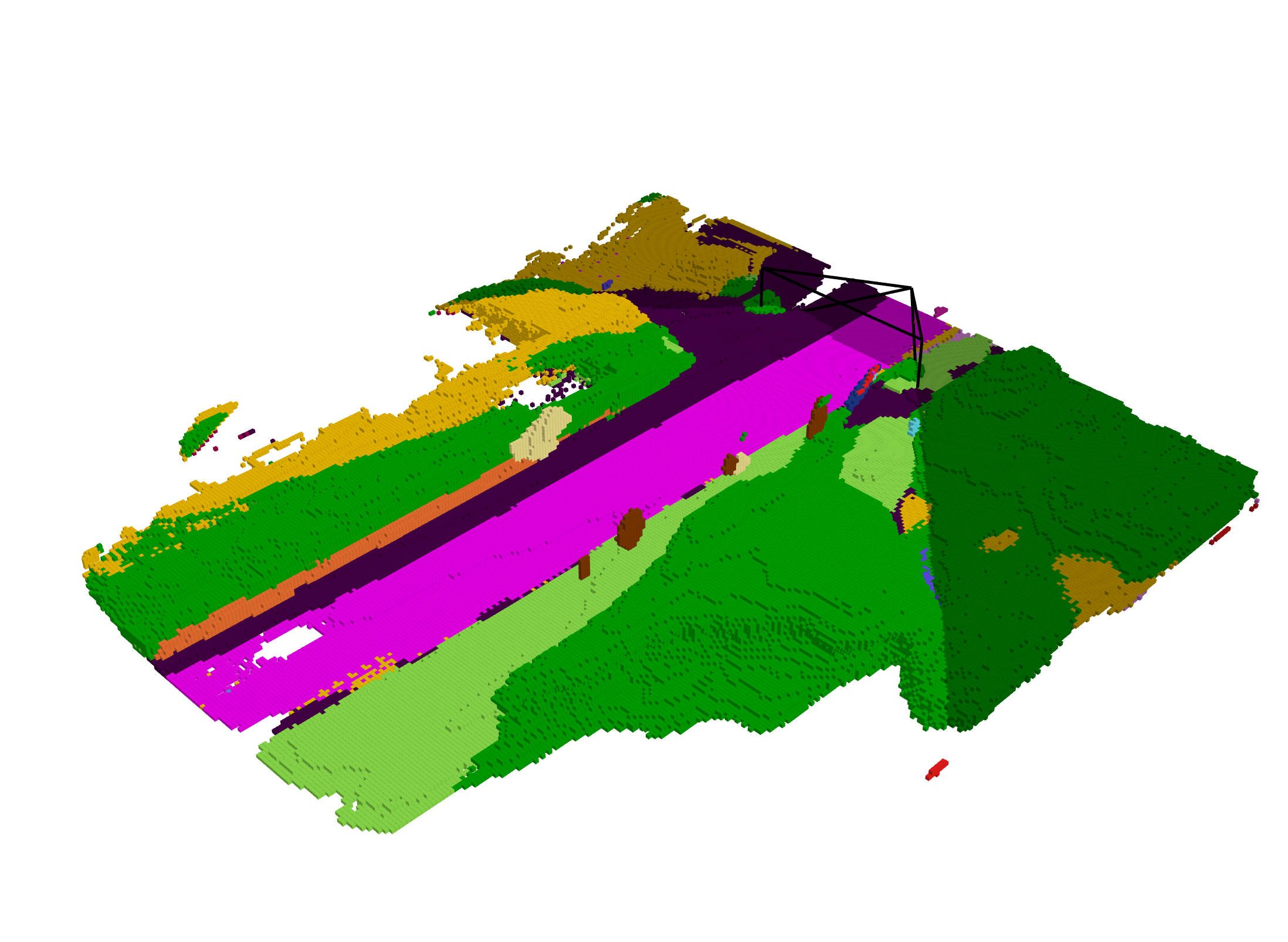}}
\end{minipage}
\begin{minipage}{0.23\linewidth}
    \vspace{3pt}
    \centerline{\includegraphics[width=\textwidth, viewport=200 250 2000 1300,clip]{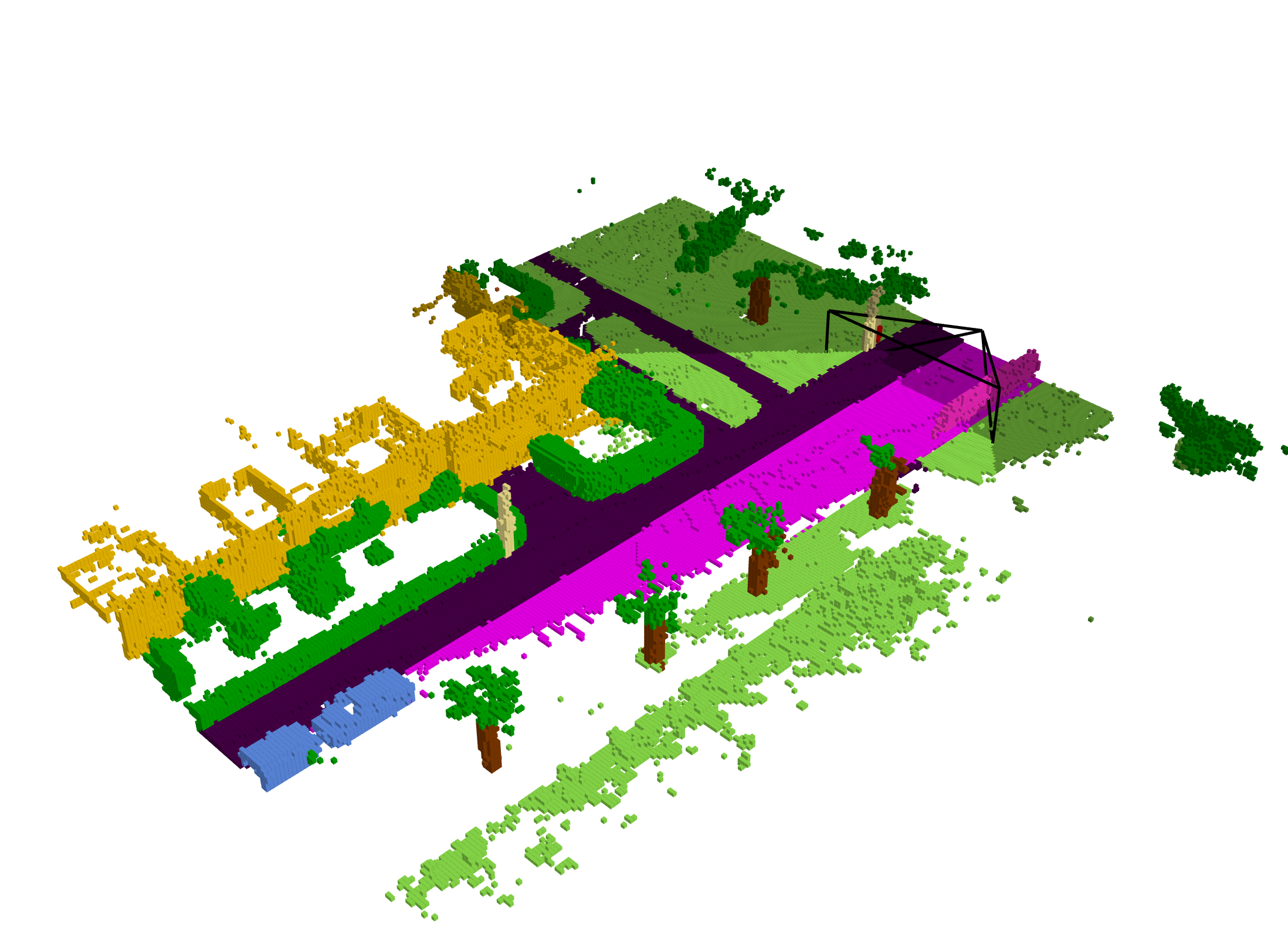}}
\end{minipage}

\begin{minipage}{0.23\linewidth}
    \vspace{3pt}
    \centerline{\includegraphics[width=\textwidth]{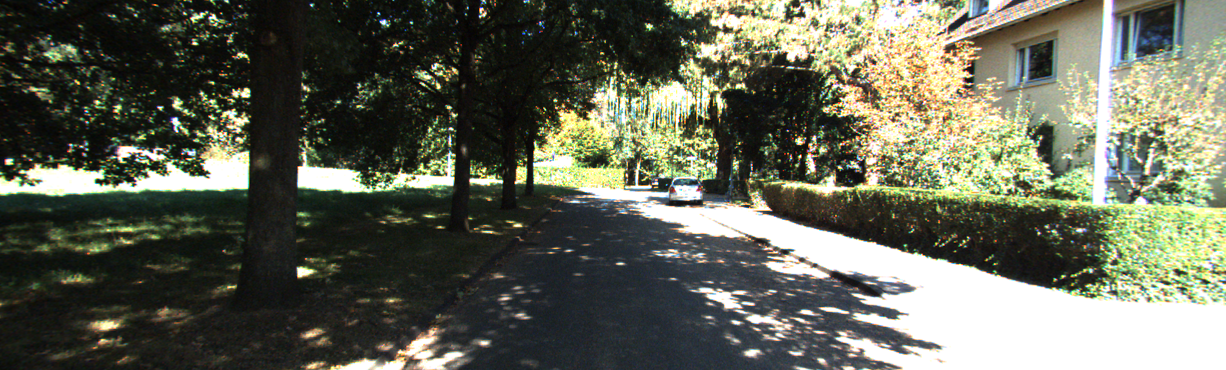}}
\end{minipage}
\begin{minipage}{0.23\linewidth}
    \vspace{3pt}
    \centerline{\includegraphics[width=\textwidth, viewport=200 250 2000 1300,clip]{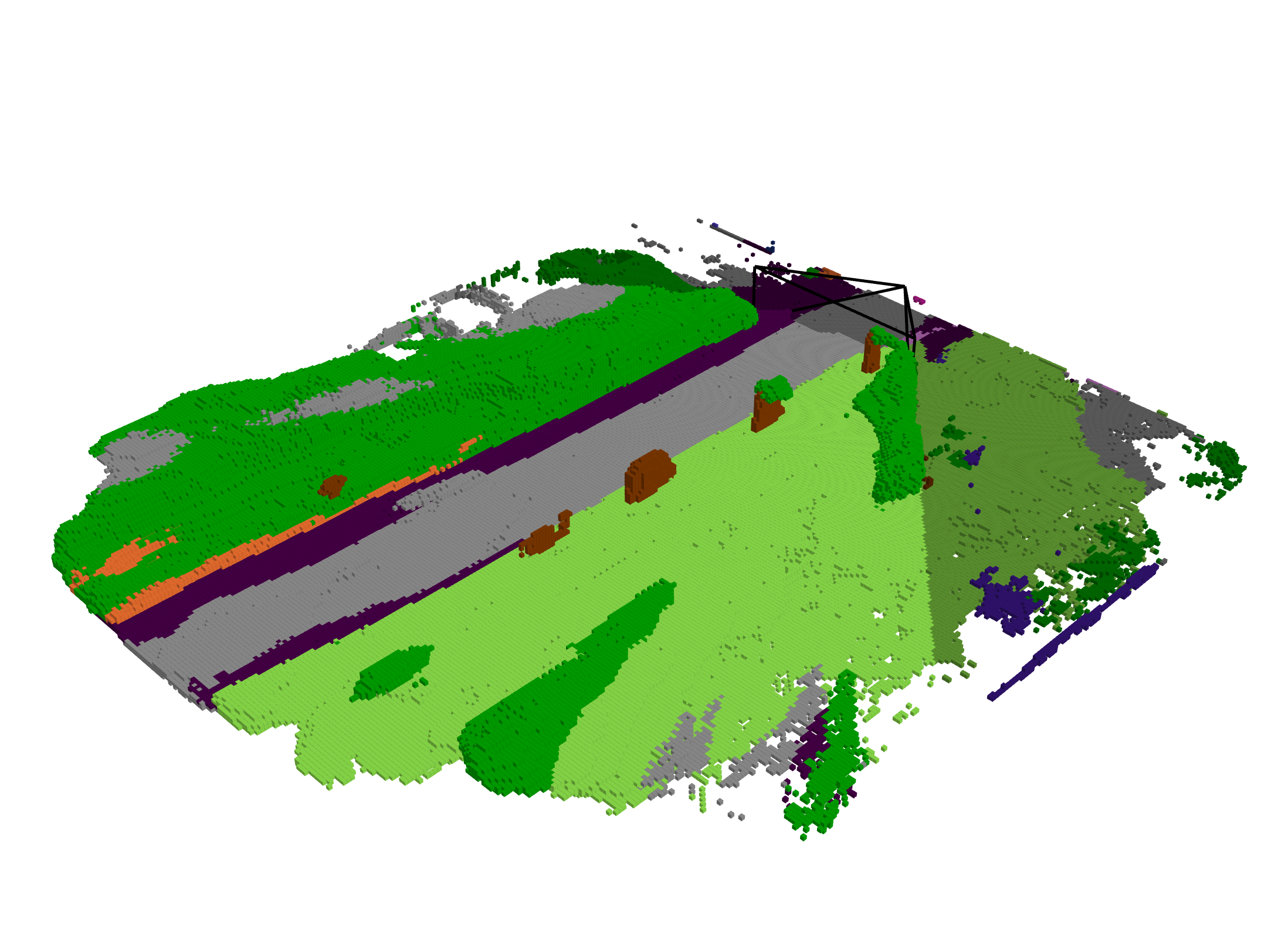}}
\end{minipage}
\begin{minipage}{0.23\linewidth}
    \vspace{3pt}
    \centerline{\includegraphics[width=\textwidth, viewport=200 250 2000 1300,clip]{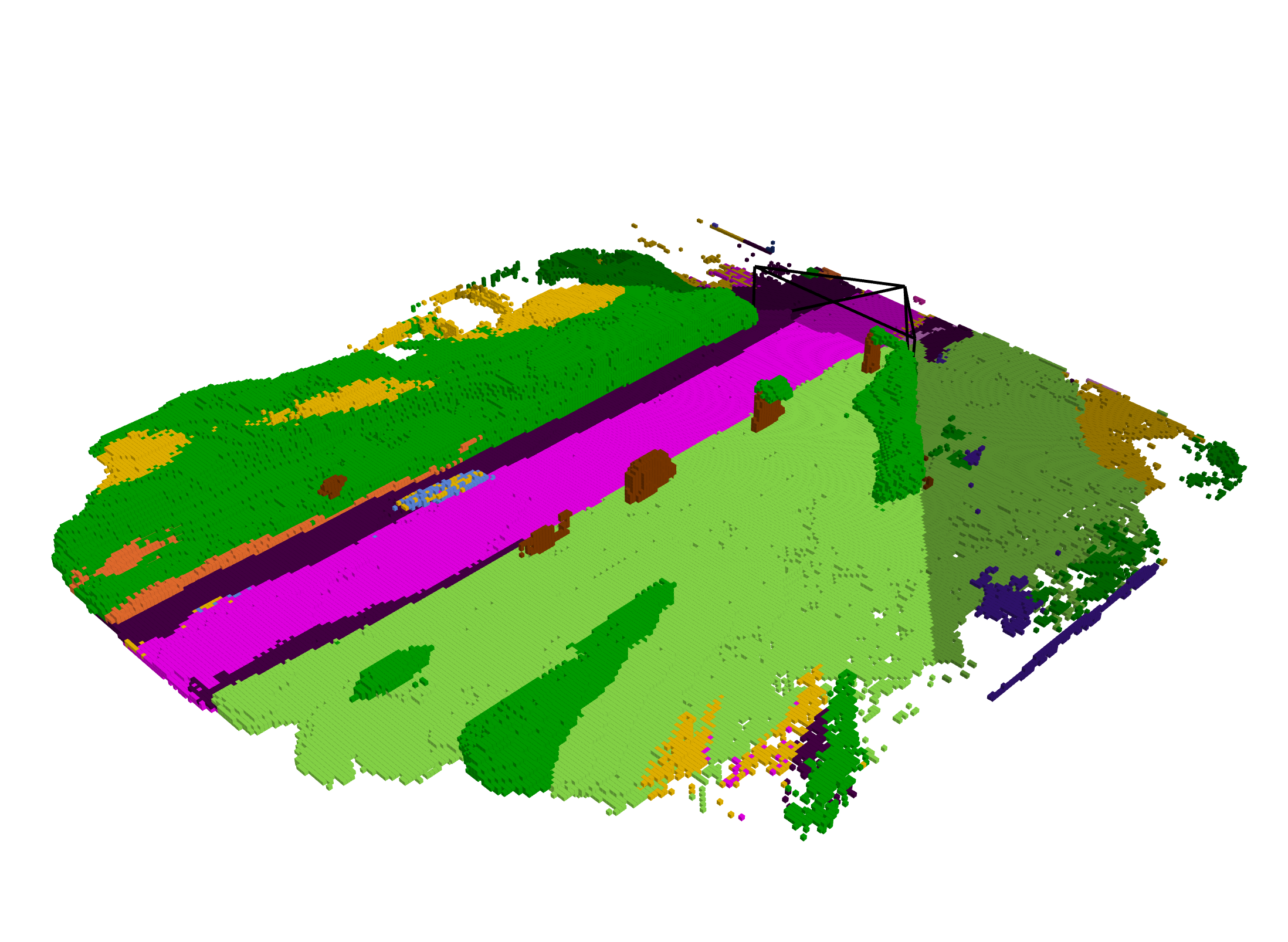}}
\end{minipage}
\begin{minipage}{0.23\linewidth}
    \vspace{3pt}
    \centerline{\includegraphics[width=\textwidth, viewport=200 250 2000 1300,clip]{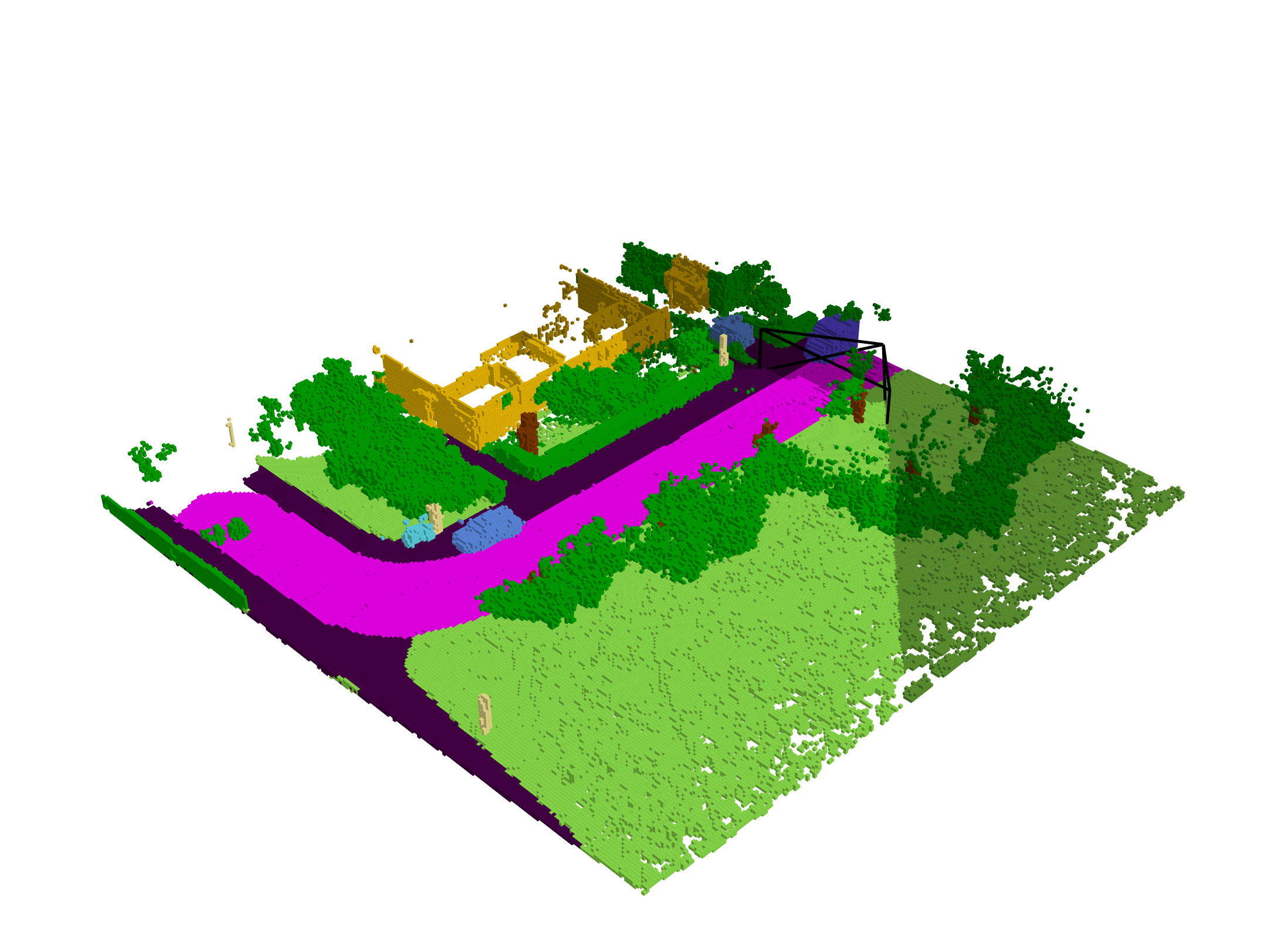}}
\end{minipage}

\begin{minipage}{0.23\linewidth}
    \vspace{3pt}
    \centerline{\includegraphics[width=\textwidth]{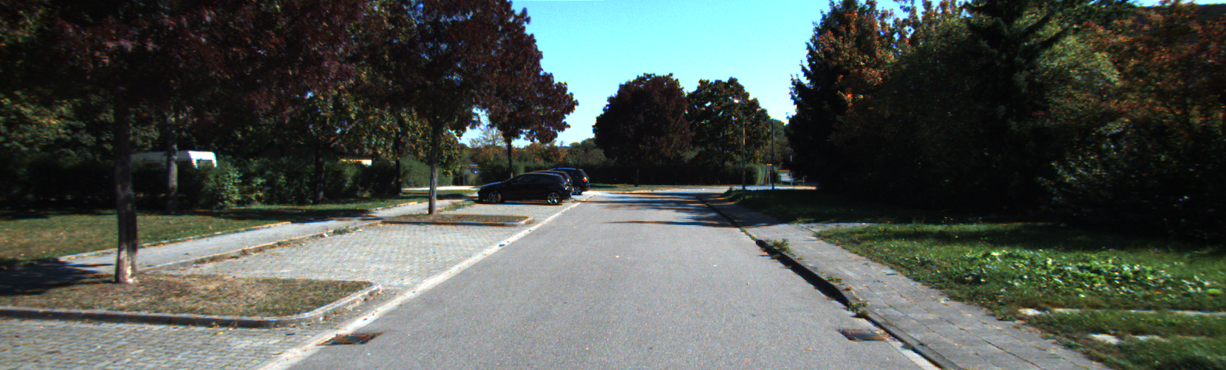}}
\end{minipage}
\begin{minipage}{0.23\linewidth}
    \vspace{3pt}
    \centerline{\includegraphics[width=\textwidth, viewport=200 250 2000 1300,clip]{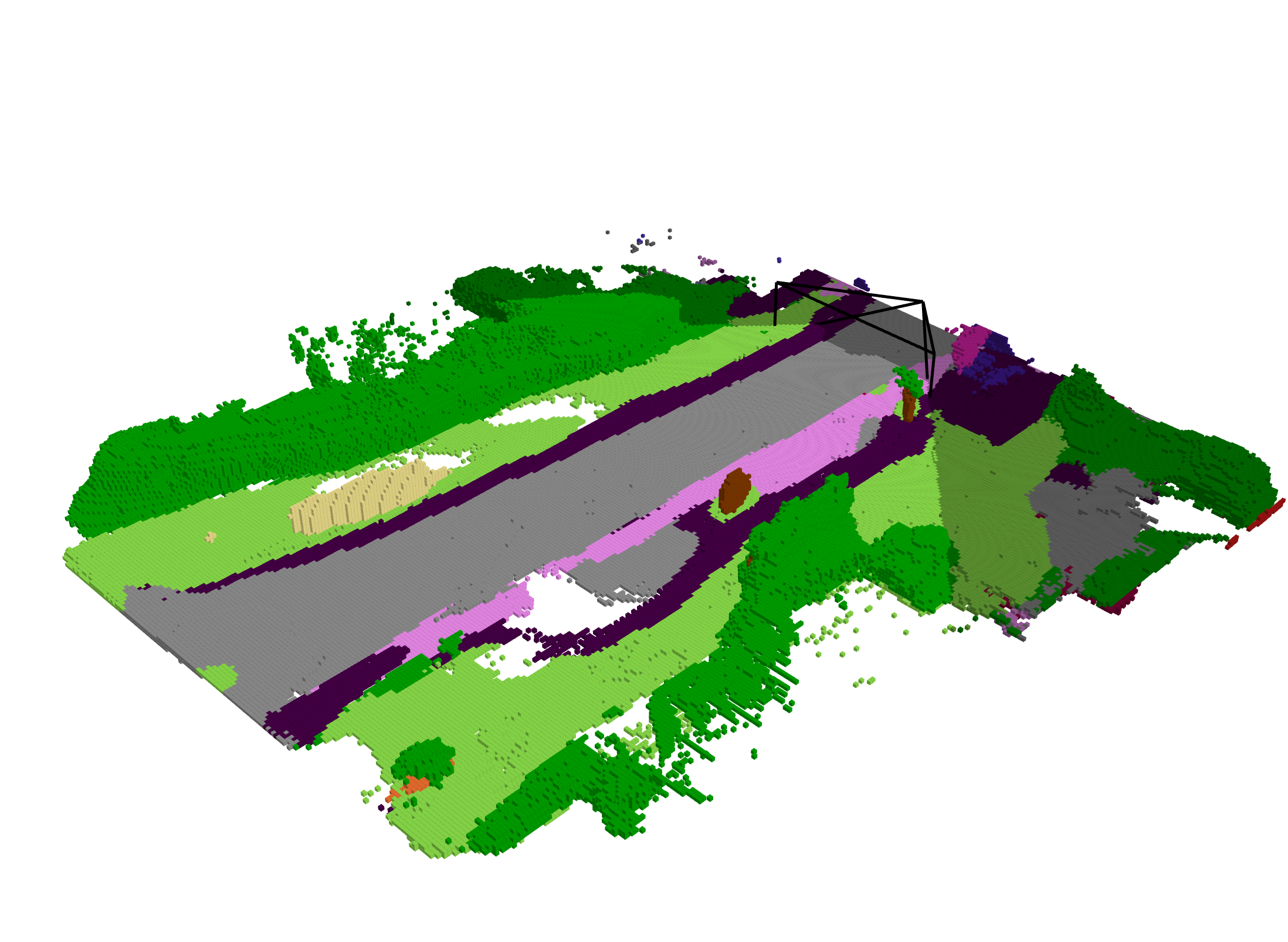}}
\end{minipage}
\begin{minipage}{0.23\linewidth}
    \vspace{3pt}
    \centerline{\includegraphics[width=\textwidth, viewport=200 250 2000 1300,clip]{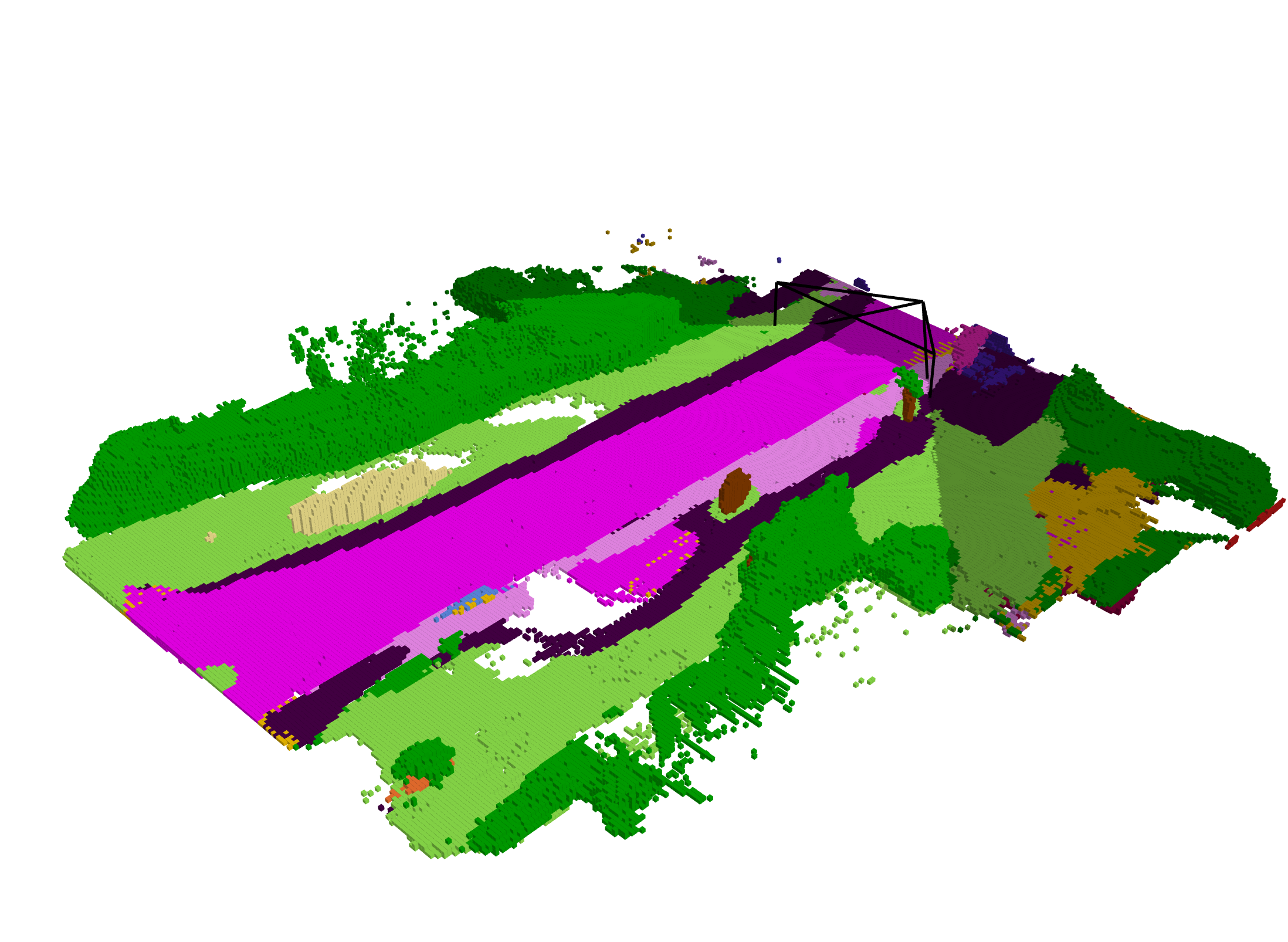}}
\end{minipage}
\begin{minipage}{0.23\linewidth}
    \vspace{3pt}
    \centerline{\includegraphics[width=\textwidth, viewport=200 250 2000 1300,clip]{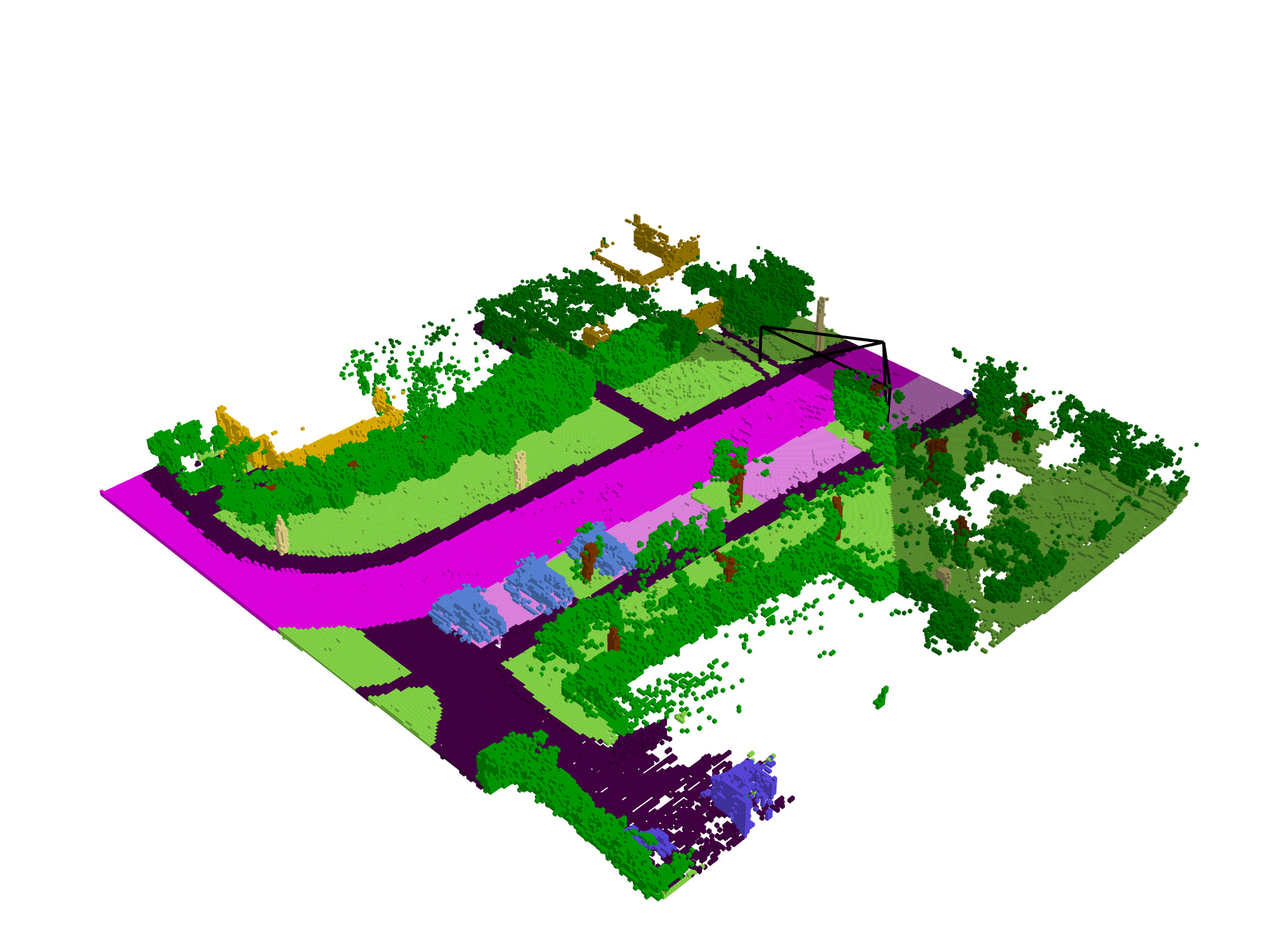}}
\end{minipage}

\begin{minipage}{0.23\linewidth}
    \vspace{3pt}
    \centerline{\includegraphics[width=\textwidth]{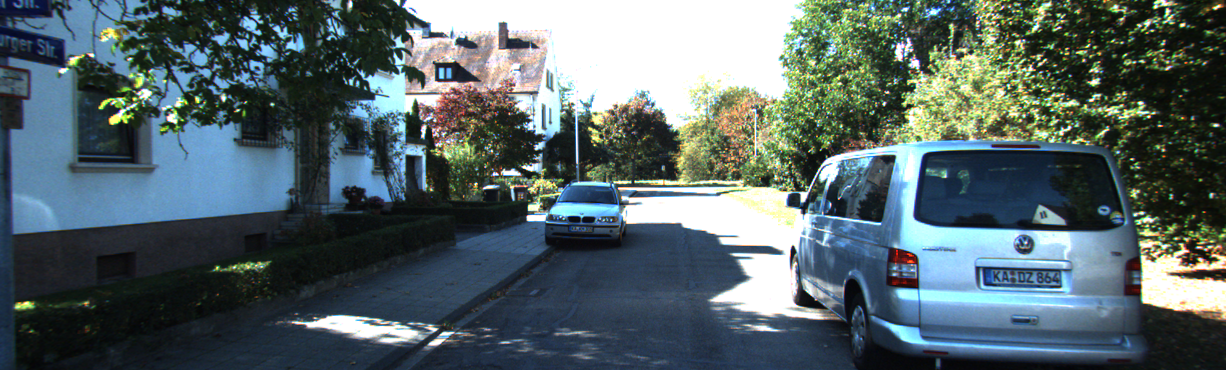}}
\end{minipage}
\begin{minipage}{0.23\linewidth}
    \vspace{3pt}
    \centerline{\includegraphics[width=\textwidth, viewport=200 250 2000 1300,clip]{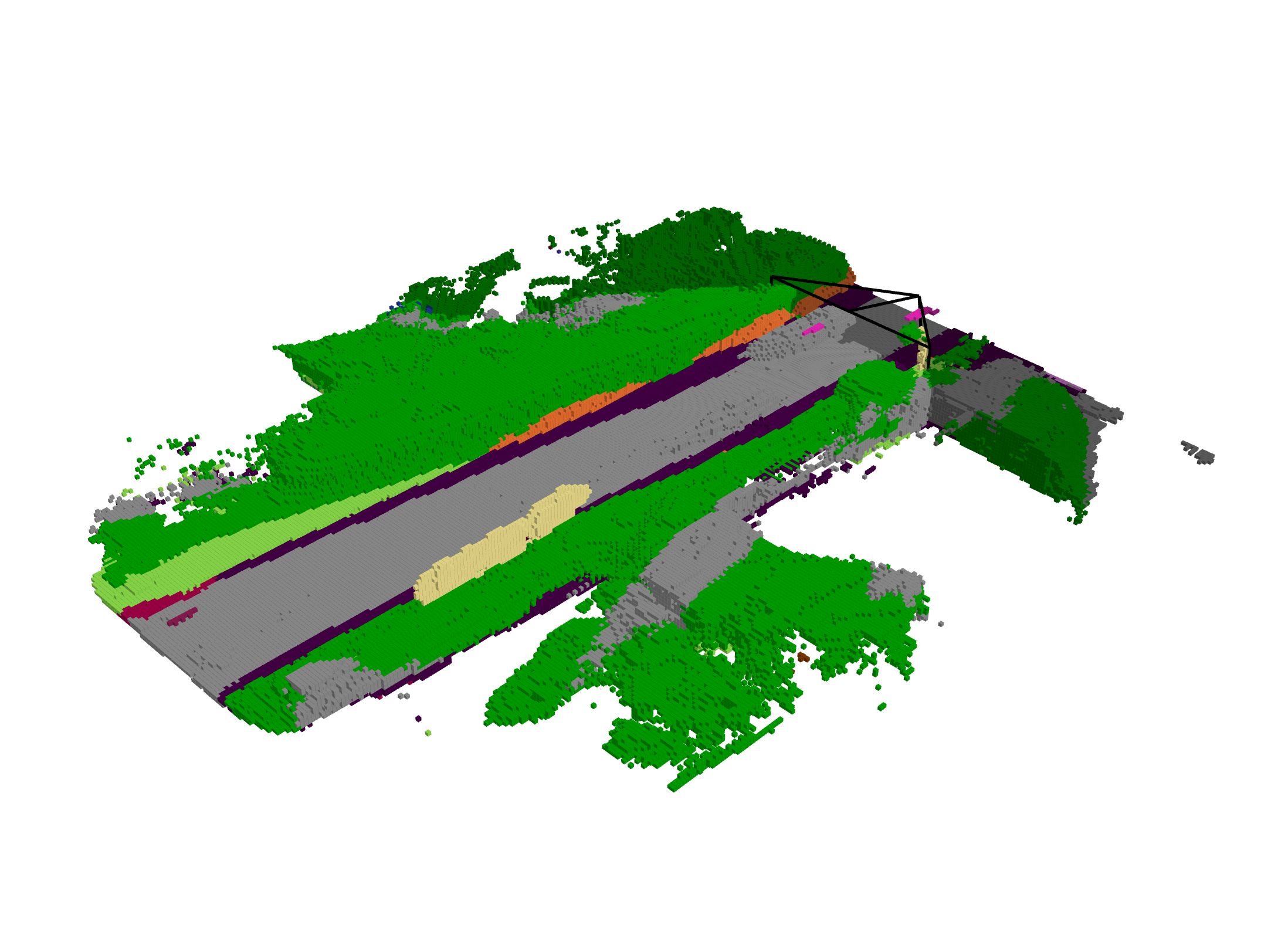}}
\end{minipage}
\begin{minipage}{0.23\linewidth}
    \vspace{3pt}
    \centerline{\includegraphics[width=\textwidth, viewport=200 250 2000 1300,clip]{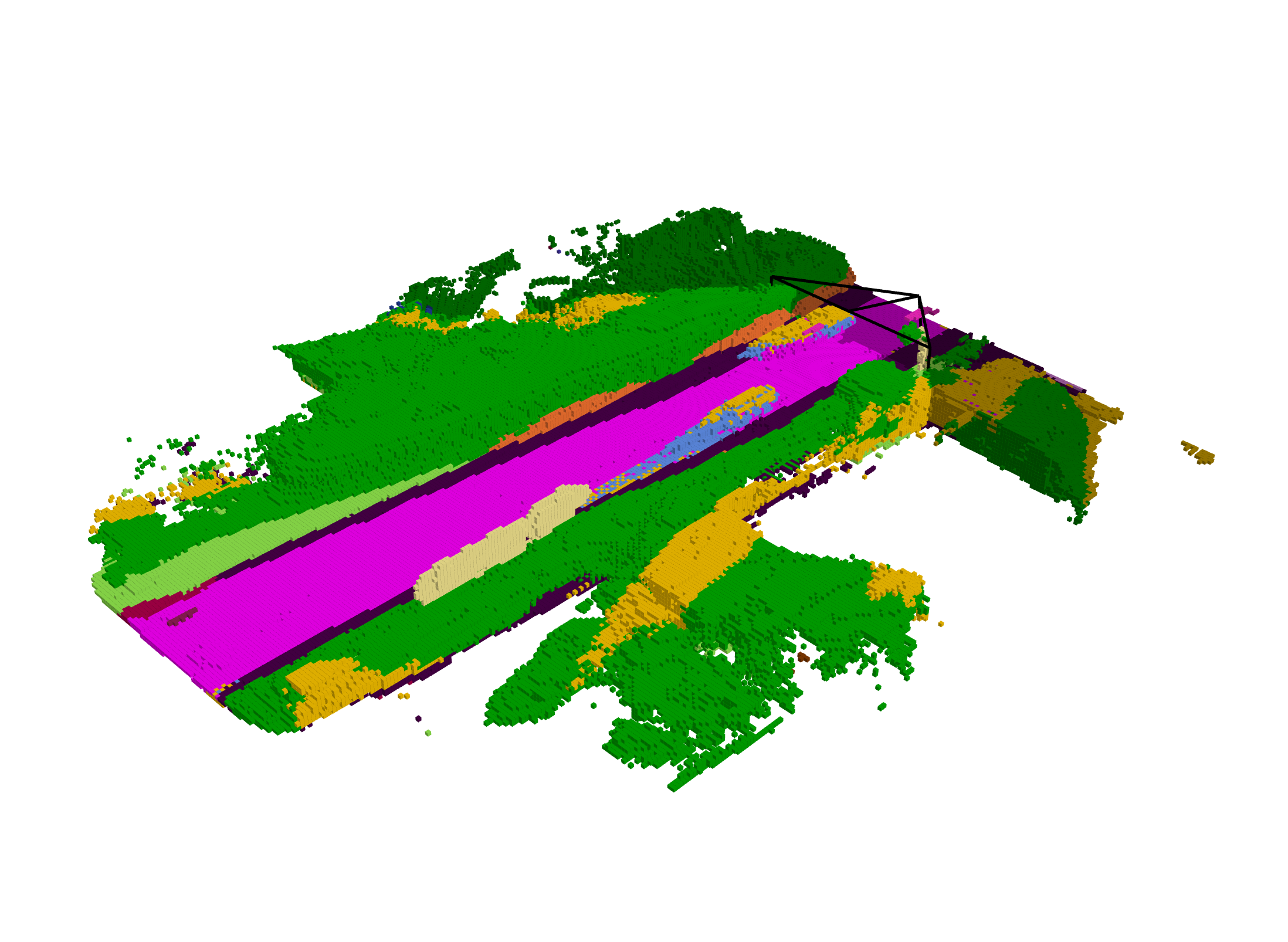}}
\end{minipage}
\begin{minipage}{0.23\linewidth}
    \vspace{3pt}
    \centerline{\includegraphics[width=\textwidth, viewport=200 250 2000 1300,clip]{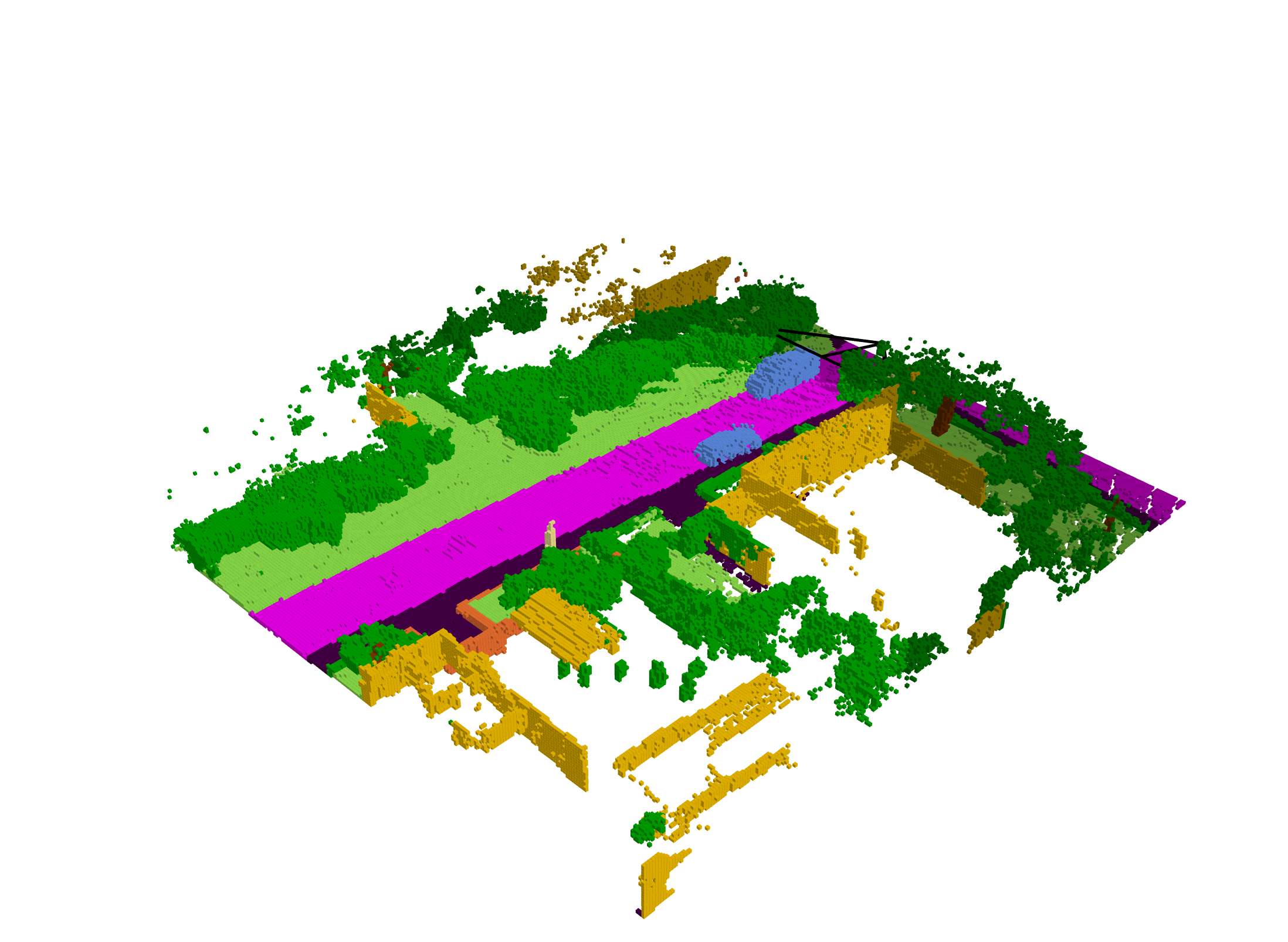}}
\end{minipage}

\begin{minipage}{0.23\linewidth}
    \vspace{3pt}
    \centerline{\includegraphics[width=\textwidth]{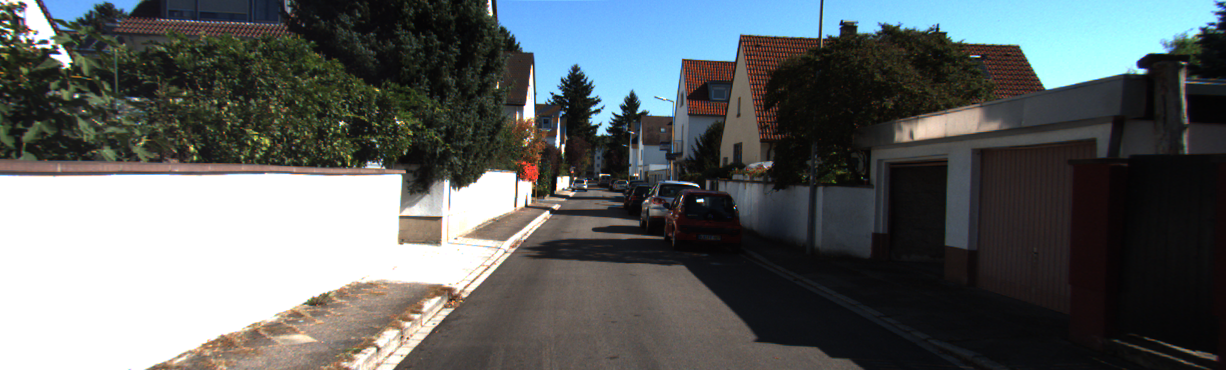}}
\end{minipage}
\begin{minipage}{0.23\linewidth}
    \vspace{3pt}
    \centerline{\includegraphics[width=\textwidth, viewport=200 250 2000 1300,clip]{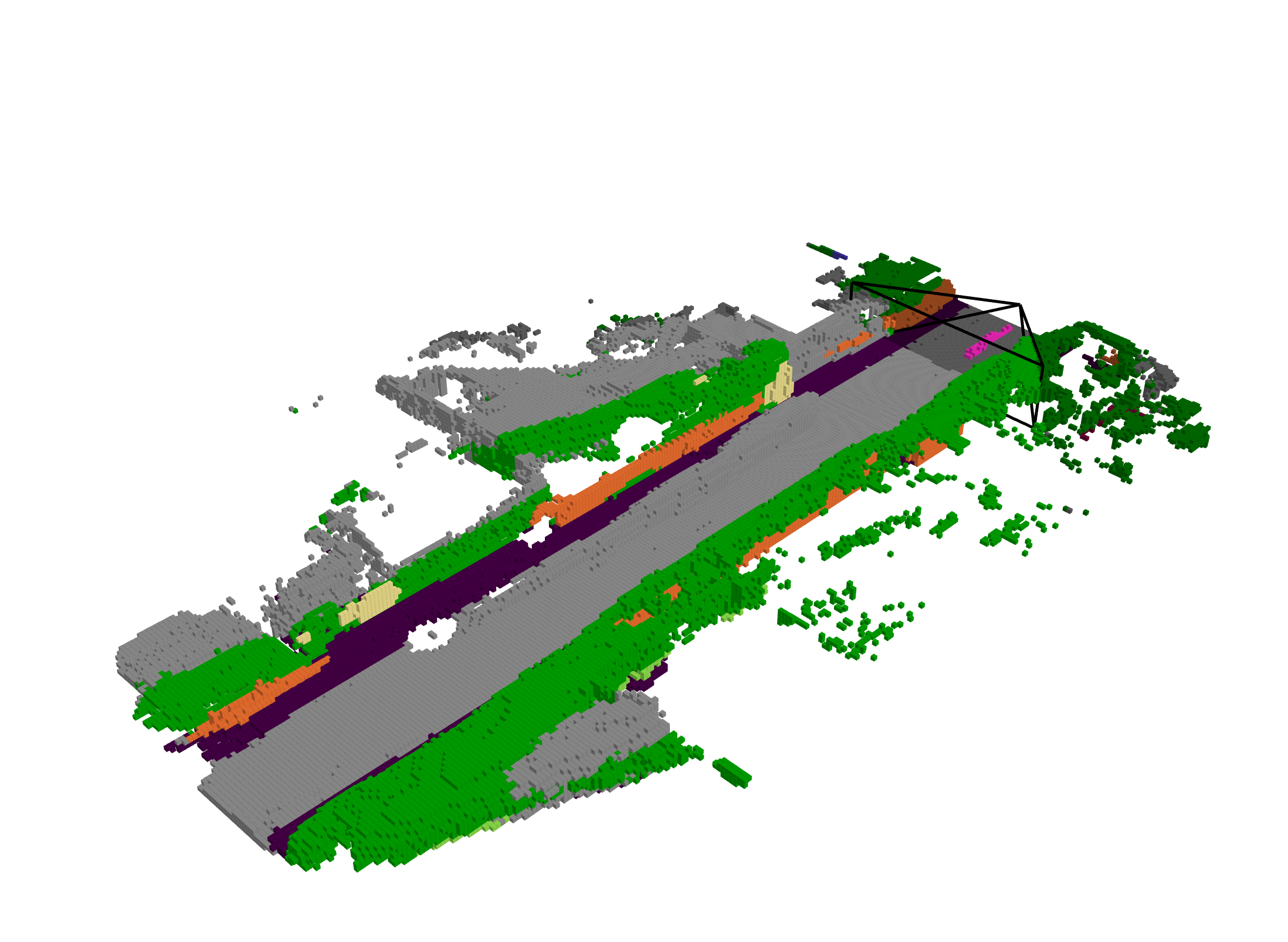}}
\end{minipage}
\begin{minipage}{0.23\linewidth}
    \vspace{3pt}
    \centerline{\includegraphics[width=\textwidth, viewport=200 250 2000 1300,clip]{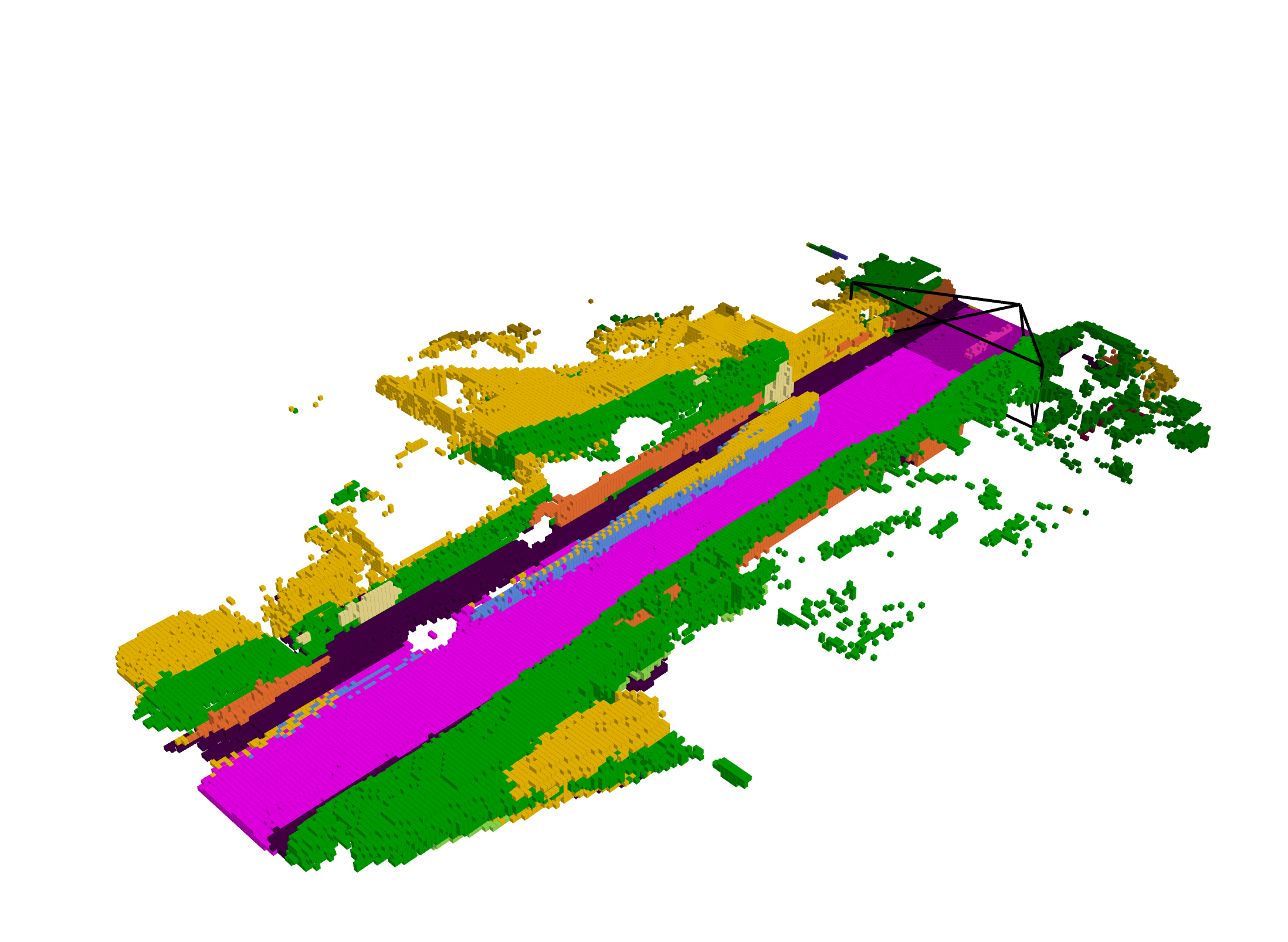}}
\end{minipage}
\begin{minipage}{0.23\linewidth}
    \vspace{3pt}
    \centerline{\includegraphics[width=\textwidth, viewport=200 250 2000 1300,clip]{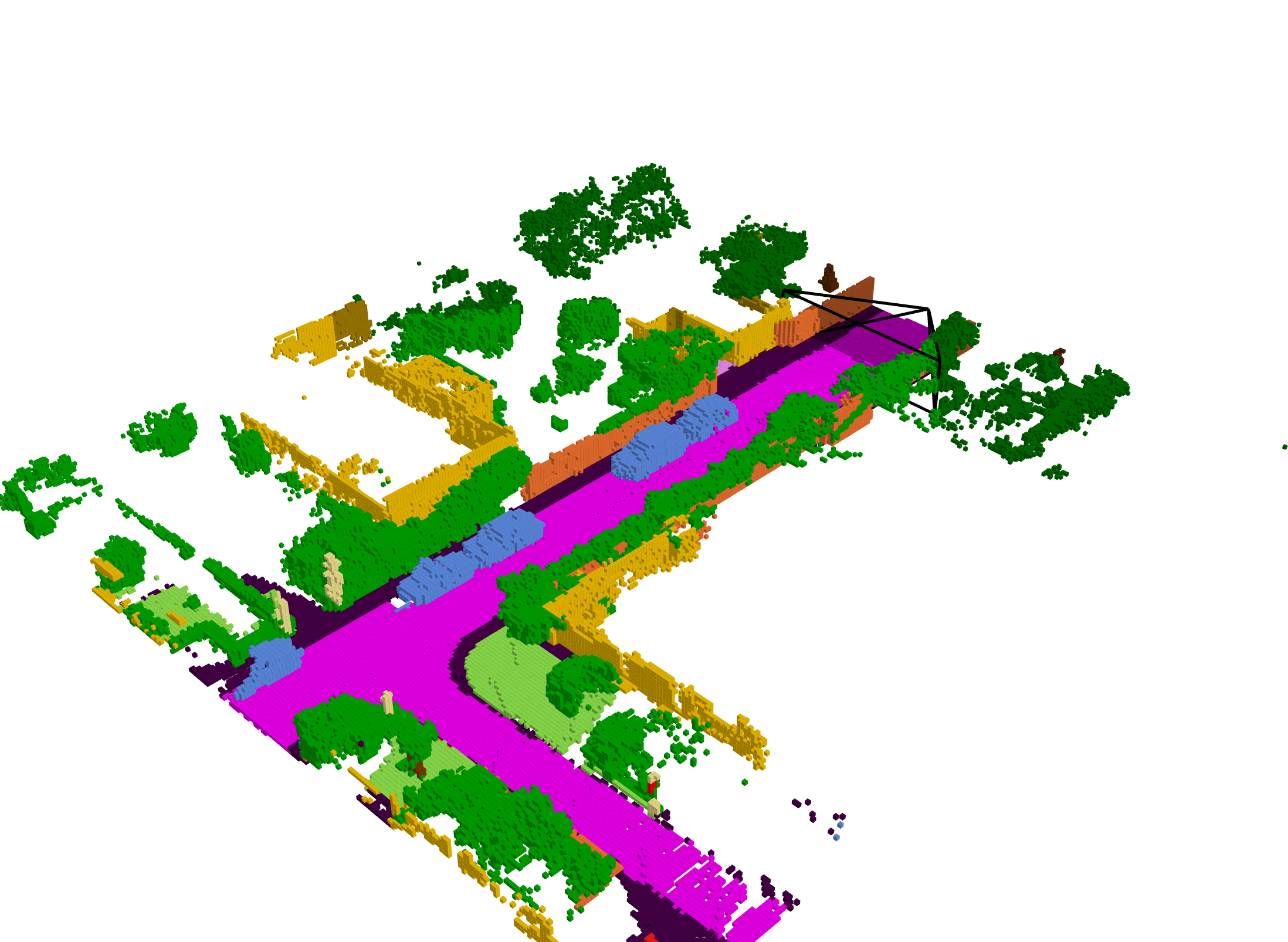}}
\end{minipage}

\begin{minipage}{0.23\linewidth}
    \vspace{3pt}
    \centerline{\includegraphics[width=\textwidth]{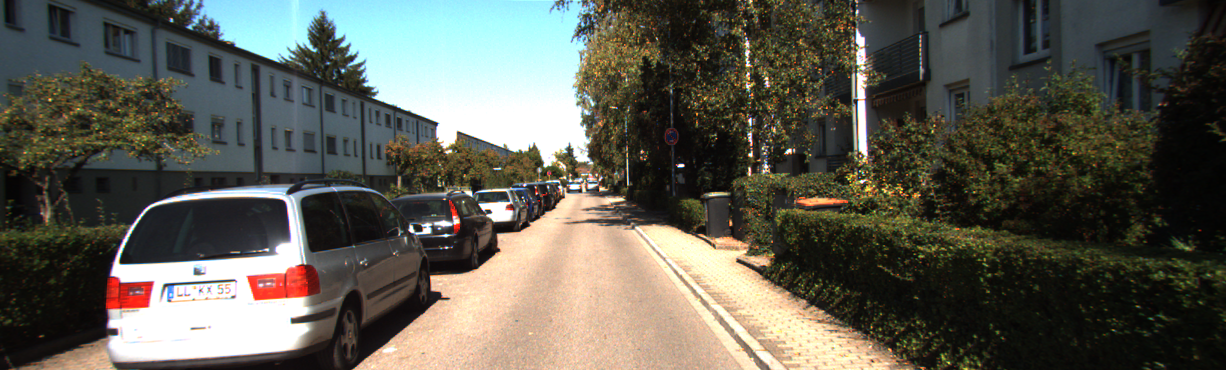}}
\end{minipage}
\begin{minipage}{0.23\linewidth}
    \vspace{3pt}
    \centerline{\includegraphics[width=\textwidth, viewport=200 250 2000 1300,clip]{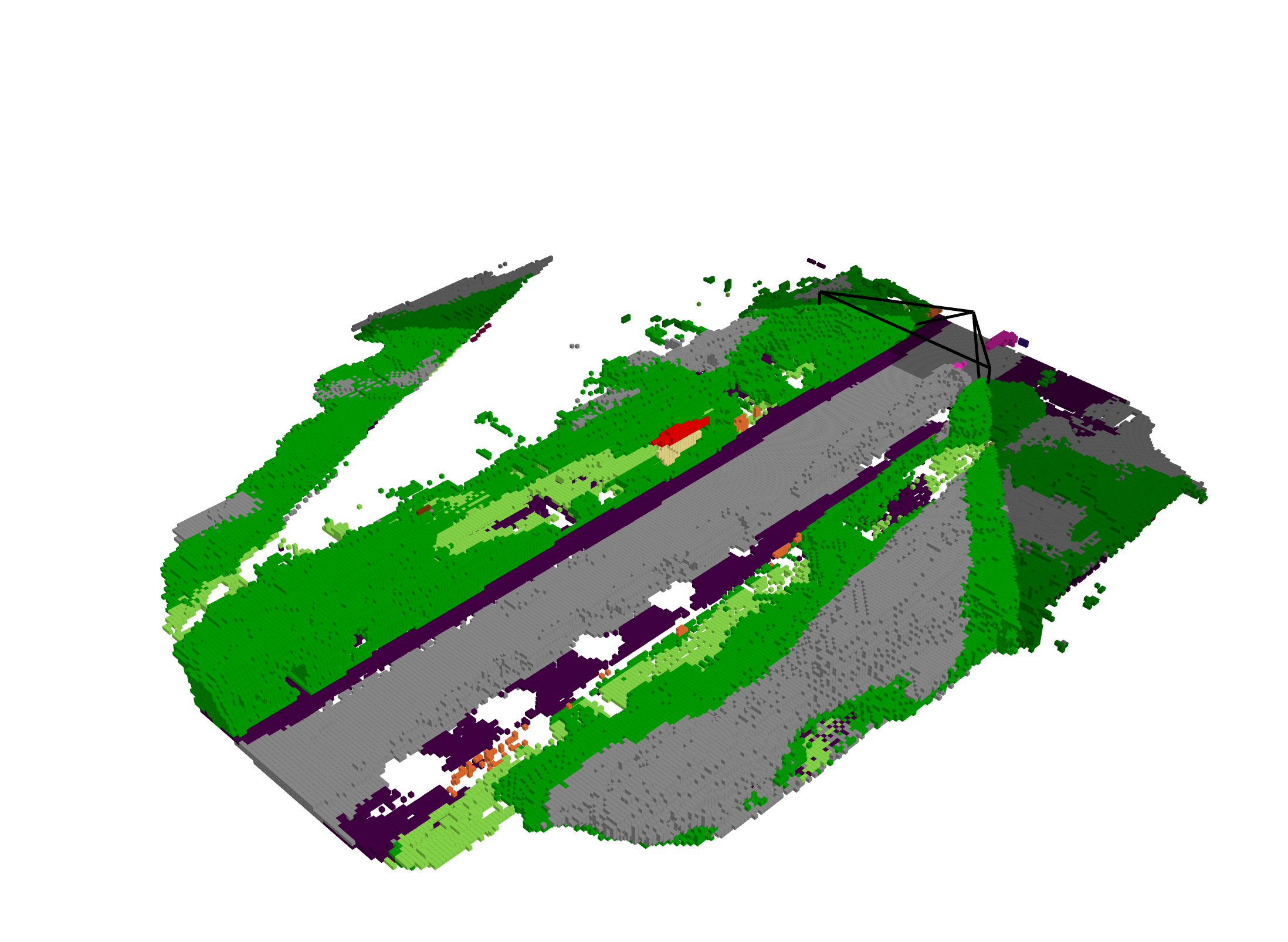}}
\end{minipage}
\begin{minipage}{0.23\linewidth}
    \vspace{3pt}
    \centerline{\includegraphics[width=\textwidth, viewport=200 250 2000 1300,clip]{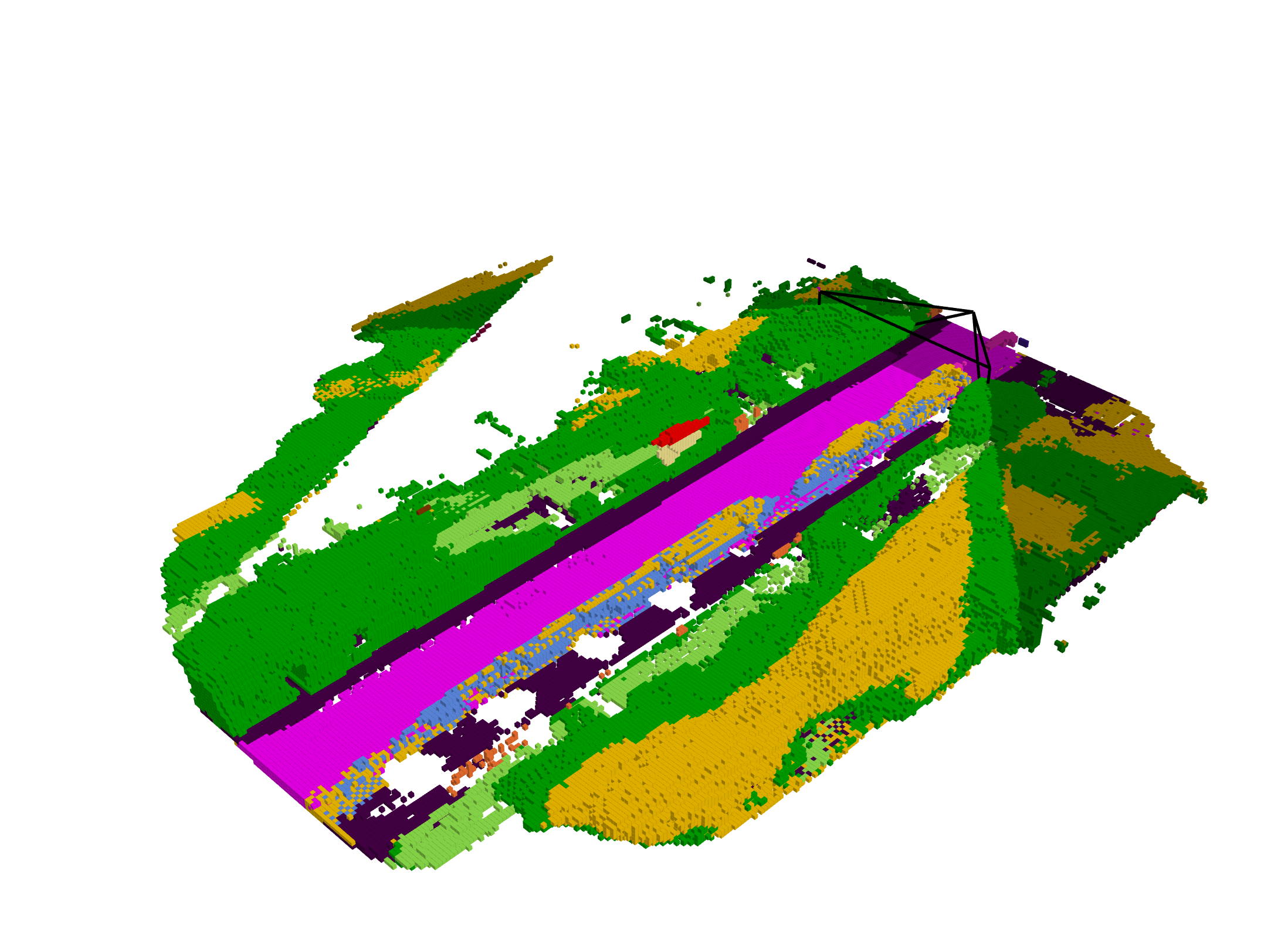}}
\end{minipage}
\begin{minipage}{0.23\linewidth}
    \vspace{3pt}
    \centerline{\includegraphics[width=\textwidth, viewport=200 250 2000 1300,clip]{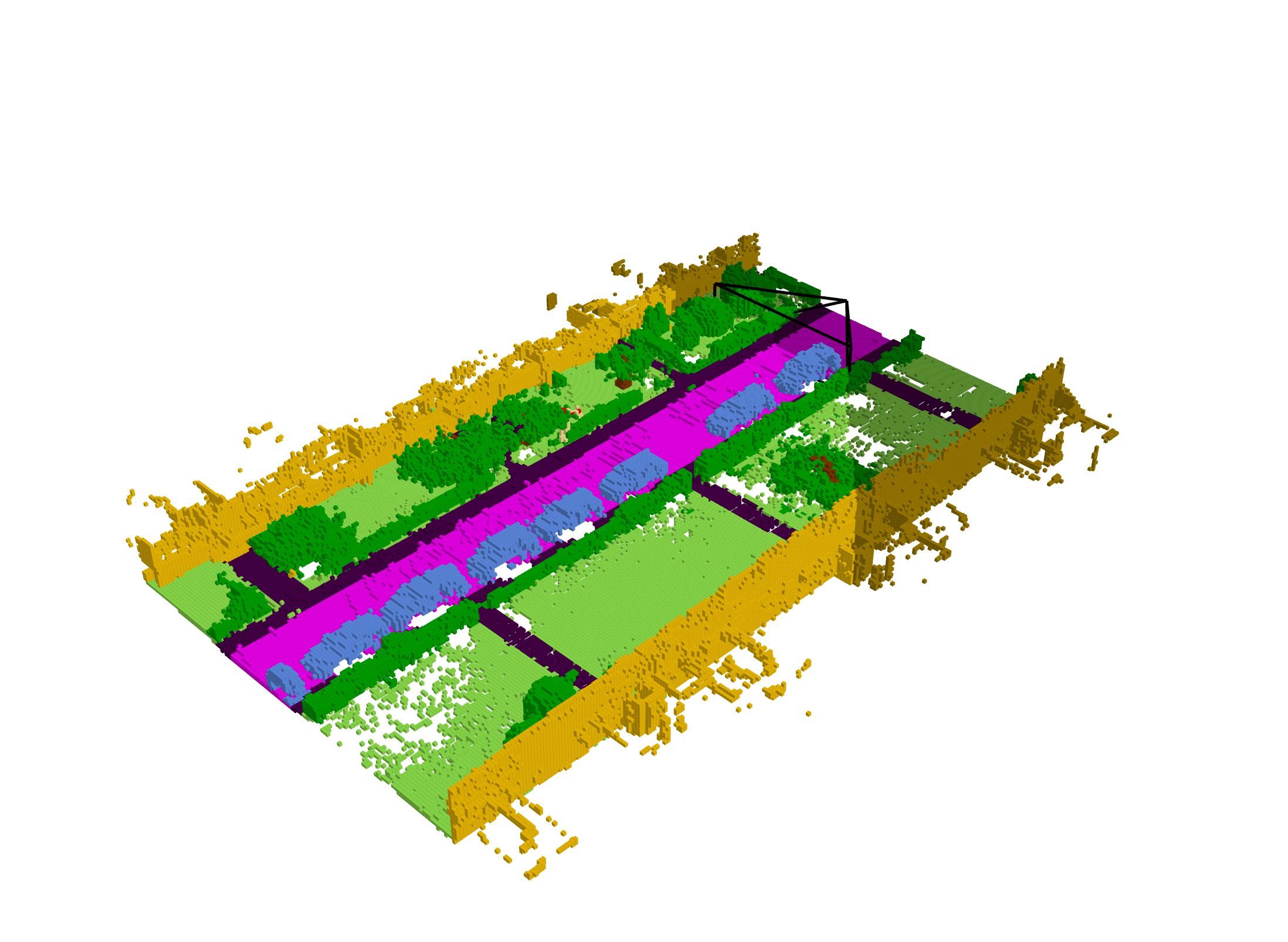}}
\end{minipage}

\begin{minipage}{0.23\linewidth}
    \centering
    RGB input
\end{minipage}
\begin{minipage}{0.23\linewidth}
    \centering
    MonoScene
\end{minipage}
\begin{minipage}{0.23\linewidth}
    \centering
    OVO (ours)
\end{minipage}
\begin{minipage}{0.23\linewidth}
    \centering
    GT
\end{minipage}

\begin{minipage}{\linewidth}
    \centering
    \raisebox{0.5ex}{\colorbox{bicycle}{}}~bicycle \raisebox{0.5ex}{\colorbox{car}{}}~car \raisebox{0.5ex}{\colorbox{motorcycle}{}}~motorcycle \raisebox{0.5ex}{\colorbox{truck}{}}~truck \raisebox{0.5ex}{\colorbox{otherVehicle}{}}~other vehicle \raisebox{0.5ex}{\colorbox{person}{}}~person \raisebox{0.5ex}{\colorbox{bicyclist}{}}~bicyclist \raisebox{0.5ex}{\colorbox{motorcyclist}{}}~motorcyclist \raisebox{0.5ex}{\colorbox{road}{}}~road \raisebox{0.5ex}{\colorbox{parking}{}}~parking \raisebox{0.5ex}{\colorbox{sidewalk}{}}~sidewalk \raisebox{0.5ex}{\colorbox{otherGround}{}}~other ground \raisebox{0.5ex}{\colorbox{building}{}}~building \raisebox{0.5ex}{\colorbox{fence}{}}~fence \raisebox{0.5ex}{\colorbox{vegetation}{}}~vegetation \raisebox{0.5ex}{\colorbox{trunk}{}}~trunk \raisebox{0.5ex}{\colorbox{terrain}{}}~terrain \raisebox{0.5ex}{\colorbox{pole}{}}~pole \raisebox{0.5ex}{\colorbox{trafficSign}{}}~traffic sign \raisebox{0.5ex}{\colorbox{unknown}{}}~unknown
\end{minipage}

\caption{\small \textbf{Qualitative results on SemanticKITTI.} We use the same visualization color code for novel class voxels following Figure~\ref{fig:nyu_qualitative}. The novel classes used for SemanticKITTI dataset include {\color{car} “car”}, {\color{road} “road”}, and {\color{building} “building”}. The visualization in the last row demonstrates that our model is capable of reasonable completion for ``road'' regions outside the field of view, showcasing the effectiveness of our model in handling scenes beyond the visible range.}
\label{fig:app_kitti_qualitative}
\end{figure*}